%% file: main.tex
\documentclass[10pt,twocolumn,letterpaper]{article}

\usepackage[pagenumbers]{cvpr} %
\usepackage{amsmath}
\usepackage{array}
\usepackage{pifont}
\usepackage{xcolor}
\usepackage{multirow}

\input{preamble}
\usepackage{listings}
\usepackage{xcolor}
\usepackage{float}
\usepackage{listings}
\definecolor{cvprblue}{rgb}{0.21,0.49,0.74}

\newlength\savewidth
\newcommand\shline{\noalign{\global\savewidth\arrayrulewidth \global\arrayrulewidth 1pt}\hline\noalign{\global\arrayrulewidth\savewidth}}
\newcommand{\tablestyle}[2]{\setlength{\tabcolsep}{#1}\renewcommand{\arraystretch}{#2}\centering\footnotesize}
\renewcommand{\paragraph}[1]{\vspace{1.25mm}\noindent\textbf{#1}}

\usepackage{algorithm}
\usepackage{algpseudocode}
\usepackage{subcaption}

\usepackage{colortbl}
\usepackage{xcolor}

\definecolor{hr}{gray}{0.7}  %
\definecolor{dt}{HTML}{ADCAD8}  %

\newcommand{\xhdr}[1]{\vspace{4pt} \noindent {\textbf{#1}}}

\newcolumntype{*}{>{\global\let\currentrowstyle\relax}}
\newcolumntype{^}{>{\currentrowstyle}}

\newcolumntype{H}{>{\setbox0=\hbox\bgroup}c<{\egroup}@{}}
\newcolumntype{Z}{>{\setbox0=\hbox\bgroup}c<{\egroup}@{\hspace*{-\tabcolsep}}}

\newcommand{\cmark}{\ding{51}} %
\newcommand{\xmark}{\ding{55}} %

\usepackage[pagebackref,breaklinks,colorlinks,allcolors=cvprblue]{hyperref}

\hypersetup{
    colorlinks=true,
    linkcolor=magenta,
    urlcolor=magenta
}

\title{VideoAuteur: Towards Long Narrative Video Generation}

\author{
Junfei Xiao\textsuperscript{1}, Feng Cheng\textsuperscript{2}, Lu Qi\textsuperscript{2}, Liangke Gui\textsuperscript{2}, Jiepeng Cen\textsuperscript{2}, Zhibei Ma\textsuperscript{2}, Alan Yuille\textsuperscript{1}, Lu Jiang\textsuperscript{2}
\vspace{2mm}
\\
\textsuperscript{1}Johns Hopkins University \hspace{1cm}
\textsuperscript{2}ByteDance
\vspace{2mm}
\\
{ Project Page: \href{https://videoauteur.github.io}{https://videoauteur.github.io}}
}

\begin{document}

\twocolumn[{%
\renewcommand\twocolumn[1][]{#1}%
\maketitle  %
\vspace{-8mm}  %
\begin{center}
    \centering
    \includegraphics[width=\linewidth]{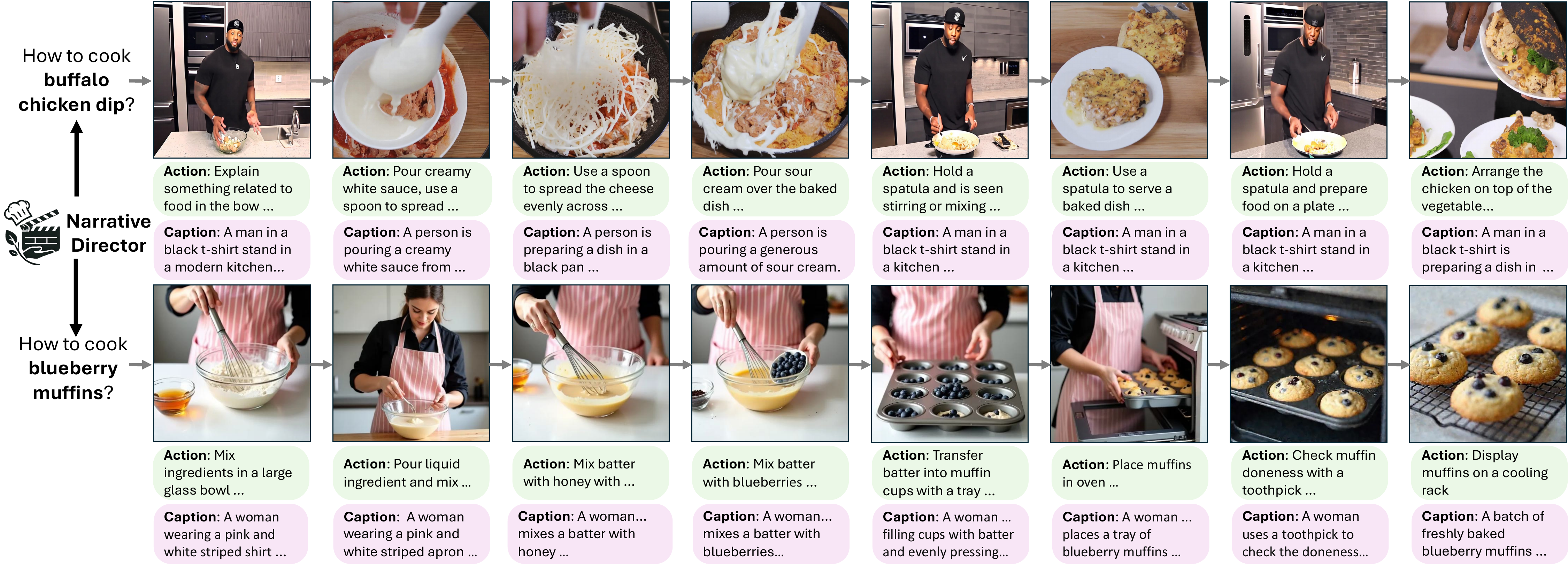}  %
    \vspace{-5mm}  %
    \captionof{figure}{\textbf{Long Narrative Video Generation.} We curate a large-scale cooking video dataset to develop an interleaved auto-regressive model -- \textbf{VideoAuteur}, which acts as a narrative director, sequentially generating actions, captions, and keyframes (two generated examples here). These elements condition a video generation model to create long narrative videos.}   

    \label{fig:teaser}
\end{center}%
}]

\input{sec/0_abstract}

\input{sec/1_intro_lu}

\input{sec/2_related_work}

\input{sec/4_data}

\input{sec/3_method}

\input{sec/5_experiments}

\input{sec/6_discussion}
\input{sec/7_conclusion}

{
    \small
    \bibliographystyle{ieeenat_fullname}
    \bibliography{main}
}

\input{sec/X_suppl}

\end{document}

%% file: sec/0_abstract.tex
\begin{abstract}
Recent video generation models have shown promising results in producing high-quality video clips lasting several seconds. However, these models face challenges in generating long sequences that convey clear and informative events, limiting their ability to support coherent narrations. In this paper, we present a large-scale cooking video dataset designed to advance long-form narrative generation in the cooking domain. We validate the quality of our proposed dataset in terms of visual fidelity and textual caption accuracy using state-of-the-art Vision-Language Models (VLMs) and video generation models, respectively. We further introduce a Long Narrative Video Director to enhance both visual and semantic coherence in generated videos and emphasize the role of aligning visual embeddings to achieve improved overall video quality. Our method demonstrates substantial improvements in generating visually detailed and semantically aligned keyframes, supported by finetuning techniques that integrate text and image embeddings within the video generation process. 
\end{abstract}

%% file: sec/1_intro_lu.tex
\section{Introduction}
\label{sec:intro}

Video generation~\cite{ho2022video,singer2022make,chen2023videocrafter1,chen2024videocrafter2,ho2022imagen,wang2023modelscope,zhou2022magicvideo} has witnessed remarkable advancements with diffusion~\cite{peebles2023scalable,brooks2024video,hong2022cogvideo,yang2024cogvideox} and auto-regressive models~\cite{kondratyuk2023videopoet,sun2023emu,sun2024generative,wang2024emu3}. A primary objective is to generate video clips from text prompts and supports various downstream applications, such as image animation~\cite{chen2025livephoto,xu2024magicanimate}, video editing~\cite{feng2024ccedit,ceylan2023pix2video}, video stylization~\cite{huang2024style}.

With the maturity of generating high-fidelity short video clips, researchers begin setting their sights on the next north-star: creating videos capable of conveying a complete narrative which captures an account of events unfolding over time. The importance of narratives has been highlighted in the literature. For example, Bruner argues that narratives are essential tools for organizing experiences and memories \cite{bruner1991narrative}. The book \emph{Sapiens: A Brief History of Humankind} emphasizes that the ability to share narratives (stories) has been pivotal in human development, setting humans apart from other animals~\cite{harari2014sapiens}.

Long \textbf{N}arrative \textbf{V}ideo \textbf{G}eneration (NVG) introduces several challenges.
One particularly challenge is the scarcity of video data suitable for learning coherent narratives in video. While our community has developed many video datasets, most are unsuitable for NVG. First, most videos are tagged with descriptions that are partially to NVG. Second, even for the relevant descriptions, these descriptions may be either too coarse or lack detailed actions needed for NVG. Finally, not all videos contain meaningful narratives suitable for learning and can be well evaluated. 

Consequently, video data with clear, complete, and meaningful narratives is crucial not only for training but also for evaluating and comparing NVG methods. However, compared to story generation through a sequence of images~\cite{huang2016visual, gupta2018imagine, yang2024seedstory, maharana2021integrating}, progress in narrative video generation has been relatively slow, partly due to the absence of standardized training and evaluation benchmarks.

This paper contributes to advancing research in narrative video generation in two ways. First, we curate and annotate a large-scale video dataset on the cooking domain. The samples in our dataset are structured with clear narrative flows, each composed of sequential actions and visual states. Our dataset consists of approximately 200,000 video clips, with an average duration of 9.5 seconds per clip. We select cooking videos for their well-defined and less ambiguous narratives, making them more objective to evaluate consistently. To address video copyright concerns, we source videos from existing video datasets, YouCook2~\cite{zhou2018youcook2} and HowTo100M~\cite{miech2019howto100m}. We design various mechanisms to ensure high-quality videos and captions, organized in a structured storyboard format, as illustrated in \Cref{fig:teaser}.

Additionally, we propose a new auto-regressive pipeline for long narrative video generation, comprising three main components: a long narrative director, a rolling-context conditioned keyframe renderer, and a visual-conditioned video generation model. The long narrative director produces a coherent narrative flow by generating a sequence of visual embeddings or keyframes that represent the story's logical progression. Building upon this, the rolling-context conditioned keyframe renderer utilizes a rolling history of reference images as contextual conditioning to generate high-quality keyframes with consistency. Finally, the visual-conditioned video generation model produces video clips based on these visual conditions to do narrative.

Extensive experiments on the large-scale collected dataset demonstrate the effectiveness of the proposed pipeline for long narrative video generation. To sum up, our contributions are as follows:

\begin{itemize}
    \item We construct \textbf{CookGen}, a large-scale, structured dataset accompanied by an effective data pipeline to benchmark long-form narrative video generation. The dataset along with the necessary functionalities will be opensourced to advance future research in the area.    
    \item We propose \textbf{VideoAuteur}, a novel approach for automated long video generation. It effectively bridges interleaved auto-regressive multimodal LLMs with pretrained DiTs, employing a rolling context strategy for enhanced generation quality and visual consistency. 
    \item Extensive experimental results and ablation studies show that \textbf{VideoAuteur} achieves the state-of-the-art performance in long narrative video generation. %
\end{itemize}

%% file: sec/2_related_work.tex
\vspace{-3mm}
\section{Related Works}
\label{sec:related_works}
\vspace{-2mm}

\xhdr{Text-to-Image/Video Generation}
Text-to-image~\cite{rombach2022high,sdxl,anydoor,yi2024diffusion,customdiffusion,qi2024unigs,wang2024semflow} and video generation~\cite{ho2022video,singer2022make,chen2023videocrafter1,chen2024videocrafter2,ho2022imagen,wang2023modelscope,zhou2022magicvideo} have made remarkable progress to generate high-fidelity video clip of 5-10 seconds. For example, latent design~\cite{rombach2022high} has become mainstream, balancing effectiveness with efficiency. Building upon this design, diffusion-based models like DiT~\cite{peebles2023scalable}, Sora~\cite{brooks2024video}, and CogVideo~\cite{hong2022cogvideo,yang2024cogvideox} leveraged larger datasets and explored refined architectures and loss functions to enhance performance. In contrast, auto-regressive models such as VideoPoet~\cite{kondratyuk2023videopoet} and Emu series~\cite{sun2023emu,sun2024generative,wang2024emu3} sequentially predict image or video tokens. Instead, our work focuses on the model's ability to generate long narrative videos beyond a few seconds. 
\begin{figure*}[t!]
    \centering
    \includegraphics[width=\linewidth]{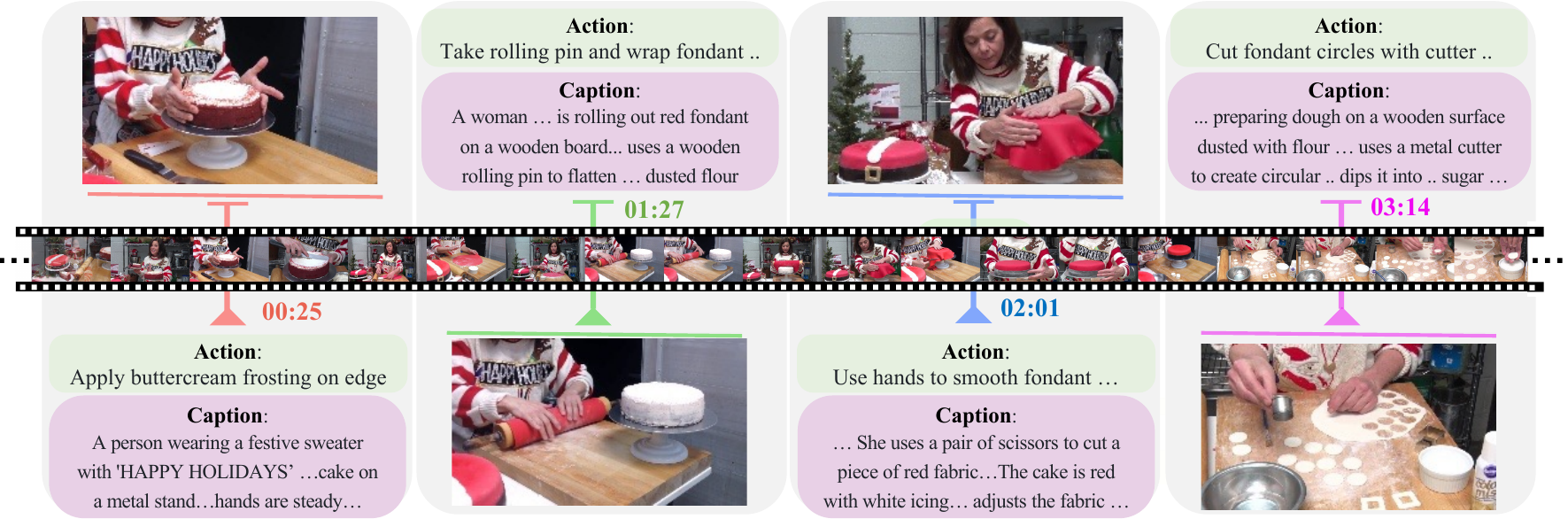}
    \vspace{-4mm}
    \caption{\textbf{CookGen contains long narrative videos annotated with actions and captions.} Each source video is cut into clips and matched with the labeled ``actions''. We use refined pseudo labels from ASR for Howto100M videos and use manual annotations for Youcook2 videos. We use state-of-the-art VLMs (\ie GPT-4o and an expert captioner) to provide high-quality captions for all video clips.}
    \vspace{-4mm}
    \label{fig:data_example_with_annotation}
\end{figure*}

\xhdr{Interleaved Image-Text Modeling}
Interleaved image-text generation~\cite{dong2023dreamllm,tian2024mm,ge2024seedx,yang2024seed,alayrac2022flamingo,ge2024seedx} has garnered attention as a compelling research area that merges visual and textual modalities to produce rich outputs. Earlier approaches~\cite{radford2021learning,li2023scaling,radford2021learning,sun2023eva} primarily relied on large-scale image-text paired datasets~\cite{gadre2024datacomp,schuhmann2022laion} but were often confined to single-modality tasks, such as captioning or text-to-image generation. With the emergence of large language models ~\cite{touvron2023llama}, various vision-language models ~\cite{li2023blip,liu2024visual,wang2024visionllm} have stepped in a new era of unified representations, leveraging well-curated datasets for interleaved generation. However, most existing works focus on the one-time generation and do not address the coherence of generated content, which is our focus.

\xhdr{Narrative Visual Generation} Existing narrative visual generation primarily focuses on addressing challenges related to semantic and visual consistency. Recent approaches such as Narrative Visual Generation, VideoDirectorGPT~\cite{Lin2023VideoDirectorGPT}, Vlogger~\cite{zhuang2024vlogger}, Animate-a-story~\cite{he2023animateastory}, VideoTeris~\cite{tian2024videotetris}, IC Lora~\cite{lhhuang2024iclora}, Vlogger~\cite{zhuang2024vlogger}, and Animate-a-story~\cite{he2023animateastory} employ various methods to enhance semantic coherence and visual continuity. Unlike most prior methods that mainly focus on consistent image generation~\cite{lhhuang2024iclora, yang2024seedstory,zhou2024storydiffusion}, our target is generating coherent narrative videos.
While some works make efforts to be language-centric using text as conditions for video generation~\cite{zhuang2024vlogger, xu2024magicanimate} or appending with keyframes~\cite{zhao2024moviedreamer}, different from these work, we propose an integrated approach that leverages multi-modal large language models (LLMs) in conjunction with in-context diffusion transformer models to ensure global narrative coherence, subsequently conditioning the video generation model.

%% file: sec/4_data.tex
\section{CookGen: a Long Narrative Video Dataset}
\label{sec:long_narrative_data}

To the best of our knowledge, datasets for long narrative video generation research is extremely limited.
To enable in-depth exploration and establish an experimental setting,
we establish \textbf{CookGen}, a large video dataset with detailed annotations on captions, actions, and annotations. As the data example provided in \Cref{fig:data_example_with_annotation}, our dataset focuses on cooking videos. We prioritize cooking over other video categories because each dish follows a pre-defined, strict sequence of action steps. These structured and unambiguous objectives in cooking videos are essential for learning and evaluating long video narrative generation.

\subsection{Overview}

We source over 30,000 raw videos about from two existing video datasets: YouCook2~\cite{zhou2018youcook2} and HowTo100M~\cite{miech2019howto100m}.
Each video is filtered and cropped with processing to remove corruptions. \Cref{tab:data_sources} provides detailed information about the dataset statistics, video and clip details, and the train/val partitioning. ~\Cref{sec:appendix_additional_data_stats} provides more details.

Table~\ref{tab:dataset_comparison} compares our dataset with existing datasets most relevant to multimodal narrative generation. Unlike existing datasets that primarily focus on image-based comic story generation, our real-world narrative dataset offers several advantages. First, the videos in our dataset depict procedural activities (\ie, cooking), providing unambiguous narratives that are easier to annotate and evaluate. Second, our dataset contains 150$\times$ the number of frames compared to the previous largest dataset, StoryStream. Third, we offer 5$\times$ denser textual descriptions, with an average of 763.8 words per video. These advantages make our dataset a better resource for narrative video generation.

\begin{table}[t!]
    \centering
    \scriptsize
    \tablestyle{1.1pt}{1.2}  %
    \begin{tabular}{c c c c c}
        \textbf{Data Source} & \textbf{\# Vid. (train/val)} & \textbf{\# Clips} & \textbf{Clip Len.} & \textbf{\# Clips / Vid.} \\ \shline
        YouCook2 & 1333 / 457 & $\sim$10K & 19.6s & 7.7 \\
        HowTo100M (subset) & 30039 / 933 & $\sim$183K & 9.5s & 5.9 \\
    \end{tabular}
    \vspace{-3mm}
    \caption{\textbf{Long narrative dataset sources.} Our dataset is built upon Youcook2 and a cooking subset of Howto100M. }
     \vspace{-2mm}
    \label{tab:data_sources}
\end{table}

\begin{table}
    \centering
    \small
    \setlength{\tabcolsep}{4pt}
    \begin{tabular}{c c c c c}
        \textbf{Datasets} & \textbf{Modality} & \textbf{Type} & \textbf{\# Images} & \textbf{Text Length} \\ 
        \shline
        Flintstones & Image & Comic & 122k & 86 \\
        Pororo & Image & Comic & 74k &  74 \\
        StorySalon & Image & Comic & 160k & 106 \\
        StoryStream & Image & Comic & 258k & 146 \\
        VIST & Image & Real world & 210K & $\sim$70\\
        \hline
        \textbf{CookGen} & Video & Real world & \textbf{39M} & \textbf{763.8} \\
    \end{tabular}
    \caption{\textbf{Comparison with multi-modal narrative datasets.} Most existing datasets focus on image-based comic story generation. In contrast, our dataset consists of long narrative videos, containing 150$\times$ the number of frames and 5$\times$ the dense text annotations compared to the previous largest dataset, StoryStream.}
    \vspace{-4mm}
    \label{tab:dataset_comparison}
\end{table}

\subsection{Annotation and Processing}
To ensure scalability and quality, we design an efficient annotation pipeline to support the annotation as below.

\paragraph{Captions.} For open-source and scalability, we train a video captioner based on open-sourced VLM.  Inspired by LLaVA-Hound~\cite{zhang2024direct}, we begin by collecting a caption dataset using GPT-4o, with a focus on object attributes, subject-object interactions, and temporal dynamics. Subsequently, we fine-tune a captioning model based on LLaVA-NeXT~\cite{zhang2024llavanextvideo} to optimize captioning performance.

\paragraph{Actions.} We use HowTo100M ASR-based pseudo labels for `actions' in each video, further refined by LLMs to provide enhanced annotations of the actions throughout the video~\cite{shvetsova2025howtocaption}. This refinement improves the action quality to capture events and narrative context. However, the annotations are still noisy and sometimes not informative due to the inherent errors in ASR scripts.

\paragraph{Caption-Action Matching and Filtering.} To ensure alignment between captions and actions, we implement a matching process based on time intervals. Using Intersection-over-Union (IoU) as a metric, we evaluate whether the overlap between the captioned clip time and action time meets a threshold. An action is considered a match if the following conditions are met: the difference between the clip start time and the action start time (\texttt{start\_diff}) is less than 5 seconds; the clip end time is later than the action end time; and the IoU between the clip and action time intervals is greater than 0.25, or if IoU\textgreater 0.5. Here, \texttt{clip\_time} and \texttt{action\_time} represent the time intervals for the clip and action, respectively. Using this rule, we filter and match captions to actions, ensuring that each caption aligns with the relevant action. We found this step is important for creating narrative consistency throughout the video.

\paragraph{Annotation Quality Reverification.} High-quality captions are essential for narrative visual generation. To verify the quality of our annotations, we build an evaluation pipline of inverse generation  and visual understanding through VLM experts, which are detailed in Appendix \S\ref{sec:inverse_video_gen} and \S\ref{sec:vlm_consistency}.

%% file: sec/3_method.tex
\begin{figure*}[h!]
    \centering
    \includegraphics[width=\linewidth]{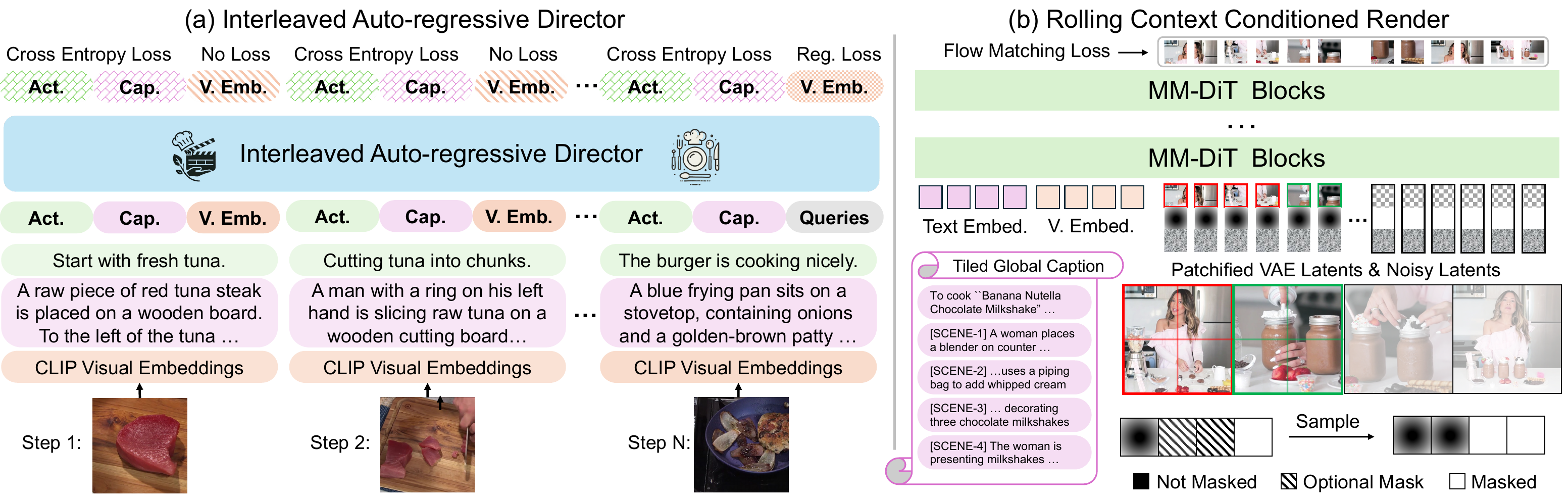}
    \vspace{-5mm}
    \caption{\textbf{Long Narrative Visual Condition Generation}. (a) {\textit{Interleaved Auto-regressive Director:}} an auto-regressive vison-language model, takes a user query (e.g., ``How to cook a tuna sandwich?") and an initial image-text pair as input. It then generates actions, captions, and visual states (\ie, visual embeddings) step-by-step. (b) \textit{Rolling Context Conditioned Render}: Apart from the semantics consistency through interleaved generation, we use a rolling of reference images as direct context conditions to further improve visual consistency with a diffusion transformer model. With them, a long narrative video can be created using these generated visual conditions (\ie, visual embeddings and/or keyframes derived from the interleaved director and the keyframe render with rolling context conditioning.)
    }
    \phantomsubcaption 
    \label{fig:long_narrative_director_a}
    \phantomsubcaption 
    \label{fig:long_narrative_director_b} 
    \label{fig:long_narrative_director}
\end{figure*}

\vspace{-2mm}
\section{Method}
\label{sec:method}

Given the text input, the task of long narrative video generation aims at generating a coherent long video $\mathcal{Y} \in \mathbb{R}^{H \times W \times F}$ that aligns with the progression of the text input sequentially. The $H$, $W$, and $F$ are generated videos' height, width, and frame numbers. To achieve this, we propose \textbf{VideoAuteur}, which involves three main components: an interleaved long narrative video director, a rolling-context conditioned keyframe renderer, and a visual-conditioned video generation model. The long narrative video director creates a sequence of language states and visual embeddings to represent the narrative flow (\S\ref{sec:long_narrative_director}). A pretrained DiT model then renders keyframes using a rolling history of reference images as contextual conditioning (\S\ref{sec:rolling_context_diffusion}). Finally, the video generation model produces video clips based on these visual conditions (\S\ref{sec:visual_condition_video_gen}).

\subsection{Long Narrative Interleaved Director} 
\label{sec:long_narrative_director}
As shown in \Cref{fig:long_narrative_director_a}, the long narrative video director generates a sequence of visual embeddings (or keyframes) that capture the narrative flow. The interleaved image-text director creates a sequence where text tokens and visual embeddings are interleaved, integrating narrative and visual content tightly. Using an auto-regressive model, it predicts the next token based on the accumulated context of both text and images. This helps maintain narrative coherence and align visuals with the text semantics.

\noindent\textbf{Interleaved auto-regressive model.} Our model performs next-token prediction for cross-modal generation, learning from sequences of interleaved image-text pairs with a context window size \( T \). Each text token is supervised with cross-entropy loss, and the final visual embedding \( \mathbf{z}_T \) is regressed using learnable query tokens, as illustrated in \Cref{fig:long_narrative_director}. The auto-regressive conditioning is given by:
\begin{align}
  p(\mathbf{y}_t \mid \mathbf{y}_{1:t-1}) = p(\mathbf{c}_t \mid \mathbf{c}_{1:t-1}) \cdot p(\mathbf{z}_t \mid \mathbf{c}_{1:t}, \mathbf{z}_{1:t-1}),
\end{align}
where \( \mathbf{c}_t \) represents texts and \( \mathbf{z}_t \) denotes visual embeddings.

\noindent\textbf{Regression latent space.} We utilize a CLIP-Diffusion visual autoencoder with a CLIP encoder \( E_{\text{clip}} \) and a diffusion decoder \( D_{\text{diff}} \) to encode raw images \(\mathbf{x}\) to visual embeddings for auto-regressive generation:
\begin{align}
\mathbf{z} = E_{\text{clip}}(\mathbf{x}), \quad \hat{\mathbf{x}} = D_{\text{diff}}(\mathbf{z})
\end{align}
This setup generates language-aligned visual embeddings and reconstructs images from them.

\noindent\textbf{Regression loss.} To align the generated visual latents \( \mathbf{z}_{\text{pred}} \) with the target latents \( \mathbf{z}_{\text{target}} \), we use a combined loss:
\begin{align}
L_{\text{reg}} = \alpha \left(1 - \frac{\mathbf{z}_{\text{pred}} \cdot \mathbf{z}_{\text{target}}}{\|\mathbf{z}_{\text{pred}}\| \|\mathbf{z}_{\text{target}}\|}\right) + \beta \frac{1}{N} \sum_{i=1}^N (\hat{z}_i - z_i)^2
\end{align}
where \( \alpha \) and \( \beta \) are hyper-parameters.

\noindent\textbf{Narrative from ``actions" to ``visual states".} The interleaved model generates a coherent narrative sequence by progressively conditioning each step on the cumulative context from previous steps, \Cref{fig:long_narrative_director}.  At each time step \( t \), the model generates an action \( \mathbf{a}_t \), a caption \( \mathbf{c}_t \), and a visual state \( \mathbf{z}_t \), conditioned on the cumulative history $\mathcal{H}_{t-1}$:

\begin{equation}
\begin{aligned}
\mathcal{H}_{t-1} = \{ \mathbf{a}_{1:t-1}, \mathbf{c}_{1:t-1}, \mathbf{z}_{1:t-1} \} \\
\mathbf{a}_t \mid \mathcal{H}_{t-1} \rightarrow \mathbf{c}_t \mid \{\mathcal{H}_{t-1}, \mathbf{a}_t\} \rightarrow \mathbf{z}_t \mid \{\mathcal{H}_{t-1}, \mathbf{a}_t, \mathbf{c}_t\}
\end{aligned}
\end{equation}
This layered conditioning improves coherence across the sequence, aligning actions, language, and visuals.

\subsection{Rolling Context Conditioned Render}
\label{sec:rolling_context_diffusion}

While the interleaved auto-regressive director model can learn visual consistency, the CLIP representation space struggles to preserve fine visual details (\eg, character features, clothing patterns), as demonstrated in Appendix \Cref{fig:autoencode_comparison}. To address this limitation and improve generation quality, we employ a pretrained Text-to-Image diffusion transformer model to render high-quality keyframes, conditioning on a rolling history of reference images. The context length can vary dynamically from 1 to 3, balancing flexibility and efficiency when generating keyframes.

As illustrated in \Cref{fig:long_narrative_director_b}, we use a rolling history of two reference images, \(I_t\) and \(I_{t-1}\). This setup is further conditioned by the tiled global caption
\begin{align}
    \mathbf{c}_{\text{tiled}} = \text{tiled}( \mathbf{c}_{t-3},\, \mathbf{c}_{t-2},\, \mathbf{c}_{t-1},\, \mathbf{c}_{t} ),
\end{align}
the predicted visual embeddings \(\mathbf{z}_t\) and \(\mathbf{z}_{t-1}\), as well as the reference images \(I_{t-3}\) and \(I_{t-2}\). 

\begin{equation}
\begin{aligned}
\text{D}(\mathbf{c}_{\text{tiled}},\, \mathbf{z}_{t-1},\, \mathbf{z}_t,\, I_{t-3},\, I_{t-2})  \rightarrow I_{t-1}, I_t,
\end{aligned}
\end{equation}
where \(\text{D}(\cdot)\) denotes the diffusion model for synthesizing a new keyframe \(I_t\) by integrating the rolling context of images, captions, and visual embeddings. This layered conditioning improves coherence across frames.

\vspace{0.3em}\noindent\textbf{Flow Matching Loss.} We employ a flow matching loss that aligns the learned drift function $f_{\theta}$ with the ground-truth path from $\mathbf{x}_{T}$ to $\mathbf{x}_{T+1}$. We define:
\begin{align}
    \mathcal{L}_{\text{flow}}(\theta)
    \;=\;
    \mathbb{E}_{\mathbf{x}_{T}, \,\mathbf{x}_{T+1}, \,T}
    \Bigl[
        \bigl\|
            f_{\theta}(\mathbf{x}_{T}, T)
            \;-\;
            \mathbf{v}(T)
        \bigr\|^2
    \Bigr],
\end{align}
where $\mathbf{v}(T)$ denotes the ideal drift path that transitions $\mathbf{x}_{T}$ towards $\mathbf{x}_{T+1}$. This objective enforces consistency across frames without relying on a separate diffusion loss.

\begin{figure}[t!]
    \centering
    \includegraphics[width=\linewidth]{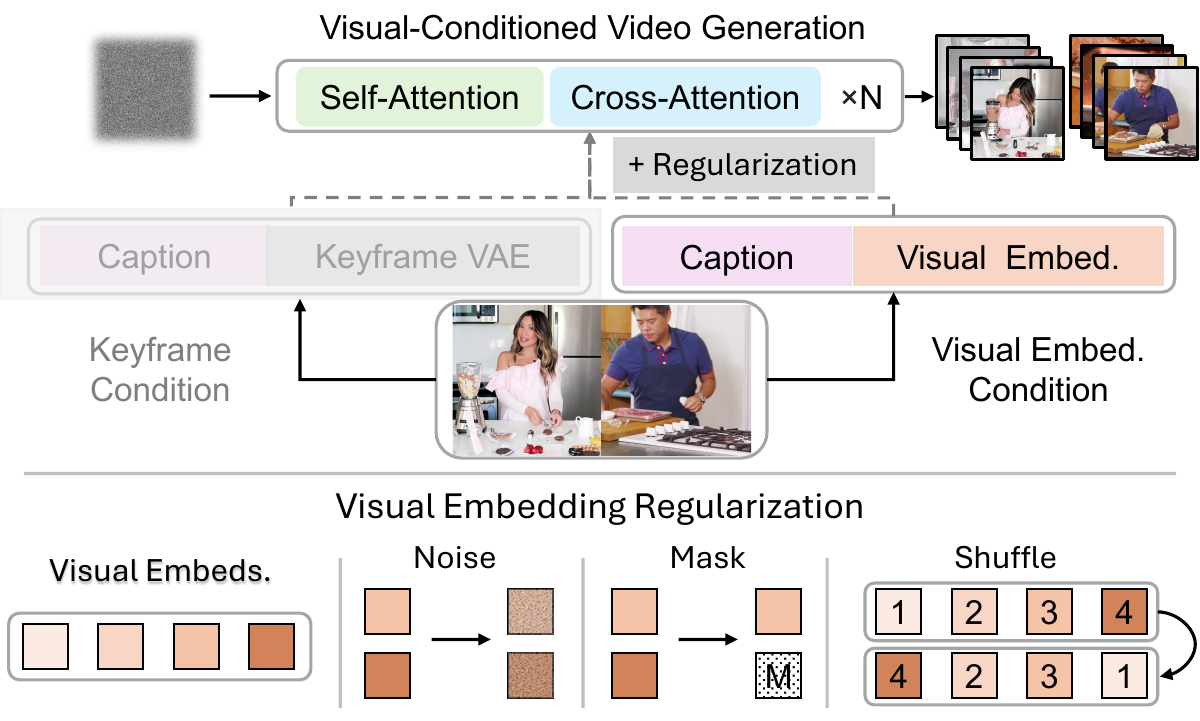}
    \vspace{-3mm}
    \caption{\textbf{Visual-conditioned video generation.} Our interleaved auto-regressive director and rolling context renderer generates both text and visual conditions, enabling the video generation process to be conditioned on keyframes (VAE embeddings) and CLIP latents. We apply Gaussian noise, random masking and random shuffling as regularization during the training process to improve robustness with the imperfect visual embeddings.}
    \vspace{-3mm}
    \label{fig:visual_conditioned_gen}
\end{figure}

\subsection{Visual-Conditioned Video Generation}
\label{sec:visual_condition_video_gen}

Using the sequence of actions \(\mathbf{a}_t\), captions \(\mathbf{c}_t\),visual states \(\mathbf{z}_t\) and keyframe \({I}_t\) generated by the interleaved director and rolling context conditioned render, we condition a video generation model to produce coherent long narrative videos. Unlike the classic Image-to-Video (I2V) pipeline that only uses an image as the starting frame, our approach leverages the regressed visual latents \(\mathbf{z}_t\) as continuous conditions throughout the sequence (see \S\ref{sec:method_beyond_keyframes}). Furthermore, we improve the robustness and quality of the generated videos by adapting the model to handle noisy visual embeddings, since the regressed visual latents may not be perfect due to regression errors and keyframe uncertainty (see \S\ref{sec:method_learn_from_noisy}).

\subsubsection{Visual Conditions Beyond Keyframes}
\label{sec:method_beyond_keyframes}
Conventional visual-conditioned video generation typically uses initial keyframes to guide the model, where each frame \( \mathbf{x}_t \) is generated as \( \mathbf{x}_t = D_{\text{visual}}(I_t) \). Our interleaved auto-regressive director supports generating visual states \( \mathbf{z}_t \) in a semantically aligned latent space, allowing direct conditioning from a pretrained visual decoder, as shown in \Cref{fig:visual_conditioned_gen}. By using these regressed visual latents \( \mathbf{z}_t \) directly, each frame is generated as \( \mathbf{x}_t = D_{\text{visual}}(\mathbf{z}_t) \). This follows the narrative and enhancing consistency by relying on narrative-aligned embeddings.

\vspace{-4mm}
\subsubsection{Learning from Noisy Visual Conditions}
\label{sec:method_learn_from_noisy}

To enhance the robustness over imperfect visual embeddings \(\mathbf{z}_t\) from the auto-regressive director, we fine-tune the model using noisy embeddings \(\mathbf{z}_t'\) defined by:
\begin{align}
\mathbf{z}_t' = \mathcal{S}(\mathcal{M}(\mathbf{z}_t + \boldsymbol{\epsilon}))
\end{align}
where \(\boldsymbol{\epsilon} \sim \mathcal{N}(0, \sigma^2 \mathbf{z}_t)\) represents Gaussian noise, \(\mathcal{M}\) is a masking operator that sets a fraction of elements to zero, and \(\mathcal{S}\) is a shuffling operator that permutes the order.

%% file: sec/5_experiments.tex
\begin{figure}
    \centering
    
    \includegraphics[width=\linewidth]{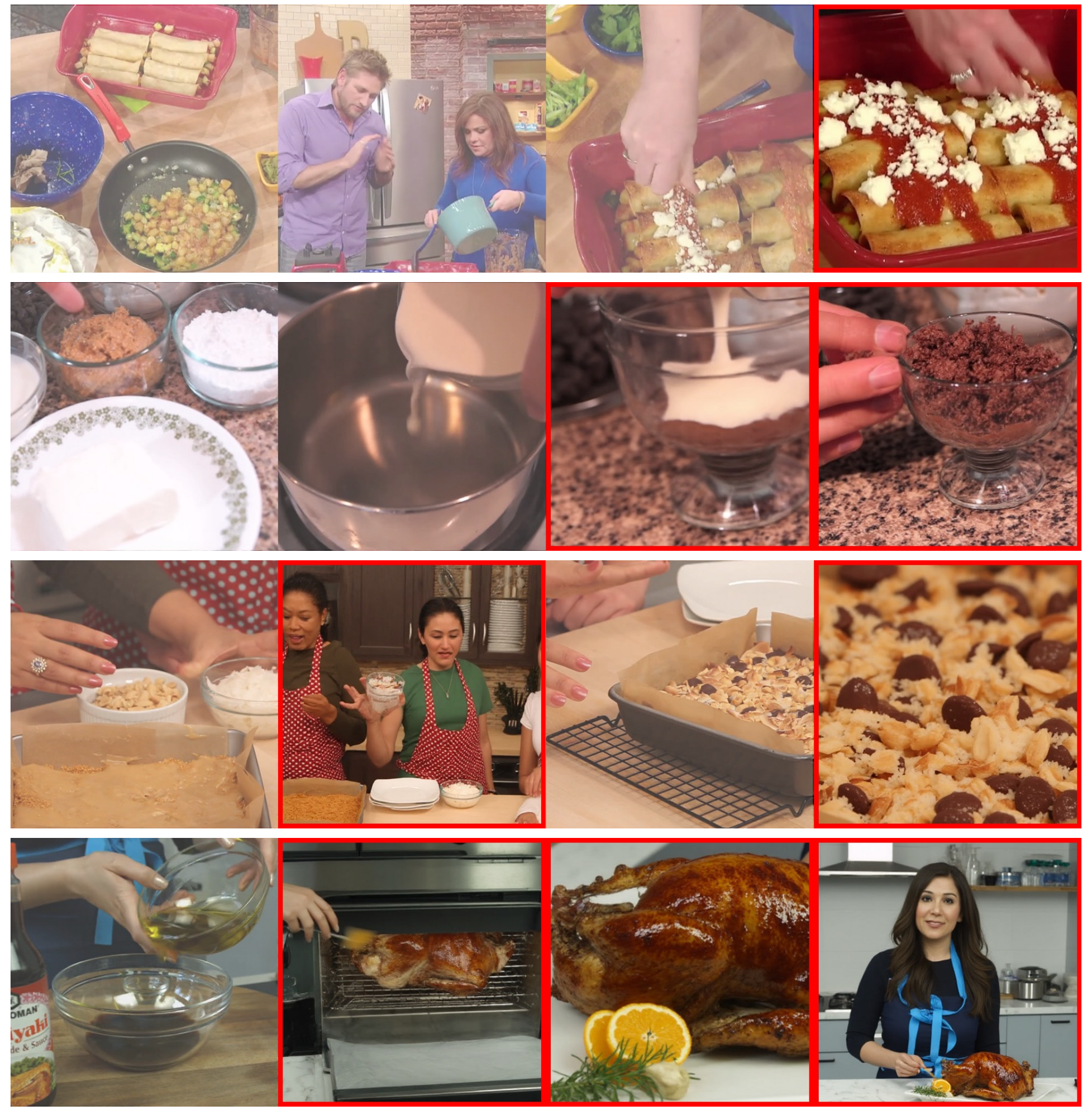}
    \vspace{-3mm}
    \caption{\textbf{Rolling Context Conditioned Render.} We integrate tiled global captions, predicted visual embeddings, and a rolling context of previous keyframes to render new keyframes throughout the narrative. By combining semantic conditioning from textual captions and CLIP embeddings with detailed information from VAE embeddings, the diffusion transformer maintains consistency in visual details such as clothing, food details, and character identities. Generated frames are highlighted with red edges.}
    \vspace{-4mm}
    \label{fig:rolling_context}
\end{figure}

\begin{figure*}[ht]
    \centering
    \includegraphics[width=\linewidth]{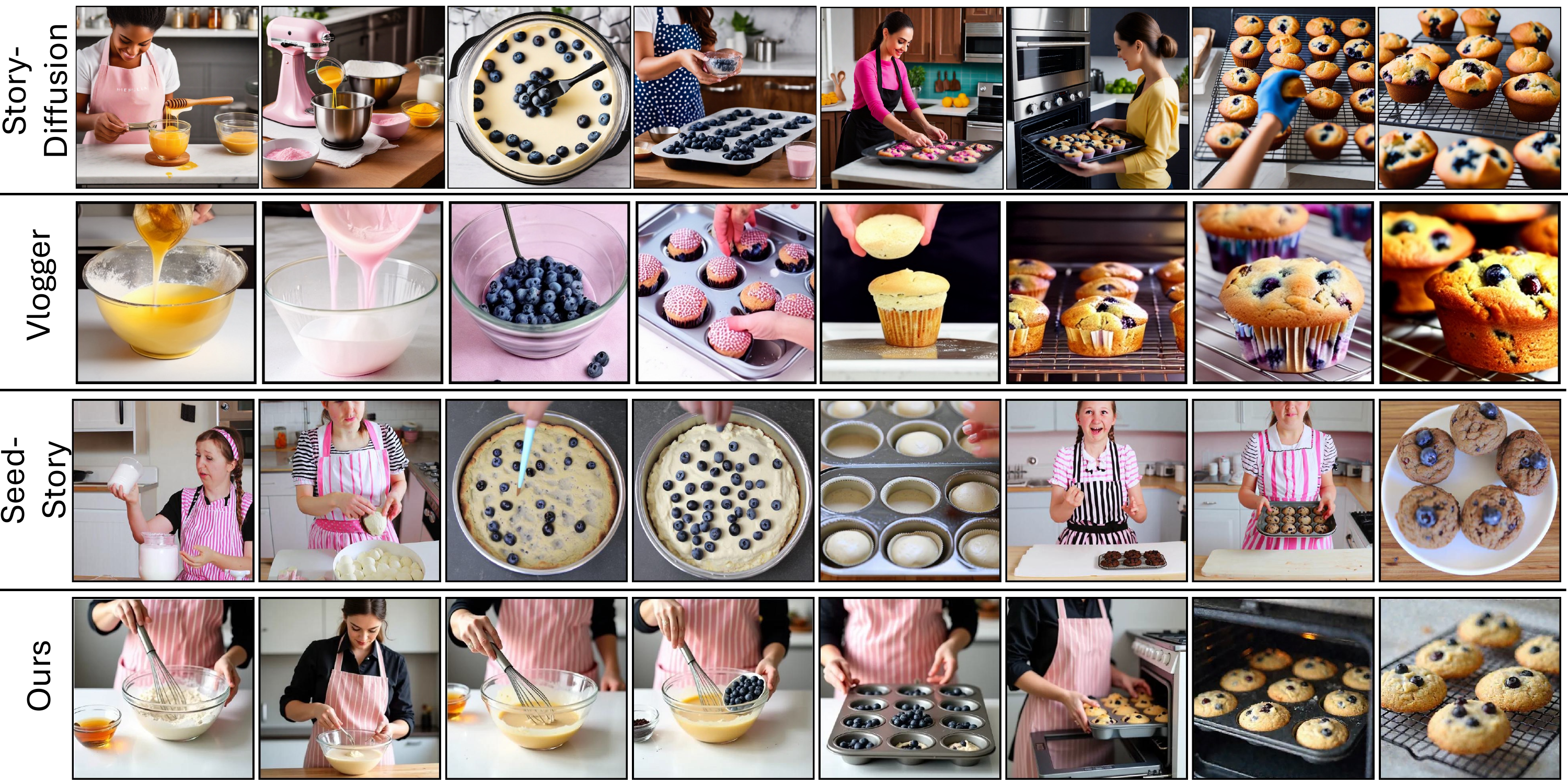}
    \vspace{-4mm}
    \caption{\textbf{Quality comparison on long narrative generation}. Here is a case with a narrative topic of \textit{``Step-by-step guide to cooking blueberry muffins"}. Our interleaved director sequentially generates "actions," "captions," and image embeddings to construct a narrative on how to cook the dish step by step and then render keyframes. Our method shows state-of-the-art visual quality with superior consistency.}
    \vspace{-2mm}
    \label{fig:quality_evaluation_main}
\end{figure*}

\section{Experiments}

\subsection{Experimental Setup}

\noindent\textbf{Models.} We initialize the auto-regressive model with~\cite{ge2024seedx}, a pretrained 7B multi-modal LLM. We initialize the context conditioned render model with FLUX.1 Fill model~\cite{flux2024}.
For video generation, we employ a video generation model which has been pre-trained on large-scale video-text pairs and could accept both text and visual conditions.

\noindent\textbf{Data.} We use a total of $\sim$32K narrative videos for model training and another $\sim$1K videos for validation. All the videos are resized to 448 (short-side) and then center-cropped with 448$\times$448 resolution.

\noindent\textbf{Training \& Evaluation.} We train the interleaved auto-regressive director model for 5,000 steps by default. Training loss is a combination of cosine similarity loss and MSE loss for visual tokens and CrossEntropy loss for language tokens. For rolling context conditioned render, we use the flow matching loss following FLUX~\cite{flux2024}. For visual-conditioned video generation, we use the diffusion loss following DiT~\cite{peebles2023scalable} and Stable Diffusion 3~\cite{esser2024sd3}. Narrative generation is mostly evaluated on the Youcook2 validation set because of the high-quality of action annotations and the Howto100M validation set is mostly used for data quality evaluation and I2V generation. 
Please refer to the appendix for implementation details.

\noindent\textbf{Evaluation Metrics.}
The common metrics CLIP score~\cite{hessel2021clipscore} and FVD~\cite{unterthiner2018towards} are used to assess overall video quality, while the FID~\cite{heusel2017gans} score evaluates the quality of the generated keyframes. Additionally, when comparing to state-of-the-art baselines, human evaluation is used to assess generation aesthetics, realism, visual consistency across video clips, and the narrative score which reflects the coherence of the generated cooking steps, and if the cooking process has been successfully completed.

\vspace{-1mm}
\subsection{Rolling Context Conditioning}
As detailed in \Cref{sec:rolling_context_diffusion}, we leverage the in-context conditioning capabilities of the transformer architecture and adopt a rolling context conditioning strategy to enable DiT to render keyframes with superior visual consistency, while adhering to the extended narrative semantics produced by the interleaved auto-regressive director model. As shown in \Cref{fig:rolling_context}, our keyframe renderer preserves fine visual details and exhibits high visual quality and aesthetics with the help of large-scale pretraining~\cite{flux2024}. The reason is that the in-context conditioned VAE features could preserve visual details and the semantics are preserved through the auto-regressive model. Notably, the rolling context conditioning approach allows the renderer to strike a flexible balance between generation efficiency and visual consistency by dynamically adjusting the number of frames generated in each forward pass (\ie, a dynamic number of frames).

\subsection{Visual-Conditioned Video Generation}
As detailed in \Cref{sec:visual_condition_video_gen}, we fine-tune the model to be directly conditioned on the visual latents and  generated by our interleaved director and keyframes generated by rolling-context renderer. \Cref{tab:visual_conditions} compares the keyframe-conditioned approach with our visual embedding-conditioned strategy. Our method improves CLIP-T~\cite{hessel2021clipscore} scores on both validation sets—from 25.9 to 26.4 on YouCook2 and from 26.6 to 27.3 on HowTo100M. Additionally, FVD scores decrease, indicating better video quality (557.7 vs. 512.6 on YouCook2, 541.1 vs. 520.7 on HowTo100M). Videos conditioned on visual embeddings demonstrate higher semantic alignment and improved generation quality.  We also provide qualitative samples on the demo page and in the appendix.

\begin{table}[t!]
    \centering
    \tablestyle{7pt}{1.0}  %
    \begin{tabular}{c |c c c c}
        \multirow{2}{*}{\textbf{Visual Condition}} & \multicolumn{2}{c}{\textbf{YouCook2}} & \multicolumn{2}{c}{\textbf{HowTo100M}} \\
        & CLIP-T~$\uparrow$ & FVD~$\downarrow$ & CLIP-T~$\uparrow$ & FVD~$\downarrow$ \\ \shline
        Keyframe & 25.9 & 557.7 & 26.6 & 541.1 \\
        Embedding (w/o Reg.) & 25.5 & 590.3 & 26.4 & 554.3 \\
        Embedding (w/ Reg.) & \textbf{26.4} & \textbf{512.6} & \textbf{27.3} & \textbf{520.7} \\
    \end{tabular}
    \caption{\textbf{Visual-conditioned Video Generation with Regularization.} Evaluate CLIP-T and FVD scores for video generation conditioned on keyframes versus visual embeddings generated by our interleaved director with and without regularization.}
    \label{tab:visual_conditions}
\end{table}

\subsection{Comparisons of Long Narrative Generation}
\label{sec:exp_interleaved_visual_generation}
As most existing narrative generation methods~\cite{yang2024seed, zhou2024storydiffusion} only support image generation, we compare our model with state-of-the-art methods on the task of long narrative keyframe generation. We provide both quantitative comparisons in (\S\ref{sec:quantitative_comparison}) and qualitative comparisons (\S\ref{sec:qualitative_comparison}).
\begin{table}[t!]
    \centering
    \tablestyle{1pt}{1.3}
    \scriptsize
    \begin{tabular}{c c c | c c | c c c c}
        \multirow{2}{*}{\textbf{Method}} 
        & \multicolumn{2}{c|}{\textbf{Prompting}} 
        & \multicolumn{2}{c|}{\textbf{Gen. Metric}} 
        & \multicolumn{4}{c}{\textbf{Human Evaluation}} \\
        \cline{2-3}\cline{4-5}\cline{6-9}
        & \textbf{Prompt Src.} 
        & \textbf{Cond.} 
        & \textbf{CLIP-T} 
        & \textbf{FID} 
        & \textbf{Aes.}
        & \textbf{Real.} 
        & \textbf{Consist.} 
        & \textbf{Narr.} \\
        \shline        
        \textcolor{gray}{SD-XL}~\cite{sdxl} & \textcolor{gray}{External} & \textcolor{gray}{Text} & \textcolor{gray}{27.1} & \textcolor{gray}{-} & \textcolor{gray}{4.0} & \textcolor{gray}{2.9} & \textcolor{gray}{3.3} & \textcolor{gray}{N/A} \\
    \textcolor{gray}{FLUX.1-s}~\cite{flux2024} & \textcolor{gray}{External} & \textcolor{gray}{Text} & \textcolor{gray}{27.9} & \textcolor{gray}{-} & \textcolor{gray}{4.8} & \textcolor{gray}{3.1} & \textcolor{gray}{3.4} & \textcolor{gray}{N/A} \\
        IC Lora~\cite{lhhuang2024iclora} & External & Text & 27.9 & 34.1 & 4.7 & 4.1 & 4.7 & N/A \\
        StoryDiffusion~\cite{zhou2024storydiffusion} & External & Text & 25.9 & 36.4 & 3.9 & 2.9 & 3.7 & N/A \\ \hline
        Vlogger~\cite{zhuang2024vlogger} & LLM & Text & 25.5 & 45.5 & 4.0 & 2.4 & 3.1 & 3.7 \\
        Seed-Story~\cite{yang2024seed} & Interleaved & V. Emb. & 24.1 & 32.1 & 1.9 & 4.1 & 4.2 & 4.1 \\
        \rowcolor{cyan!15}
       \textbf{ Ours (w.o RCC)} & Interleaved & V. Emb. & 26.1 & \textbf{25.3} & 2.1 & 4.3 & 4.5 & 4.4 \\
        \rowcolor{cyan!15}
        \textbf{Ours (w. RCC)} & Interleaved & T+V. Emb. & \textbf{28.0} & 29.4 & \textbf{4.8} & \textbf{4.5} & \textbf{4.8 }& \textbf{4.6} \\
    \end{tabular}
    \vspace{-2mm}
    \caption{\textbf{Quantitative comparisons with metrics and human evaluation.} Each method is evaluated by both image generation metrics (CLIP-T and FID) and human ratings. Higher values indicate better performance for all human-evaluation metrics (5 tiers, from 1 to 5, higher is better).  SD-XL and FLUX.1-s use narrative captions generated by our model and IC-Lora uses a tiled version. RCC: Rolling Context Conditioning. We use our generated narrative captions for the text-conditioned methods (row 1-5).}
    \vspace{-2mm}
    \label{tab:method_comparison}
\end{table}

\vspace{-3mm}

\subsubsection{Long Narrative Keyframe Generation}
\label{sec:quantitative_comparison} We compare our method with leading narrative keyframe generation approaches, including IC Lora~\cite{lhhuang2024iclora}, StoryDiffusion~\cite{zhou2024storydiffusion}, Vlogger~\cite{zhuang2024vlogger}, and Seed-Story~\cite{yang2024seed}, as well as a language-centric strategy that relies solely on captions (using models such as SD-XL~\cite{sdxl} and FLUX.1-schnell~\cite{flux2024}). Except for IC Lora and Seed-Story, which are fine-tuned on our CookGen dataset for two epochs, all other methods follow their official inference guidelines with the official checkpoints. As shown in \Cref{tab:method_comparison}, our approach achieves the highest generation scores, with a CLIP-Text score of \textbf{28.0} and an FID score of \textbf{25.3}. We also conduct a human evaluation (\Cref{tab:method_comparison}) using a five-tier rating scale, where higher is better. Our method attains top performance in aesthetics (\textbf{4.8} vs. \textbf{4.7}, IC Lora), realism (\textbf{4.5} vs. \textbf{4.1}, Seed-Story), and visual consistency (\textbf{4.8} vs. \textbf{4.7},  IC Lora), as well as the highest narrative score of \textbf{4.6}. These results demonstrate that our method achieves  state-of-the-art performance for long narrative generation.

\vspace{-3mm}

\subsubsection{Qualitative Comparisons}
\label{sec:qualitative_comparison}

In \Cref{fig:quality_evaluation_main}, we compare our method with state-of-the-art long narrative keyframe generation approaches, including StoryDiffusion, Vlogger, and Seed-Story, and observe that our results maintain superior visual quality and consistency. In particular, our keyframes balance realism with appealing aesthetics while preserving character identities and smooth transitions. In contrast, competing methods often exhibit color inconsistencies or lose track of concepts—Vlogger occasionally produces uneven color schemes between frames, StoryDiffusion can introduce visual confusion, and Seed-Story sometimes generates mismatched clothing across different scenes. This comparison aligns with the human evaluation results in \Cref{tab:method_comparison}, demonstrating our method achieves state-of-the-art performance for long narrative visual generation. The generated keyframes can be extended into full video clips with consistent visuals and coherent storytelling.

\subsection{Ablation Studies}
\label{sec:ablation_studies}

In this section, we ablate important designs in VideoAuteur, which improve the interleaved auto-regressive model and the visual-conditioned video generation model for interleaved narrative visual generation.

\noindent\textbf{Latent scale and direction matter.} To determine an effective supervision strategy for visual embeddings, we firstly test the robustness of the latents to pseudo regression errors by rescaling (multiplying by a factor) and adding random Gaussian noise. \Cref{fig:scale_direction} indicates that both scale and direction are critical in latent regression. Notably, rescaling primarily affects object shape while preserving key semantic information (\ie object type and location), whereas adding noise drastically impacts reconstruction quality.
As shown in \Cref{tab:scale_and_direction}, combining MSE loss (for scale ) and cosine similarity (for direction) leads to the best generation quality, improving CLIP-T by 1.5 points and reducing FID by 1.8 points compared to using MSE alone.

\begin{table}[t!]
    \centering
    \tablestyle{3pt}{1.0}
    \begin{tabular}{cc|cccc}
        \multicolumn{2}{c|}{\textbf{Loss Type}} & \multicolumn{2}{c}{\textbf{Training}} & \multicolumn{2}{c}{\textbf{Validation}} \\
        MSE & Cosine & L2 Dist~$\downarrow$  & Cosine Sim.~$\uparrow$ & CLIP-T~$\uparrow$ & FID~$\downarrow$ \\
        \shline
        \cmark & \xmark & \textbf{0.41} & 0.82 & 23.6 & 31.9 \\
         \xmark & \cmark &  1.1 &  0.82 &  24.1 &  32.1 \\
        \cmark & \cmark & \textbf{0.41} & \textbf{0.83} & \textbf{25.1} & \textbf{30.1} \\
    \end{tabular}
    \label{tab:scale_and_direction}
    \vspace{-2mm}
    \caption{\textbf{Both scale and direction matter.} We track the training convergence and evaluate models with the CLIP-T and FID metrics on the validation set. The combination of both MSE loss and Cosine Similarity loss performs best on the validation metrics.}
    \vspace{-2mm}
\end{table}

\noindent\textbf{From ``Actions" to ``Visual States".} We also explore how different regression tasks influence the director's capability in narrative visual generation. Specifically, we compare various reasoning settings for the interleaved director, examining transitions from sequential actions to language states, and ultimately to visual embeddings. As shown in \Cref{tab:regression_tasks}, a chain of reasoning that progresses from actions to language states and then to visual states proves effective for long narrative visual generation. This approach enhances both training convergence, achieving a lower L2 distance (\textbf{0.41} vs. \textbf{0.43}), and generation quality, reflected in a superior FID score of 25.3 (an improvement of \textbf{+0.8}).

\begin{table}[t!]
    \centering
    \tablestyle{2pt}{1.2}  %
    \scriptsize
    \begin{tabular}{cc|cc cc}
        & & \multicolumn{2}{c}{\textbf{Training}} & \multicolumn{2}{c}{\textbf{Validation}} \\ 
        \multicolumn{2}{c|}{\textbf{Regression Task}} & L2 Dist~$\downarrow$  & Cosine Sim.~$\uparrow$ & CLIP-T~$\uparrow$ & FID~$\downarrow$  \\ \shline
        Action $\rightarrow$ Vis. Embed. & &  0.43 &  0.82 &  22.7 &  27.9 \\
        Caption $\rightarrow$ Vis. Embed. & & 0.41 & 0.82 & 25.7 & 26.1  \\
        Action $\rightarrow$ Caption $\rightarrow$ Vis. Embed. & & \textbf{0.41} & \textbf{0.83} & \textbf{26.1} & \textbf{25.3}  \\
    \end{tabular}
    \vspace{-2mm}
    \caption{\textbf{From ``Actions" to ``Visual States".} We report the L2 distance and cosine similarity scores for tracking the training convergence and evaluate the generation images with CLIP score and FID score. Models are trained and evaluated on the collected Howto100M subset. SEED-X latent is used for visual regression.}
    \vspace{-2mm}
    \label{tab:regression_tasks}
\end{table}

\noindent\textbf{Learn from noisy visual conditions.} \Cref{tab:ablation_visual_conditioned} presents an ablation study examining the effect of robustness regularization on the visual-conditioned video generation model. We evaluate the generated videos using CLIP-T and FVD. The progressively improved results from 26.4 to 27.3 on CLIP-T and 554.3 to 520.7 on FVD demonstrate the effectiveness of our regularization strategy, which combines random masking, Gaussian noise, and shuffling.

\begin{table}[]
    \small
    \tablestyle{4pt}{1.1}  %
    \begin{tabular}{lcc}
        \textbf{Regularization Setting} & \textbf{CLIP-T}~$\uparrow$ & \textbf{FVD}~$\downarrow$ \\ \shline
        Naive Baseline & 26.4 & 554.3 \\ \hline 
        +Random Masking & 26.9 & 539.7 \\
        +Random Gaussian. Noise & 27.2 & 522.1 \\
        +Random Shuffling & \textbf{27.3} & \textbf{520.7} \\
    \end{tabular}
    \vspace{-4mm}
\caption{\textbf{Learn from Noisy Visual Conditions.} Our training regularization strategy enhances the robustness of the visual-conditioned video generation model. Specifically, we apply random masking and shuffling at a rate of 25\%, and introduce Gaussian noise with 0.5 std of the embeddings of two thousand samples.}
    \label{tab:ablation_visual_conditioned}
\end{table}

%% file: sec/7_conclusion.tex
\section{Conclusion}
\label{sec:conclusion}
In this paper, we tackle the challenges of generating long-form narrative videos and empirically evaluate its efficacy in the cooking domain. We curate and annotate a large-scale cooking video dataset, capturing clear and high-quality narratives essential for training and evaluation. Our proposed two-stage auto-regressive pipeline, which includes a long narrative director, a rolling context conditioned keyframe renderer and a visual-conditioned video generation model, demonstrates promising improvements in semantic and visual consistency in generated long narrative videos with an unified pipeline. Through experiments on our dataset, we observe enhancements in spatial and temporal coherence across video sequences. We hope our work can facilitate further  research in long narrative video generation.

%% file: sec/X_suppl.tex
\clearpage

\onecolumn
\Large
\begin{center}
    Appendix
\end{center}
\appendix
\normalsize
This appendix provides comprehensive supplementary materials to support our study. Below are brief descriptions of all the sections covered in the appendix. Please visit our project page for more visualization.
\vspace{4mm}
\begin{itemize}
    \item \Cref{sec:appendix_data_with_annotations}: \textbf{Data Examples with Annotations}
    \begin{itemize}
        \item Presents data examples from our CookGen dataset.
        \item Showcases annotated ``actions'' and ``captions'' that provide detailed multimodal information of cooking processes.
    \end{itemize}
    \item \Cref{sec:appendix_additional_data_stats}: \textbf{Additional Data Statistics}
    \begin{itemize}
        \item Offers distributions of video lengths, clip lengths, and textual annotations.
        \item Demonstrates the dataset's richness and suitability for long narrative video generation.
    \end{itemize}
    \item \Cref{sec:appendix_data_eval_details}: \textbf{Data Evaluation Details}
    \begin{itemize}
        \item Details our data evaluation process.
        \item Includes inverse video generation results, the prompts used for video captioning, GPT-4o evaluations, and human evaluation results.
    \end{itemize}
    \item \Cref{sec:appendix_implementation_details}: \textbf{Implementation Details}
    \begin{itemize}
        \item Outlines the implementation details of our models.
        \item Provides key hyperparameters and training \& inference configurations.
    \end{itemize}
    \item \Cref{sec:appendix_action_caption_matching_pseudo_code}: \textbf{Action-Caption Matching Pseudo Code}
    \begin{itemize}
        \item Includes the pseudo code for our action-caption matching algorithm.
        \item Essential for aligning video clips with their corresponding annotations.
    \end{itemize}
    \item \Cref{sec:clip_beats_vae}: \textbf{CLIP beats VAE for interleaved generation}
    \begin{itemize}
        \item Introduces three autoencoders (EMU-2, SEED-X, SDXL-VAE) and compares reconstruction vs. generation performance.
        \item Demonstrates that CLIP-diffusion embeddings (EMU-2, SEED-X) outperform SDXL-VAE in language-driven visual generation due to better vision-language alignment.
    \end{itemize}
    \item \Cref{sec:appendix_generated_video_examples}: \textbf{Generated Video Examples}
    \begin{itemize}
        \item Showcases generated video examples.
        \item Illustrates the effectiveness of our pipeline in producing long narrative videos for cooking recipes like ``Fried Chicken'' and ``Shish Kabob.''
    \end{itemize}
    \item \Cref{sec:appendix_limitations}: \textbf{Limitations}
    \begin{itemize}
        \item Discusses the limitations of our approach.
        \item Includes issues with noisy ``actions'' from automatic speech recognition and potential failure cases in video generation.
    \end{itemize}
\end{itemize}

\clearpage
\section{Data Examples with Annotations}
\label{sec:appendix_data_with_annotations}
\Cref{fig:appendix_data_example,fig:appendix_data_example_jello} shows two data examples from our CookGen dataset, annotated with high-quality descriptions that provide detailed multi-modal information of cooking processes. The examples clearly show structured annotations of key actions and corresponding visual descriptions, making the dataset ideal for generating long narrative videos. 

\begin{figure*}[h]
    \centering
    \begin{subfigure}[t]{0.48\textwidth}
        \includegraphics[width=\textwidth]{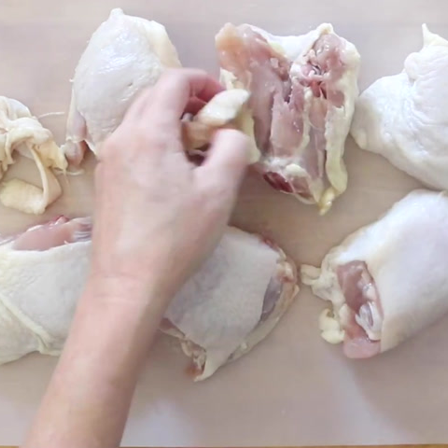}
        \caption{\textbf{Action:} Elise works with chicken thighs, advises to trim excess skin and fat \\
        \textbf{Caption:} A person is preparing chicken on a wooden cutting board. He uses a pair of black-handled scissors to cut through the chicken pieces, which are spread out on a clear cutting mat.}
    \end{subfigure}
    \hfill
    \begin{subfigure}[t]{0.48\textwidth}
        \includegraphics[width=\textwidth]{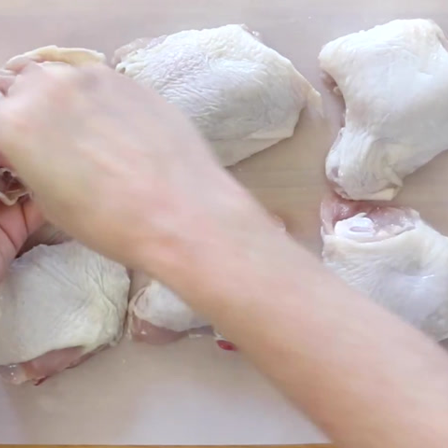}
        \caption{\textbf{Action:} She offers alternatives with chicken breast bone-in skin-on or chicken drumsticks \\
        \textbf{Caption:} A person with light skin is preparing raw chicken pieces on a wooden surface. He places several pieces of chicken on a white cutting board.}
    \end{subfigure}
    \vspace{1em} %
    \begin{subfigure}[t]{0.48\textwidth}
        \includegraphics[width=\textwidth]{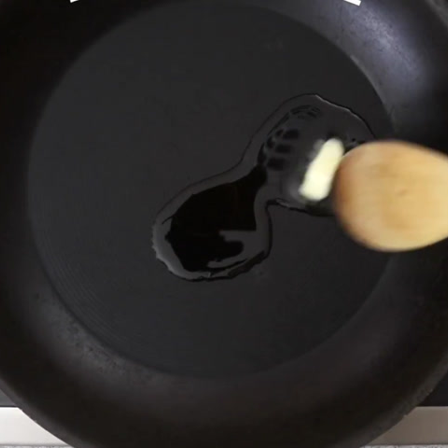}
        \caption{\textbf{Action:} Elise heats up a large skillet with two teaspoons of olive oil and a teaspoon of butter \\
        \textbf{Caption:} A person is seen in a kitchen setting, holding a wooden spoon. He places a small piece of butter into a black frying pan on a gas stove.}
    \end{subfigure}
    \hfill
    \begin{subfigure}[t]{0.48\textwidth}
        \includegraphics[width=\textwidth]{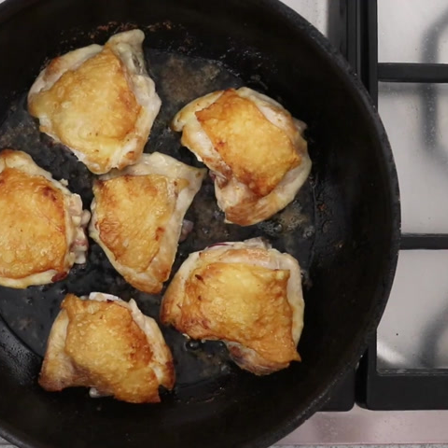}
        \caption{\textbf{Action:} Turn over the chicken pieces and cook for another 4 minutes Remove the chicken from the pan but keep the browned pieces in the pan \\
        \textbf{Caption:} Golden-brown chicken pieces are sizzling in a black frying pan on a gas stove.}
    \end{subfigure}
    \vspace{-1cm}
\end{figure*}

\begin{figure*}[ht]
    \ContinuedFloat 
    \centering
    \begin{subfigure}[t]{0.48\textwidth}
        \includegraphics[width=\textwidth]{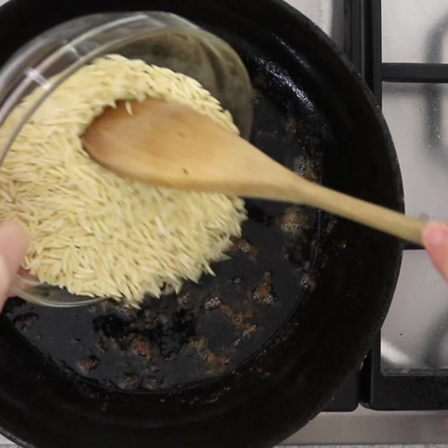}
        \caption{\textbf{Action:} Use the remaining oil in the pan to brown the orzo Cook the orzo like a traditional rice pilaf, using the same method as before \\
        \textbf{Caption:} A person is cooking rice in a black frying pan on a gas stove. He pours the rice from a glass bowl into the pan, then uses a wooden spatula to spread and stir the rice.}
    \end{subfigure}
    \hfill
    \begin{subfigure}[t]{0.48\textwidth}
        \includegraphics[width=\textwidth]{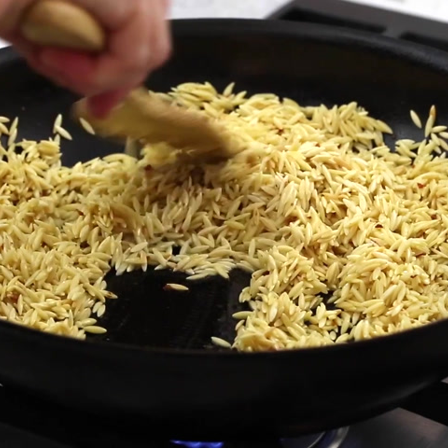}
        \caption{\textbf{Action:} Add 2 cups of gordo's to a hot pan \\
        \textbf{Caption:} A person wearing a blue shirt is cooking rice in a black frying pan on a stovetop. Using a wooden spatula, he stirs the rice, ensuring it is evenly cooked.}
    \end{subfigure}
    \vspace{1em} %
    \begin{subfigure}[t]{0.48\textwidth}
        \includegraphics[width=\textwidth]{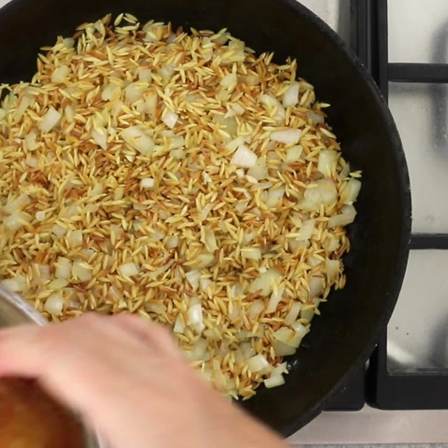}
        \caption{\textbf{Action:} Combine the mixture with the orzo and cook for a few minutes until the sauce thickens \\
        \textbf{Caption:} A woman is cooking on a stovetop, adding pieces of breaded chicken to a pan filled with chopped onions and rice.}
    \end{subfigure}
    \hfill
    \begin{subfigure}[t]{0.48\textwidth}
        \includegraphics[width=\textwidth]{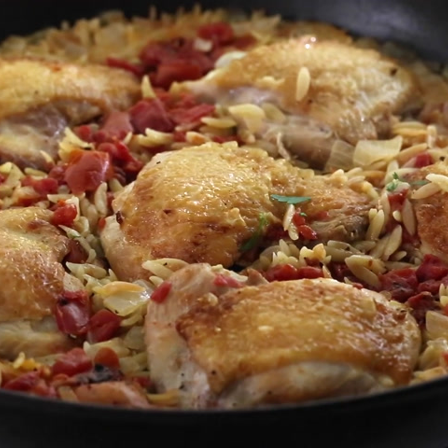}
        \caption{\textbf{Action:} Stock is cooked until orzo has fully absorbed liquid and chicken is cooked through, about 10-12 minutes Dish is removed from heat and left to sit for five minutes Dish is sprinkled with unspecified seasoning \\
        \textbf{Caption:} A delicious dish of roasted chicken pieces is presented in a black skillet, surrounded by a colorful mix of diced vegetables and grains.}
    \end{subfigure}
    \caption{\textbf{Data examples with annotated ``actions" and ``captions"}. A video of cooking recipe of \textit{``One Pot Chicken and Orzo"}.}
    \label{fig:appendix_data_example}
\end{figure*}

\begin{figure*}[h]
    \centering
    \begin{subfigure}[t]{0.48\textwidth}
        \includegraphics[width=\textwidth]{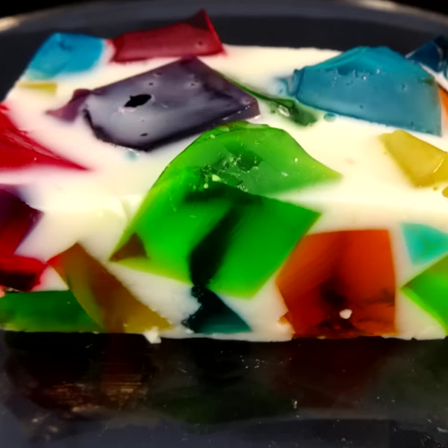}
        \caption{\textbf{Action:} Hi everyone, this one's called rainbow broken glass jello \\
        \textbf{Caption:} A colorful, multi-layered dessert is displayed on a black surface. The dessert features vibrant red, green, blue, and purple segments, arranged in a geometric pattern.}
    \end{subfigure}
    \hfill
    \begin{subfigure}[t]{0.48\textwidth}
        \includegraphics[width=\textwidth]{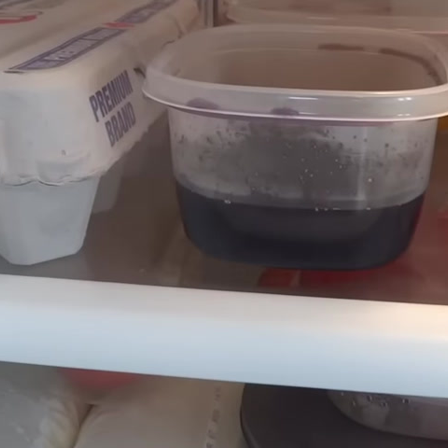}
        \caption{\textbf{Action:} Now normally when you make jello you use two cups of boiling water, but in this case we're only using one cup because we want the jello to be extra firm \\
        \textbf{Caption:} The video shows the interior of a refrigerator, focusing on the door shelf. The containers are filled with dark, blue, orange, and red liquids.}
    \end{subfigure}

    \vspace{1em} %

    \begin{subfigure}[t]{0.48\textwidth}
        \includegraphics[width=\textwidth]{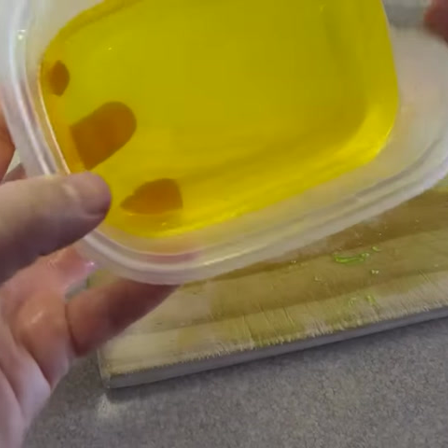}
        \caption{\textbf{Action:} I find the easiest way to do this is to put the small container into a larger container of hot water \\
        \textbf{Caption:} A person with light skin is holding a clear plastic container filled with a yellow liquid, inspecting its contents.}
    \end{subfigure}
    \hfill
    \begin{subfigure}[t]{0.48\textwidth}
        \includegraphics[width=\textwidth]{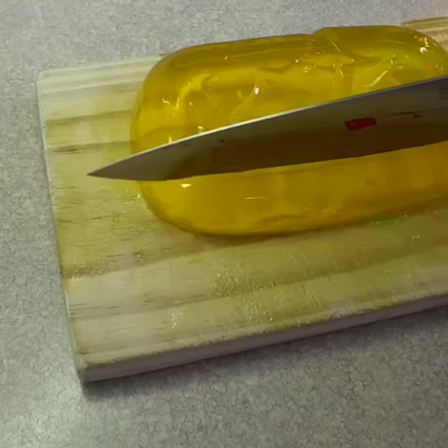}
        \caption{\textbf{Action:} Loosen the edges of the Jello piece Slide the Jello piece out and cut it into cubes Cut the Jello cubes into half-inch pieces \\
        \textbf{Caption:} A person is slicing a block of yellow gelatin on a wooden cutting board, cutting it into uniform strips.}
    \end{subfigure}
\end{figure*}

\begin{figure*}[ht]
    \ContinuedFloat
    \centering
    \begin{subfigure}[t]{0.48\textwidth}
        \includegraphics[width=\textwidth]{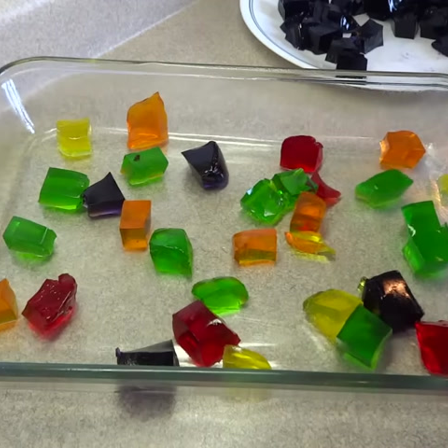}
        \caption{\textbf{Action:} Spread out the different colored Jello pieces in a 9 by 13 inch baking dish \\
        \textbf{Caption:} A person is arranging colorful gelatin cubes in a glass baking dish, adjusting the placement of green, orange, purple, and black cubes.}
    \end{subfigure}
    \hfill
    \begin{subfigure}[t]{0.48\textwidth}
        \includegraphics[width=\textwidth]{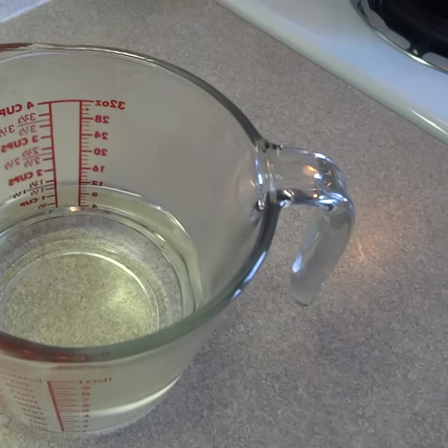}
        \caption{\textbf{Action:} Make a separate gelatin mixture by boiling two cups of water and adding two envelopes of gelatin \\
        \textbf{Caption:} A clear glass measuring cup is placed on a countertop, containing water. A person pours a white powder into it.}
    \end{subfigure}

    \vspace{1em} %

    \begin{subfigure}[t]{0.48\textwidth}
        \includegraphics[width=\textwidth]{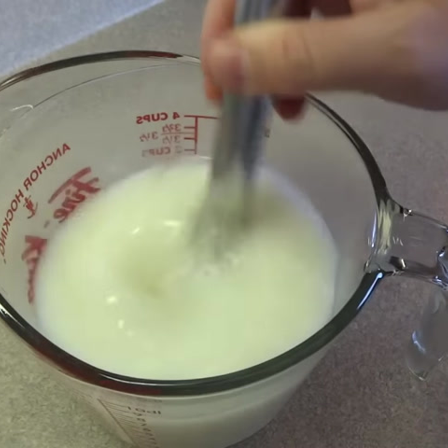}
        \caption{\textbf{Action:} Stir the sweetened condensed milk into the gelatin and water mixture \\
        \textbf{Caption:} A person is vigorously whisking a creamy mixture in a clear glass measuring cup.}
    \end{subfigure}
    \hfill
    \begin{subfigure}[t]{0.48\textwidth}
        \includegraphics[width=\textwidth]{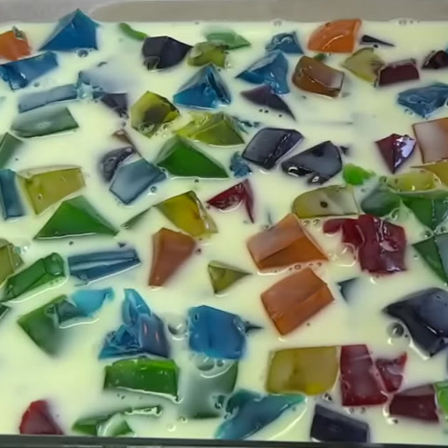}
        \caption{\textbf{Action:} Let it set for several hours, then cut it into squares and serve \\
        \textbf{Caption:} A glass baking dish is filled with a creamy white liquid, topped with colorful, triangular-shaped glass pieces.}
    \end{subfigure}

    \caption{\textbf{Data examples with annotated ``actions" and ``captions"}. A video of preparing \textit{``Rainbow Broken Glass Jello"}.}
    \label{fig:appendix_data_example_jello}
\end{figure*}

\clearpage
\section{Additional Data Statistics}
\label{sec:appendix_additional_data_stats}

\begin{figure*}[ht!]
    \centering
    \includegraphics[width=\linewidth]{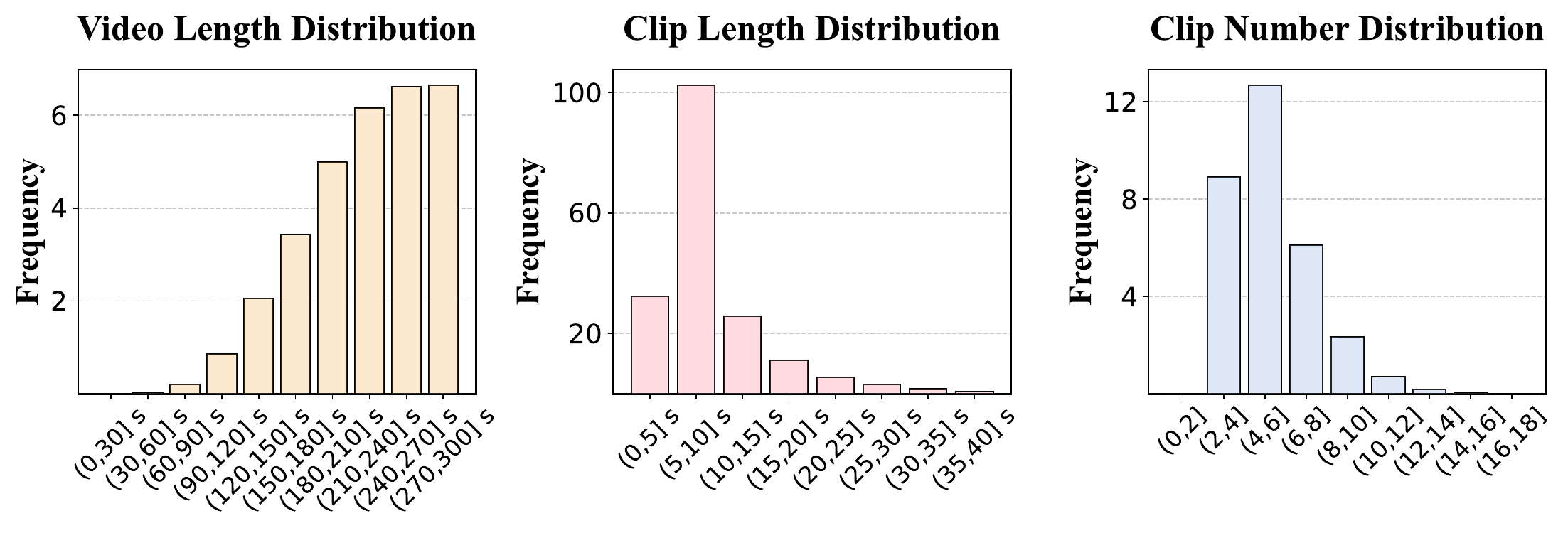}
    \vspace{-3mm}
    \caption{\textbf{Statistics on the video data.} We do statistics on the video lengths of the collected whole videos, the clip lengths of the scene-cut video clips, and the number of clips selected for each video.}
    \vspace{-4mm}
    \label{fig:stats_clip_distribution}
\end{figure*}

\begin{figure*}[ht!]
    \centering
    \includegraphics[width=0.95\linewidth]{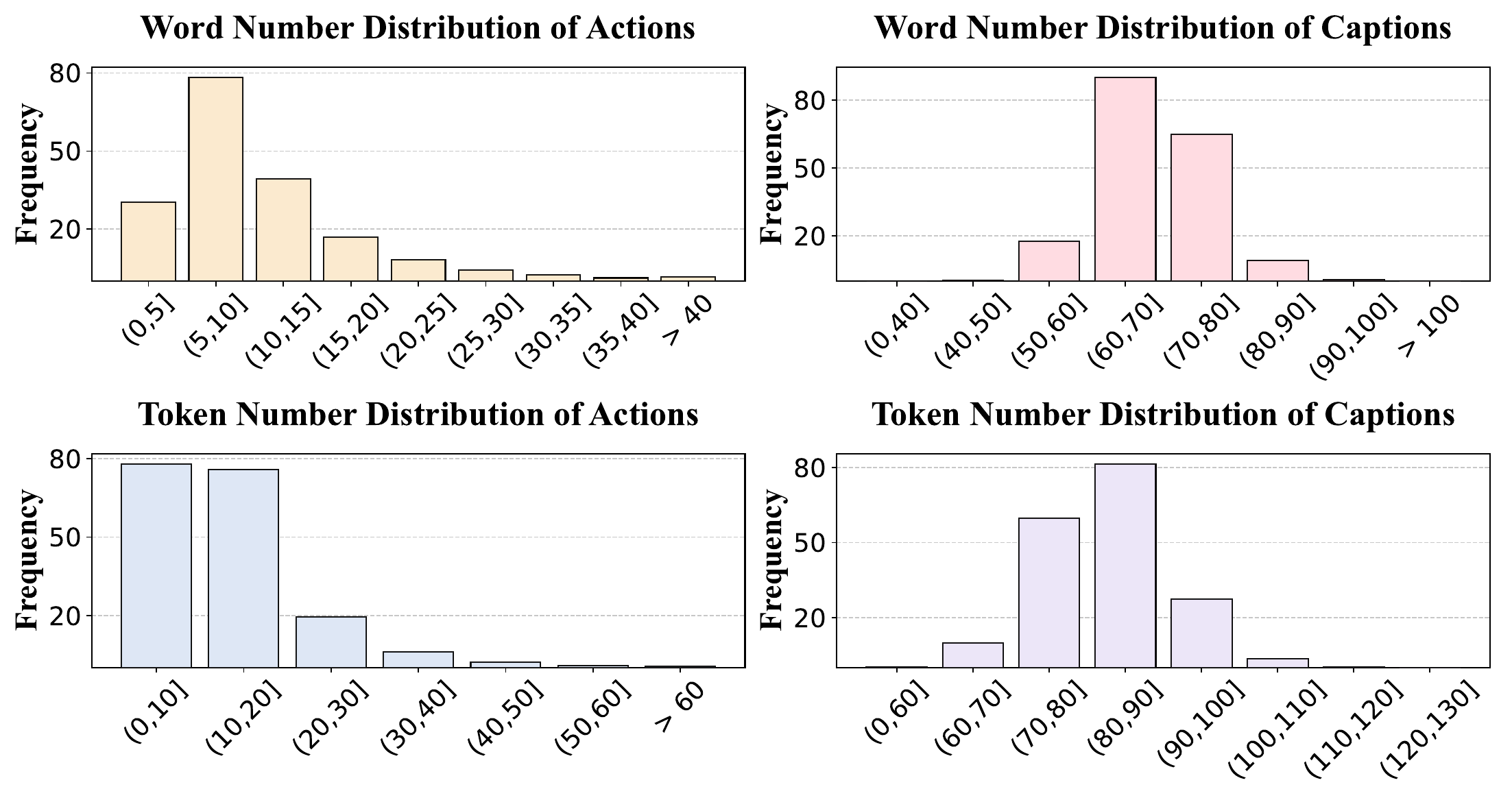}
    \vspace{-3mm}
    \caption{\textbf{Statistics on the text annotations.} We do statistics on the number of words and tokens (Llama~\cite{touvron2023llama2} tokenized) of annotated ``actions" and ``captions," respectively.}
    \label{fig:stats_word_number_distribution}
    \vspace{-3mm}
\end{figure*}

The statistics in \Cref{fig:stats_clip_distribution} and \Cref{fig:stats_word_number_distribution} demonstrate the high quality and suitability of our dataset for long narrative video generation. \Cref{fig:stats_clip_distribution} reveals that the video lengths range broadly, with most videos falling between \textbf{30} and \textbf{150} seconds. Clip lengths are primarily distributed between \textbf{5} and \textbf{30} seconds, ensuring manageable segments for modeling. Additionally, the majority of videos contain \textbf{4} to \textbf{12} clips, providing a balanced structure for narrative flow. \Cref{fig:stats_word_number_distribution} shows that the word counts for "actions" predominantly range from \textbf{10} to \textbf{25}, while "captions" range from \textbf{40} to \textbf{70}. Token distributions further highlight their richness, with "actions" having \textbf{20} to \textbf{60} tokens and "captions" extending up to \textbf{120} tokens. These detailed annotations ensure well-aligned and contextually rich representations of the video content.

Overall, the dataset's design ensures coherent sequences of actions and captions with reasonable clip and video lengths, making it well-suited for generating high-quality, long-form narrative videos.

\section{Annotation Quality Reverification Details}
\label{sec:appendix_data_eval_details}

High-quality captions are essential for narrative visual generation. To verify the quality of our annotations, we build an evaluation pipline of inverse generation (\S\ref{sec:inverse_video_gen}) and visual understanding through VLM experts (\S\ref{sec:vlm_consistency}).

\subsection{Inverse Video Generation} 
\label{sec:inverse_video_gen}
\begin{figure}[h!]
\begin{minipage}{0.48\linewidth}
    \centering
    \includegraphics[width=\linewidth]{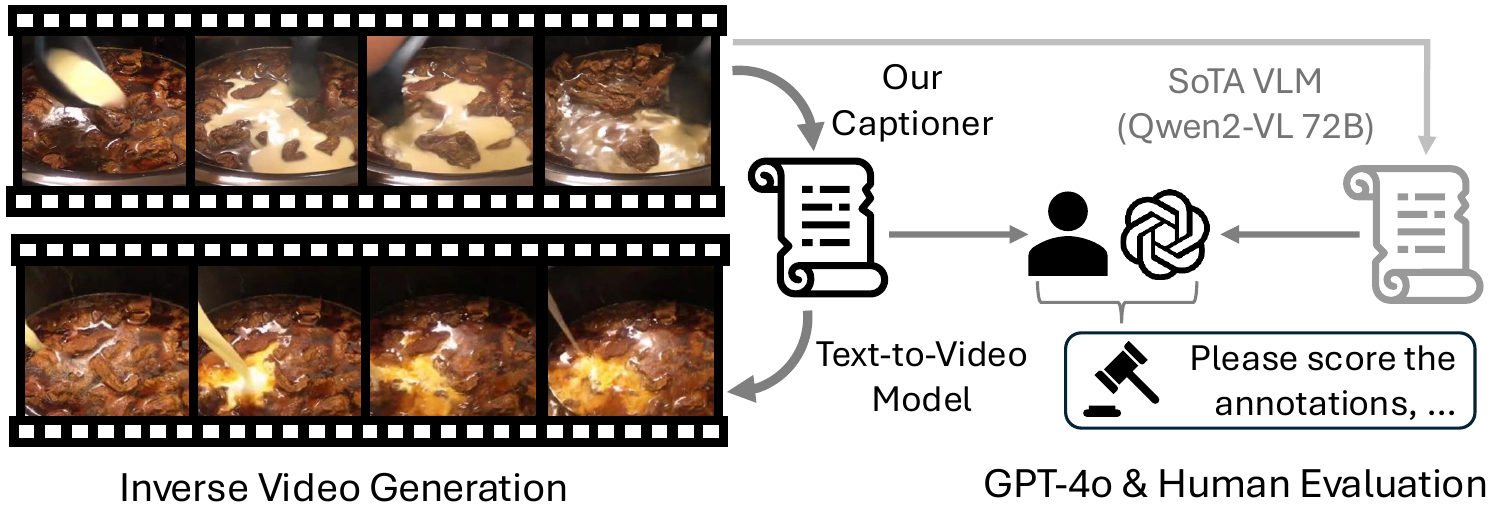}
    \caption{\textbf{Annotation quality evaluation pipeline}. We verify our annotation quality through a pipeline of two major aspects: 1) inverse video generation 2) GPT-4o and human evaluation.  }
    \label{fig:annotation_eval}
\end{minipage}
\hfill
\begin{minipage}{0.48\linewidth}
    \centering
    \tablestyle{10pt}{1.0}
    \begin{tabular}{c|c|c}
         \textbf{Validation Set} & \textbf{w/. GT keyframe} & \textbf{W/o. GT keyframe} \\ \shline
        \# Clips  & FVD & FVD \\ 
        5504 & \textbf{116.3} & 561.1 \\ 
    \end{tabular}
    \caption{\textbf{High-quality captions enable inverse video generation.} We utilize the annotated captions and actions to inversely generate video clips using a pretrained Text-to-Video diffusion model. Higher reconstruction fidelity (i.e., similarity to the original videos) indicates superior captions and actions. Our inversely generated videos achieve very low FVD scores compared to the original videos, highlighting the high quality of our annotations.}
    \label{tab:inverse_gen}
\end{minipage}
\end{figure}

This evaluation is motivated by the understanding that high quality captions, when combined with ground truth keyframes, more effectively reconstruct the original videos. We evaluate the dataset’s ability to reconstruct original videos using the annotated captions, with and without conditioning with ground truth keyframes. For this evaluation, we assess the validation set ($\sim$5,000 video clips). We measure reconstruction quality using FVD~\cite{unterthiner2018towards}. The results, shown in Table~\ref{tab:inverse_gen}, indicate that our captions capture sufficient semantic information, enabling effective representation of the original videos. When generating with ground-truth keyframes, the video quality is very high and closely aligned with the original videos, as shown by the low FVD score (116.3). Without keyframes, the captions alone still provide reasonable alignment. Examples of reconstructed videos are included in the supplementary materials.

\begin{table}[h!]
    \centering
    \large
    \tablestyle{4pt}{1.0}  %
    \begin{tabular}{c|cc|cc}
         & \multicolumn{2}{c|}{\textbf{GPT-4o Evaluation}} & \multicolumn{2}{c}{\textbf{Human Evaluation}} \\ \shline
       \multirow{2}{*}{Score (0-100)}& Qwen2-VL-72B & \textbf{Ours} & Qwen2-VL-72B & \textbf{Ours} \\
         & \textbf{98.0} & 95.2 & 79.3 & \textbf{82.0} \\ 
    \end{tabular}
    \caption{\textbf{Caption Quality Evaluation}. We compare the caption quality between our captioner and the Qwen2-VL-72B model by both GPT-4o and human annotators. Our model achieves competitive results despite a much smaller model size.}
     \vspace{-4mm}
    \label{tab:caption-quality}
\end{table}

\subsection{Semantic Consistency across VLM Experts}
\label{sec:vlm_consistency}

\noindent\textbf{GPT-4o \& human evaluation.}
We evaluate the quality of our captions using both GPT-4o and six human annotators, in which we ask humans and GPT-4o to rate our dataset provided captions according to two criteria: the coverage of video elements and the absence of hallucinations in the caption. Following~\cite{zhang2024direct}, hallucination refers to the model generating content absent or unsupported by the video, such as incorrect details about objects, actions, or counts.

To demonstrate the quality, we compare our captions with those generated by a state-of-the-art open-source VLM (Qwen2-VL-72B). As shown in Table~\ref{tab:caption-quality}, our dataset's captions receive a decent score of 95.2 out of 100, showing slightly better alignment with rigorous human evaluation than the Qwen2-VL-72B model. Results from both human evaluators and GPT-4 assessments indicate that the dataset contains high-quality captions.

\subsection{Inverse Video Generation Results}

\begin{figure*}[!h]
    \centering
    \begin{subfigure}[b]{0.48\linewidth}
        \centering
        \includegraphics[width=\linewidth]{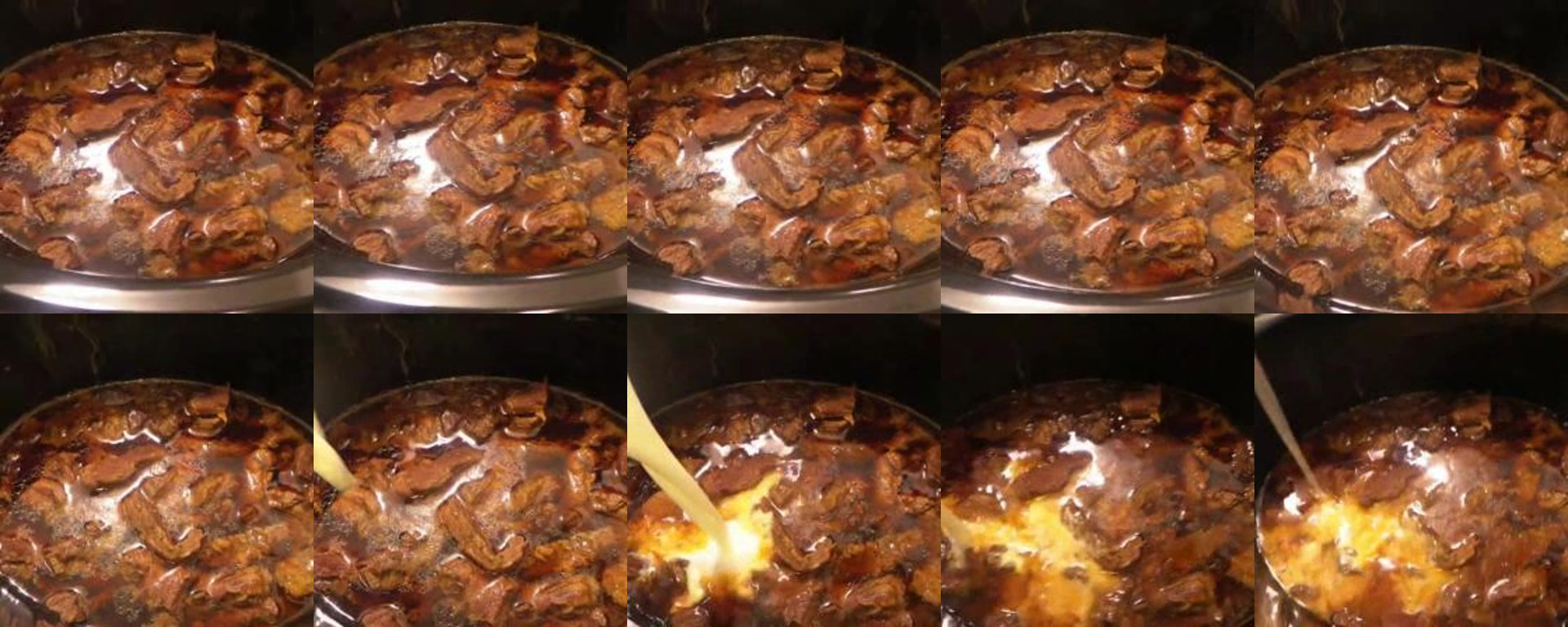}
    \end{subfigure}
    \hfill
    \begin{subfigure}[b]{0.48\linewidth}
        \centering
        \includegraphics[width=\linewidth]{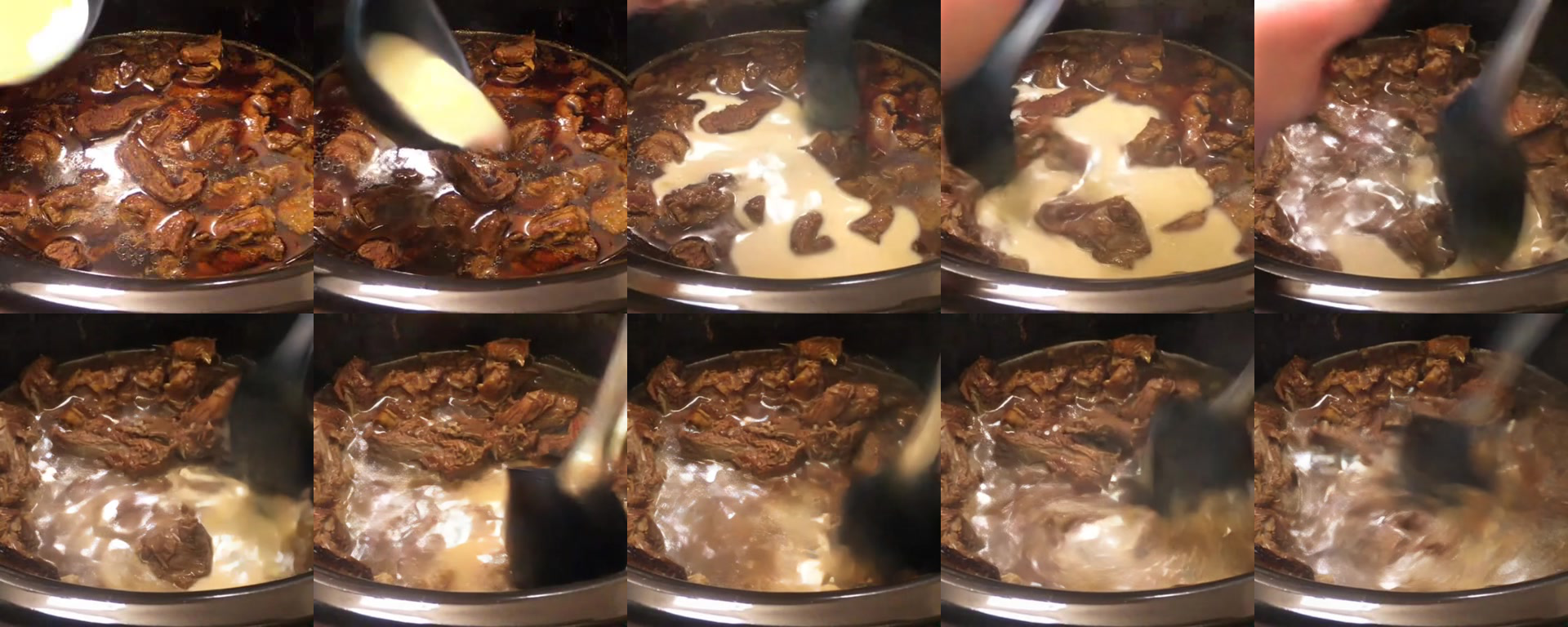}
    \end{subfigure}
    \vspace{-3mm}
    \caption{\textbf{Left:} Ground truth, \textbf{Right:} Inverse generation with GT keyframe. \textbf{Caption:} Chunks of meat are simmering in a dark-colored slow cooker. A hand pours a creamy liquid into the pot, causing the liquid to mix with the meat and broth. The mixture bubbles and thickens as the liquid is added. The person stirs the contents with a black spoon, ensuring the ingredients are well combined. The slow cooker continues to cook the meat, which appears tender and well-cooked.}
    \label{fig:second}
\end{figure*}

\begin{figure*}[!h]
    \centering
    \begin{subfigure}[b]{0.48\linewidth}
        \centering
        \includegraphics[width=\linewidth]{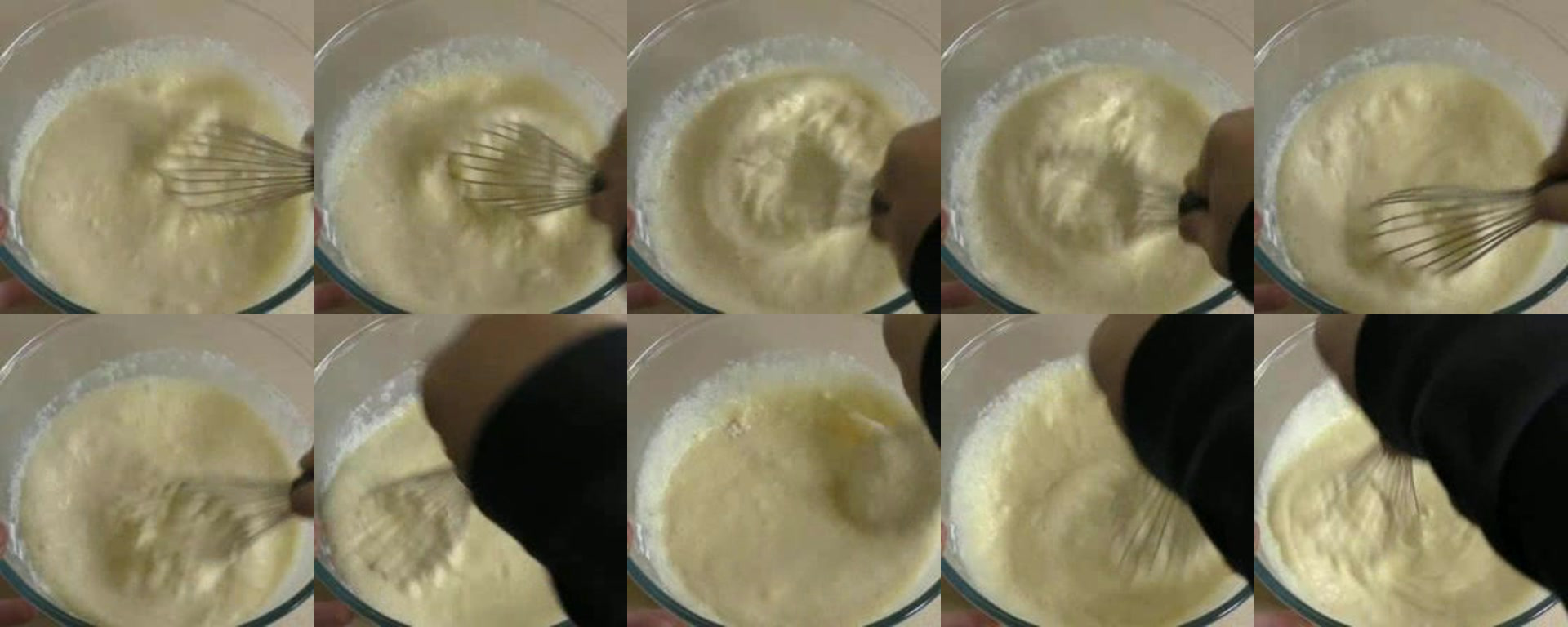}
    \end{subfigure}
    \hfill
    \begin{subfigure}[b]{0.48\linewidth}
        \centering
        \includegraphics[width=\linewidth]{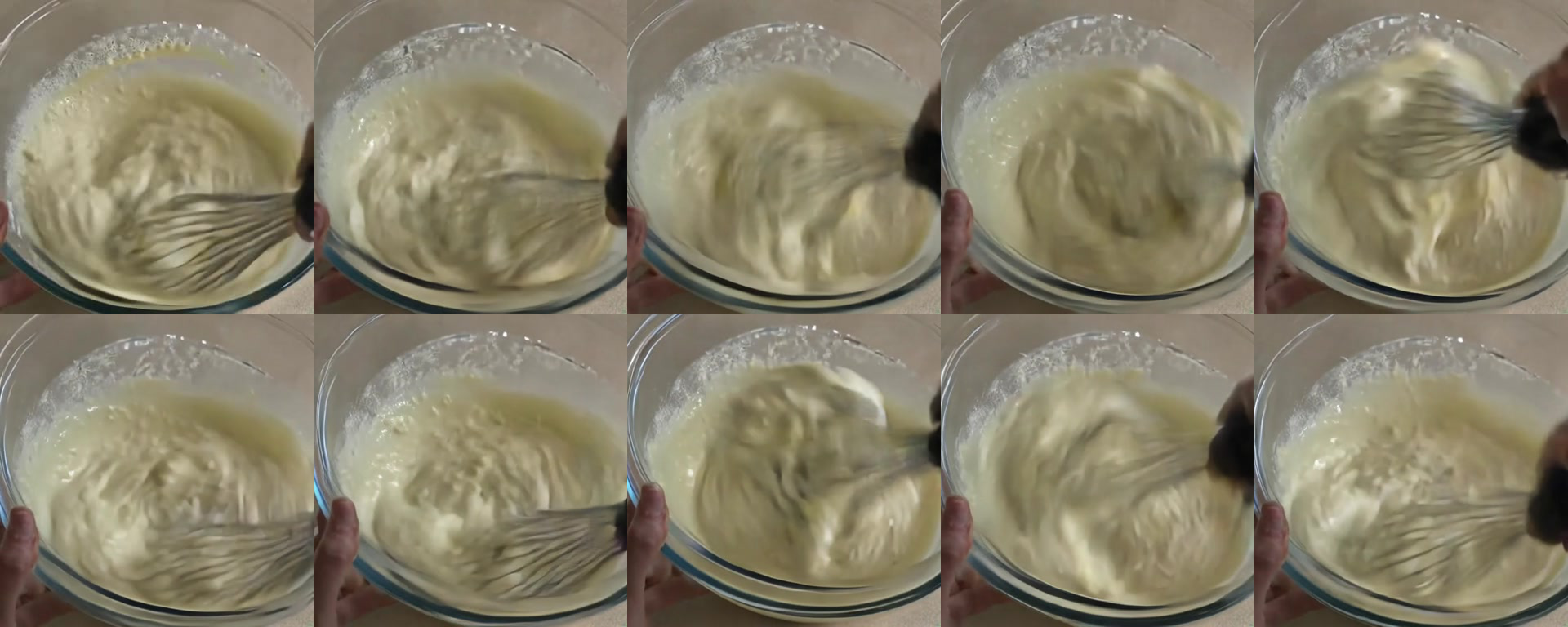}
    \end{subfigure}
    \vspace{-3mm}
    \caption{\textbf{Left:} Ground truth, \textbf{Right:} Inverse generation with GT keyframe. \textbf{Caption:} A person wearing a black sleeve is whisking a creamy mixture in a clear glass bowl. The mixture appears to be a batter or dough, gradually becoming smoother and more uniform. The person's left hand holds the bowl steady on a light-colored countertop. The whisking motion is consistent and thorough, ensuring the mixture is well-blended. The background is plain, focusing attention on the mixing process.}
    \label{fig:third}
\end{figure*}

\begin{figure*}[!h]
    \centering
    \begin{subfigure}[b]{0.48\linewidth}
        \centering
        \includegraphics[width=\linewidth]{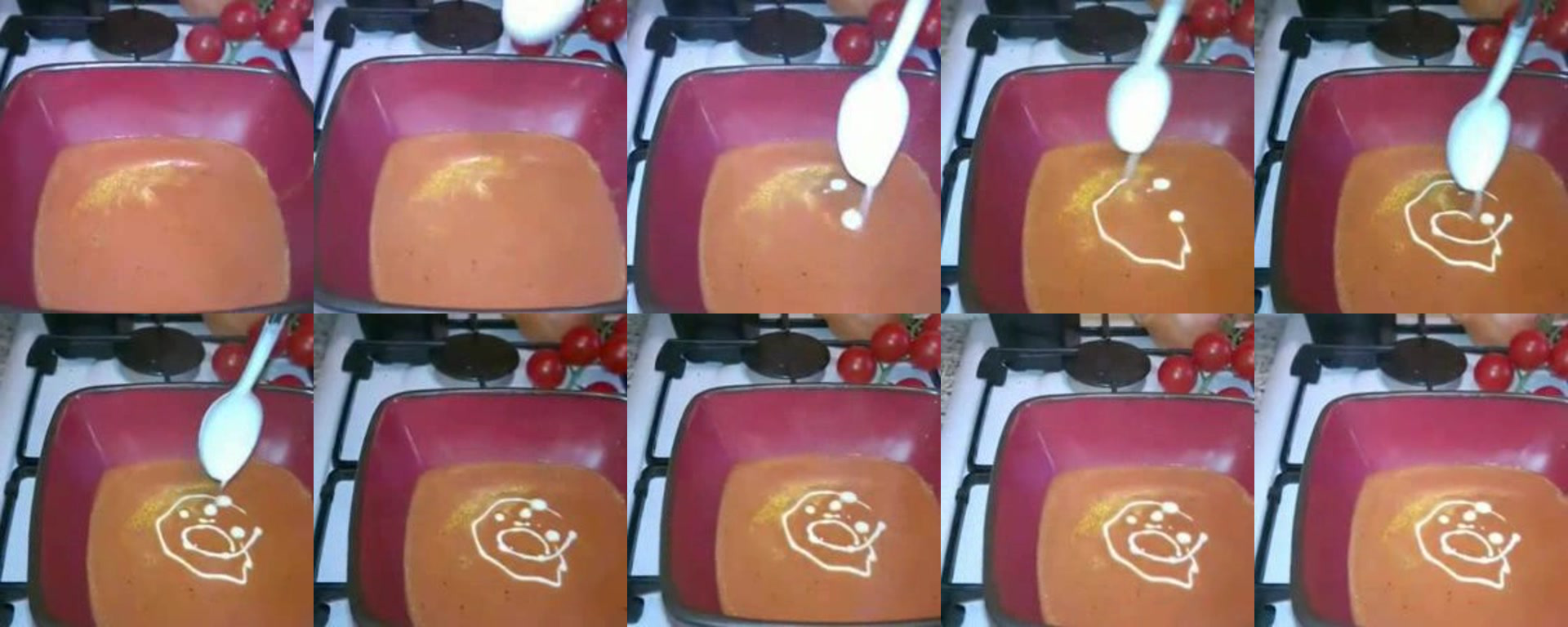}
    \end{subfigure}
    \hfill
    \begin{subfigure}[b]{0.48\linewidth}
        \centering
        \includegraphics[width=\linewidth]{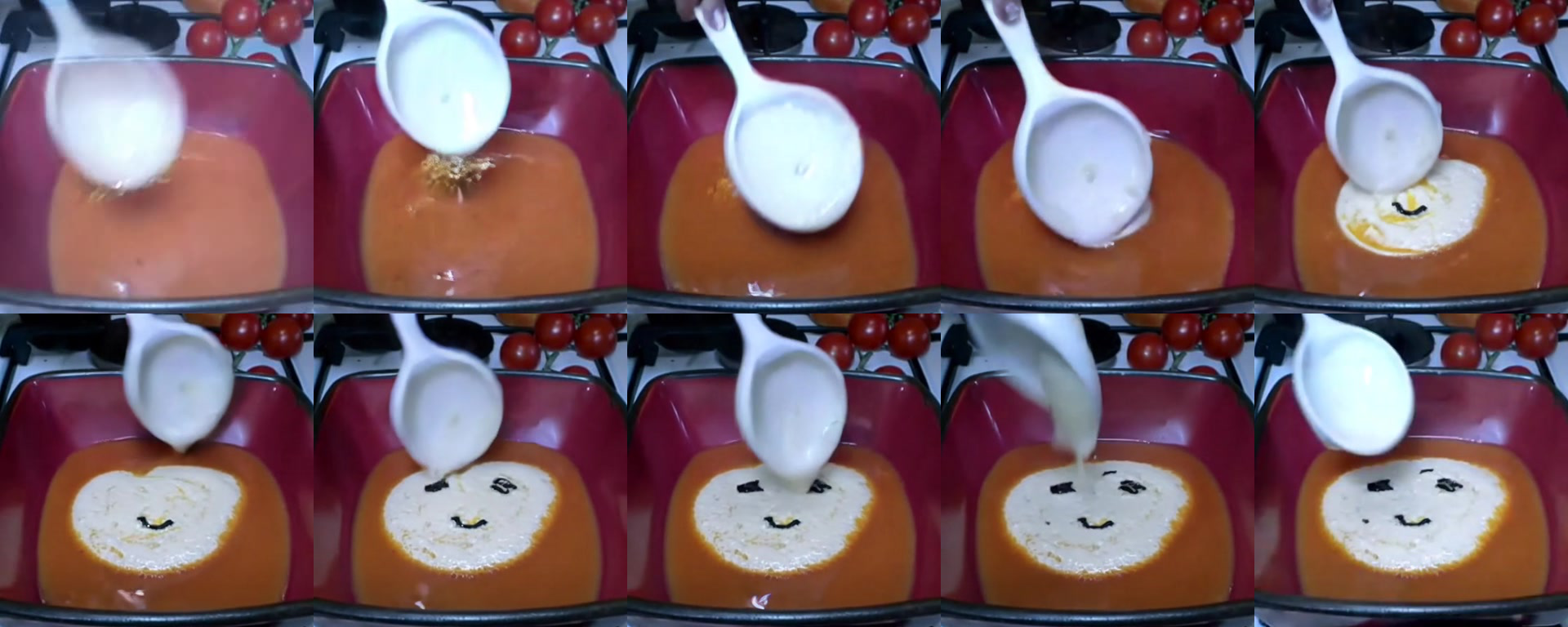}
    \end{subfigure}
    \vspace{-3mm}
    \caption{\textbf{Left:} Ground truth, \textbf{Right:} Inverse generation with GT keyframe. \textbf{Caption:} A red bowl filled with a thick, orange liquid is placed on a stovetop. A woman's hand, holding a white spoon, appears and begins to draw on the surface of the liquid. She creates a face with white cream, adding details to the eyes and mouth. The background shows a granite countertop with a bunch of red tomatoes and a white pot. The woman continues to add finishing touches to the face.}
    \label{fig:fourth}
\end{figure*}

As discussed in \Cref{sec:inverse_video_gen}, high-quality captions, especially with ground truth keyframes, enable effective video reconstruction. We compare ground truth video frames with inversely generated frames using the GT first keyframe and annotated captions, as shown in \Cref{fig:second,fig:third,fig:fourth}. The reconstruction aligns well with the narrative, accurately capturing actions, though patterns and interactions differ slightly from the original video. This shows that while the captions convey crucial information for reconstruction, they lack finer visual details, a limitation for current vision-language models and human annotators.

For example, in \Cref{fig:second}, the ground truth shows a hand pouring creamy liquid into a slow cooker and stirring, while the generated frames replicate the actions with slight differences in texture and liquid mixing. Similarly, in \Cref{fig:fourth}, the ground truth shows a face drawn with cream on orange liquid, but the generated frames vary in precision and interaction details. These examples highlight the captions' strength in preserving narrative flow while exposing gaps in capturing fine-grained visual detail.

\subsection{Prompt for Video Captioning}

Below is the prompt we designed to effectively caption video clips and also for benchmarking VLMs, ensuring detailed and accurate descriptions while avoiding redundancy:

\begin{lstlisting}[basicstyle=\ttfamily\small, caption={\textbf{Video Captioning Prompt}}]
You are an expert in describing videos and catching the sequential motions from video frames. 

For the given ten video frames, you need to generate a detailed good description within five 
sentences / 80 words. Please do not include the word 'frame' or 'frames' in your answer. If 
the gender of a person is clear, use 'he' or 'she' instead of they. Do not describe a single 
motion/action twice like 'xxx continues doing yyy'. Don't assume actions like discussion or 
having a conversation unless it is very clear in the frames. Describe the video given the 
frame sequence. Describe both the appearance of people (gender, clothes, etc), objects, 
background in the video, and the actions they take.
\end{lstlisting}
    \vspace{-4mm}

\subsection{GPT-4o Evaluation on Captions}

Below is the evaluation prompt designed to objectively assess the quality of video captions generated by a Large Multimodal Model (LMM), focusing on coverage and hallucination.

\begin{lstlisting}[basicstyle=\ttfamily\small, backgroundcolor=\color{gray!10}, caption={\textbf{GPT-4o Evaluation Prompt}}]
Your role is to serve as an impartial and objective evaluator of a video caption provided by
a Large Multimodal Model (LMM). Based on the input frames of a video, assess primarily on 
two criteria: the coverage of video elements in the caption and the absence of hallucinations
in the response. In this context, 'hallucination' refers to the model generating content not
present or implied in the video, mainly focused on incorrect details about objects, actions,
counts, temporal order, or other aspects not evidenced in the video frames.

To evaluate the LMM's response:
    Start with a brief explanation of your evaluation process.
    Then, assign a rating from the following scale:
        Rating 6: Very informative with good coverage, no hallucination
        Rating 5: Very informative, no hallucination
        Rating 4: Somewhat informative with some missing details, no hallucination
        Rating 3: Not informative, no hallucination
        Rating 2: Very informative, with hallucination
        Rating 1: Somewhat informative, with hallucination
        Rating 0: Not informative, with hallucination
Do not provide any other output symbols, text, or explanation for the score.
\end{lstlisting}
\vspace{-4mm}

\subsection{Human Evaluation on Captions}

\begin{table*}[!ht]
    \centering
    \tablestyle{4pt}{1.2}
    \begin{tabular}{c|cccc}
        \textbf{Matching Tier} & \textbf{Action (Important Info.)} & \textbf{Object (Important Info.)} & \textbf{Score} \\ \shline
        Very Match   & Good Coverage, No Hallucination   & Good Coverage, No Hallucination   & \textbf{100}  \\
        Good Match   & Good Coverage, Limited Hallucination & Good Coverage, Limited Hallucination & \textbf{85}  \\
        Somehow Match & Fair Coverage, Some Hallucination & Fair Coverage, Some Hallucination & \textbf{70}  \\
        Not Match    & Little Coverage or High Hallucination & Little Coverage or High Hallucination & \textbf{0}   \\
    \end{tabular}
    \caption{\textbf{Human Evaluation Matching Rules.} Captions are rated based on coverage and hallucination levels, using four matching tiers.}
    \vspace{-2mm}
    \label{tab:matching-tier}
\end{table*}

We assess the quality of our captions through evaluations by six human annotators, who rate the captions based on two key criteria: the coverage of video elements (such as objects and actions) and the absence of hallucinations, defined as generating content unsupported or absent in the video~\cite{zhang2024direct}. As shown in Table~\ref{tab:caption-quality}, our captions achieve a high human evaluation score of \textbf{82.0}, surpassing the state-of-the-art open-source VLM (Qwen2-VL-72B) score of \textbf{79.3}. These results demonstrate the superior quality of our captions, which are more aligned with human preferences and exhibit better narrative accuracy.

For evaluation, annotators rate the captions across four tiers—Very Match, Good Match, Somehow Match, and Not Match—based on consistency with video content. The scoring rubric, detailed in Table~\ref{tab:matching-tier}, considers both coverage and hallucination levels. Our captioner consistently achieves high scores in the top tiers, validating its reliability and quality for narrative video generation.

\onecolumn

\section{Implementation Details}
\label{sec:appendix_implementation_details}
We provide the training and inference hyperparameters for the interleaved auto-regressive model and the visual-conditioned video generation model in \Cref{tab:appendix_implementation_details_interleaved} and \Cref{tab:appendix_implementation_details_videogen}, respectively. The interleaved auto-regressive model is trained on images with a resolution of $448 \times 448$, using a batch size of 512 and bfloat16 precision. It employs AdamW as the optimizer, with a peak learning rate of $2 \times 10^{-4}$ and a cosine decay schedule, training for 2,500 steps. Training context pairs vary between 2 and 8, while inference always uses 8 pairs for consistency. The visual-conditioned video generation model processes video data at a resolution of $448 \times 448 \times T$, with a batch size of 64 and bfloat16 precision. It uses AdamW with a peak learning rate of $1 \times 10^{-5}$ and a constant decay schedule, training for 20,000 steps to handle temporal conditioning effectively.

\begin{table}[!h]
\centering
\small
\begin{tabular}{cc}
\textbf{Configuration}               & \textbf{Setting}            \\ \shline
Image resolution                     & $448 \times 448$                     \\
Optimizer                            & AdamW                                \\
Optimizer hyperparameters            & $\beta_1 = 0.9, \beta_2 = 0.98, \epsilon = 10^{-6}$ \\
Peak learning rate                   & $2 \times 10^{-4}$                   \\
Learning rate schedule               & Linear warm-up, cosine decay         \\
Gradient clip                        & $1.0$                                \\
Total training steps                 & $2,500$                              \\
Warm-up steps                        & $200$                                \\
Batch size                           & $512$                                \\
Numerical precision                  & bfloat16                             \\
Training context pairs               & [2, 8]                               \\
Inference context pairs              & 8                                    \\
\end{tabular}
\caption{\textbf{Implementation details of the interleaved auto-regressive model.}}
\label{tab:appendix_implementation_details_interleaved}
\end{table}

\begin{table}[!h]
\centering
\small
\begin{tabular}{cc}
\textbf{Configuration}               & \textbf{Setting}            \\ \shline
Image Resolutions             & $448 \times 448 \times 4$             \\
Optimizer                            & AdamW                                \\
Optimizer hyperparameters            & $\beta_1 = 0.9, \beta_2 = 0.999, \epsilon = 10^{-8}$ \\
Peak learning rate                   & $5 \times 10^{-4}$                   \\
Learning rate schedule               & Linear warm-up, constant             \\
Gradient clip                        & $1.0$                                \\
Total training steps                 & $5,000$                             \\
Warm-up steps                        & $500$                              \\
Batch size                           & $16$                                 \\
Optional Masking Probability & 0.25 \\
Numerical precision                  & bfloat16                             \\
\end{tabular}
\caption{\textbf{Implementation details of the rolling context conditioned render.}}
\label{tab:appendix_implementation_details_context_conditioned_render}
\end{table}

\begin{table}[!h]
\centering
\small
\begin{tabular}{cc}
\textbf{Configuration}               & \textbf{Setting}            \\ \shline
Image/Video resolution               & $448 \times 448 \times T$             \\
Optimizer                            & AdamW                                \\
Optimizer hyperparameters            & $\beta_1 = 0.9, \beta_2 = 0.95, \epsilon = 10^{-8}$ \\
Peak learning rate                   & $1 \times 10^{-5}$                   \\
Learning rate schedule               & Linear warm-up, constant             \\
Gradient clip                        & $1.0$                                \\
Total training steps                 & $20,000$                             \\
Warm-up steps                        & $1,000$                              \\
Batch size                           & $64$                                 \\
Numerical precision                  & bfloat16                             \\
\end{tabular}
\caption{\textbf{Implementation details of the visual-conditioned video generation model.}}
\label{tab:appendix_implementation_details_videogen}
\end{table}

\vspace{1cm}

\clearpage
\onecolumn

\section{Action-Caption Matching Pseudo Code}
\label{sec:appendix_action_caption_matching_pseudo_code}
The action-caption matching algorithm detailed in \Cref{alg:action_matching} aligns video clips with actions based on temporal overlap and specific rules. It uses the Intersection over Union (IoU) to measure the overlap between the time intervals of video clips and actions. A match is identified if either the IoU exceeds 0.5 or all of the following conditions are met: the start time difference (\texttt{start\_diff}) is less than 5 seconds, the clip's end time exceeds the action's end time, and the IoU is greater than 0.2.

The algorithm processes each video iteratively. For each video, it retrieves all associated actions and their time intervals. Then, for each clip in the video, it calculates the IoU with every action and evaluates the matching conditions. Valid matches, along with their metadata (clip info and descriptions), are stored in a list $\mathcal{M}$. This systematic approach ensures that the matched actions and captions are temporally consistent, providing high-quality annotations for keyframe visual states.

\begin{algorithm}[h!]
\begin{algorithmic}[1]
\Function{IoU}{$[s_1, e_1], [s_2, e_2]$}
    \State $\text{intersection} \gets \max(0, \min(e_1, e_2) - \max(s_1, s_2))$
    \State $\text{union} \gets \max(e_1, e_2) - \min(s_1, s_2)$
    \If{$\text{union} > 0$}
        \State \Return $\frac{\text{intersection}}{\text{union}}$
    \Else
        \State \Return $0$
    \EndIf
\EndFunction

\State Initialize an empty list $\mathcal{M} \gets []$

\ForAll{$v \in \mathcal{V}$}
    \State $v_{\text{id}} \gets v.\text{id}$
    \If{$v_{\text{id}} \in \mathcal{A}$}
        \State $\mathcal{A}_{v} \gets \mathcal{A}[v_{\text{id}}]$
        \State $\text{action\_times} \gets \mathcal{A}_v.\text{times}$
        \State $\text{action\_descriptions} \gets \mathcal{A}_v.\text{descriptions}$
        
        \ForAll{$c \in v.\text{clips}$}
            \State $[s_c, e_c] \gets c.\text{start\_end}$
            \ForAll{$a \in \text{action\_times}$}
                \State $[s_a, e_a] \gets a$
                \State $\text{start\_diff} \gets |s_c - s_a|$
                \State $\text{iou} \gets \Call{IoU}{[s_c, e_c], [s_a, e_a]}$
                
                \If{$(\text{start\_diff} < 5 \land e_c > e_a \land \text{iou} > 0.2) \lor \text{iou} > 0.5$}
    \State Create match: $\mathcal{M} \gets \mathcal{M} \cup m$
                \EndIf
            \EndFor
        \EndFor
    \EndIf
\EndFor

\State \Return $\mathcal{M}$
\end{algorithmic}
\caption{\textbf{Pseudo code for action-caption matching.}}
\label{alg:action_matching}
\end{algorithm}

\clearpage

\onecolumn

\section{CLIP beats VAE for interleaved generation.} 
\label{sec:clip_beats_vae}
\begin{figure*}[h]
    \centering
    \includegraphics[width=0.6\linewidth]{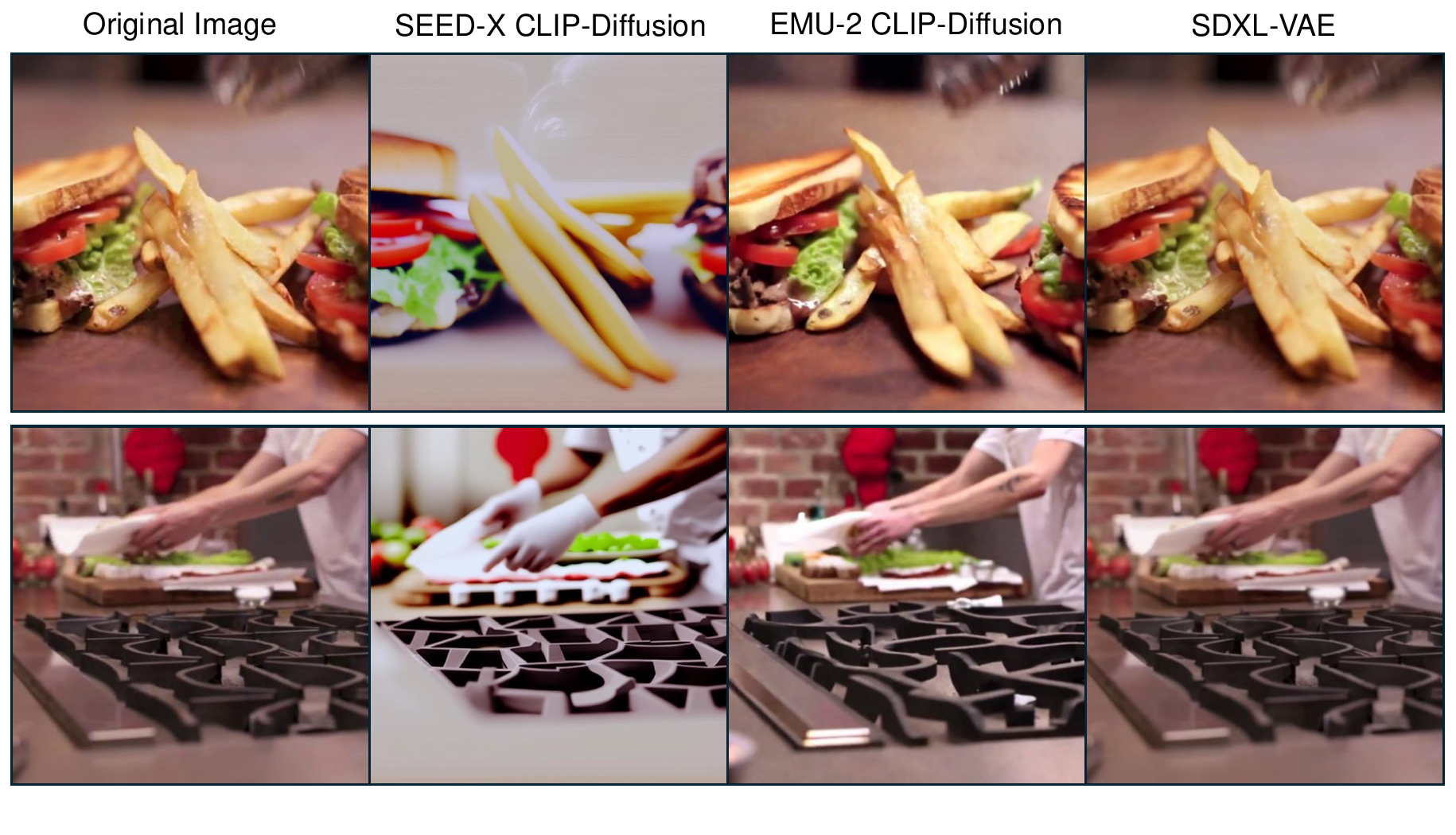}
    \vspace{-4mm}
    \caption{\textbf{Auto-encoded results with different latent spaces.} While SEED-X and EMU-2 both use a CLIP vision encoder and a diffusion model (\ie finetuned SDXL) as decoder for autoencoding visual latents, SEED-X is semantic-biased and EMU-2 keeps much more visual details. SDXL-VAE shows the best image reconstruction ability, however, the latent space is not aligned with language (\ie without pretraining on image-text pairs like CLIP). }
    \label{fig:autoencode_comparison}
\end{figure*}

We experiment with three different auto-encoded visual latent spaces for regression: the EMU-2~\cite{sun2024generative} CLIP-Diffusion autoencoder, the SEED-X CLIP-Diffusion autoencoder, and the KL Variational autoencoder (VAE) used by SDXL. Both SEED-X and EMU-2 use a CLIP vision encoder and a finetuned SDXL diffusion model as the decoder for encoding visual latent.
From appendix \Cref{fig:autoencode_comparison}, we observe that SDXL-VAE achieves the best reconstruction quality. However, in terms of visual generation quality, as shown in \Cref{tab:regression_target_comparision}, the CLIP-Diffusion based autoencoders significantly outperform VAE (\ie, \textbf{+12.2} CLIP-T score and \textbf{256.6} better FID). This suggests that CLIP embeddings are more suitable for interleaved visual generation compared to VAE's latent space. This is reasonable, as SDXL-VAE is not aligned with language and lacks semantics.

\begin{table}[h]
    \centering
    \tablestyle{1pt}{1.2}
    \scriptsize
    \begin{tabular}{cc c c |c c}

        \textbf{Method} & \textbf{Autoencoder Style} & \textbf{VL Aligned.} & \textbf{ Recon. Ability} & \textbf{CLIP-T} & \textbf{FID} \\ \shline
        SDXL-VAE & Variational U-Net & \xmark & High & 13.2 & 286.6 \\
        EMU-2 & CLIP-Diffusion & \cmark &Medium   & \textbf{25.4} & 76.7 \\
        SEED-X & CLIP-Diffusion & \cmark &Low  & 25.1 & \textbf{30.1}\\
        
    \end{tabular}
    \caption{\textbf{Visual latent spaces for visual regression.} The VAE latent space is challenging for auto-regressive models to regress in a single step due to its limited correlation with language. In contrast, the language-aligned latent spaces (EMU-2 and SEED-X) allow for easier and effective regression in an interleaved manner.}
    \label{tab:regression_target_comparision}
\end{table}

\vspace{-0.5cm}

\begin{figure}[h]
    \centering
    \includegraphics[width=0.9\linewidth]{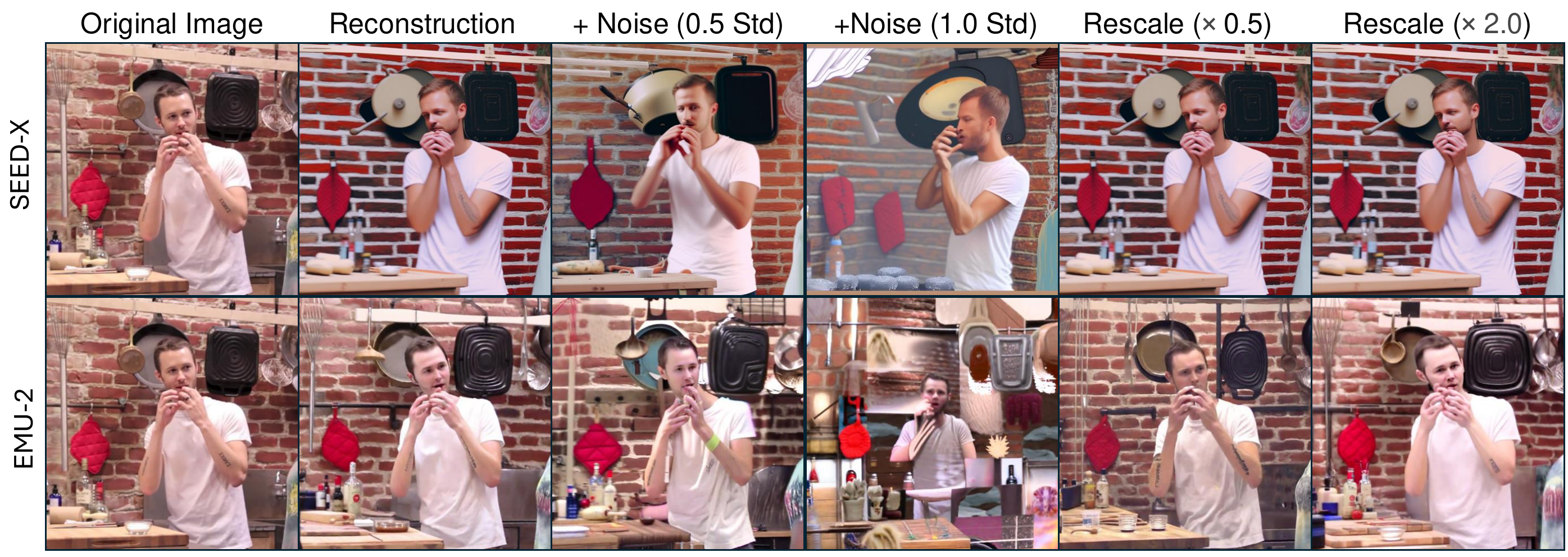}
    \vspace{-3mm}
    \caption{\textbf{Both Scale and Direction Matters.} We experiment with pseudo regression errors by altering latent direction and scale using Gaussian noise and scaling factors. Reconstruction results confirm that preserving both scale and direction is important for latent regression.}

    \label{fig:scale_direction}
\end{figure}

\clearpage
\onecolumn
\section{Generated Video Examples}
\label{sec:appendix_generated_video_examples}
\begin{figure*}[!h]
    \centering
    \begin{subfigure}[b]{0.48\linewidth}
        \centering
        \includegraphics[width=\linewidth]{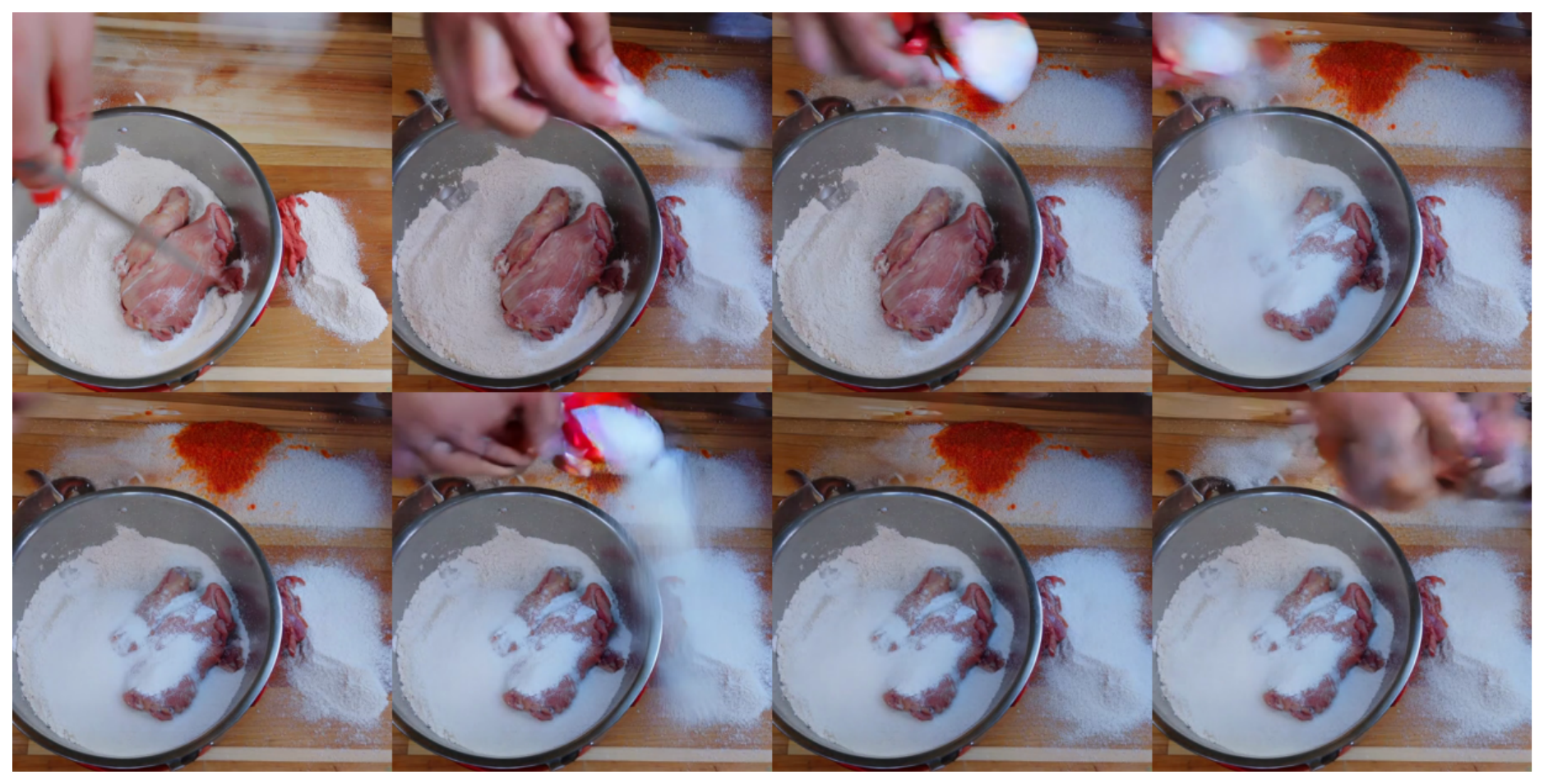}
        \caption{\textbf{Action:} Add raw chicken pieces and seasoning to a bowl of flour.}
    \end{subfigure}
    \hfill
    \begin{subfigure}[b]{0.48\linewidth}
        \centering
        \includegraphics[width=\linewidth]{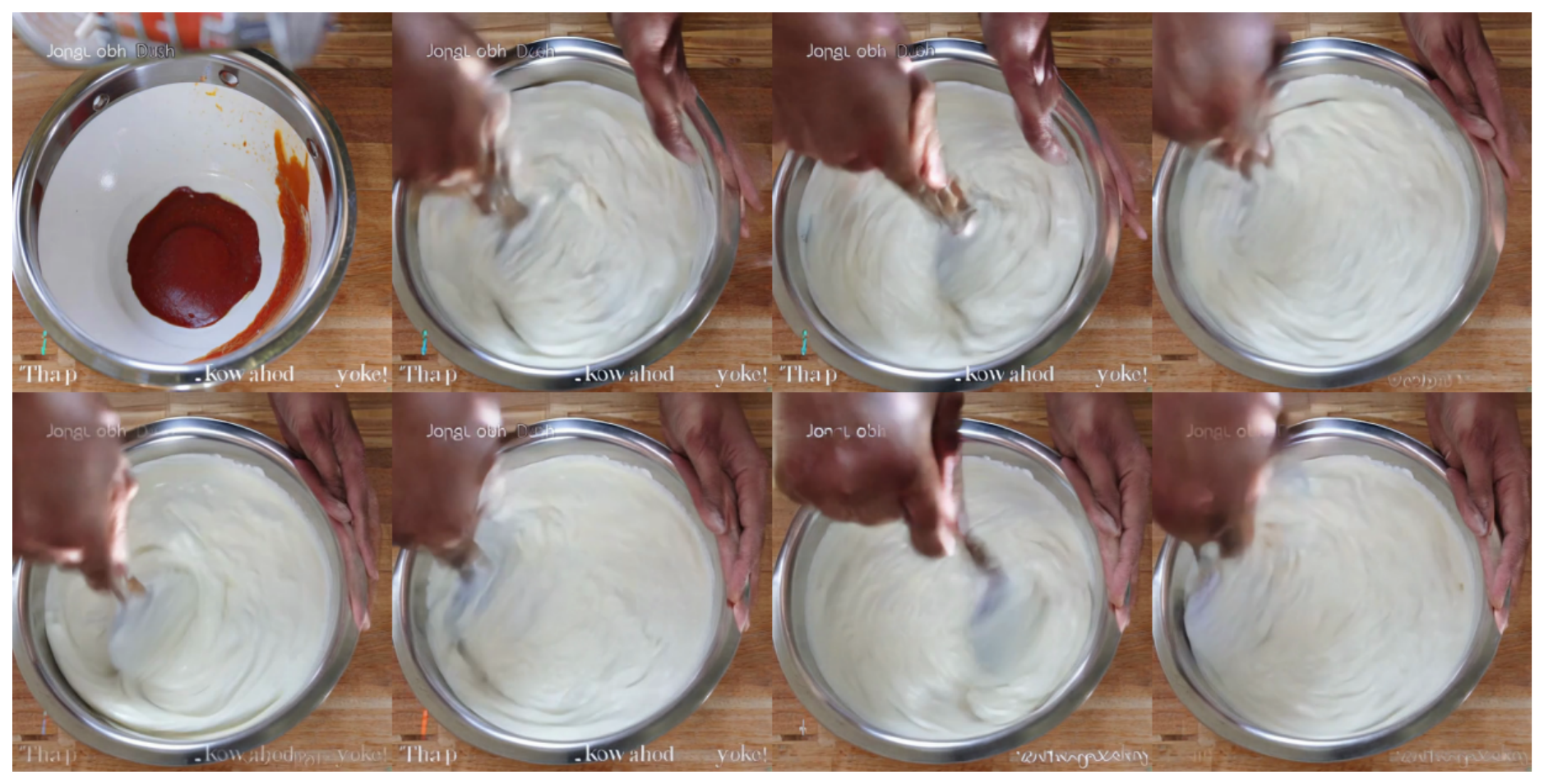}
        \caption{\textbf{Action:} Mix yogurt or buttermilk with seasoning in a bowl.}
    \end{subfigure}

    \begin{subfigure}[b]{0.48\linewidth}
        \centering
        \includegraphics[width=\linewidth]{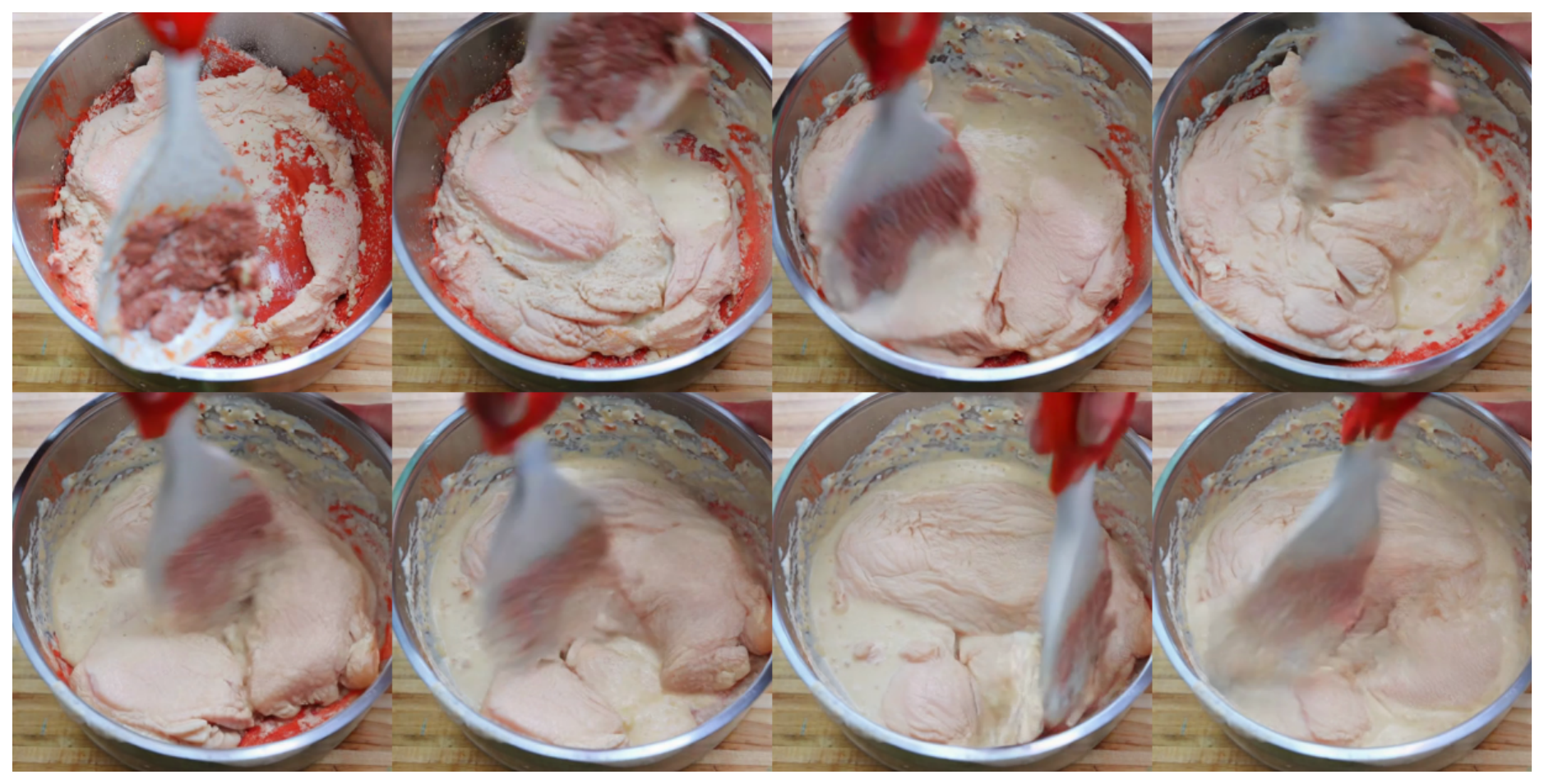}
        \caption{\textbf{Action:} Dip chicken pieces into the batter to coat evenly.}
    \end{subfigure}
    \hfill
    \begin{subfigure}[b]{0.48\linewidth}
        \centering
        \includegraphics[width=\linewidth]{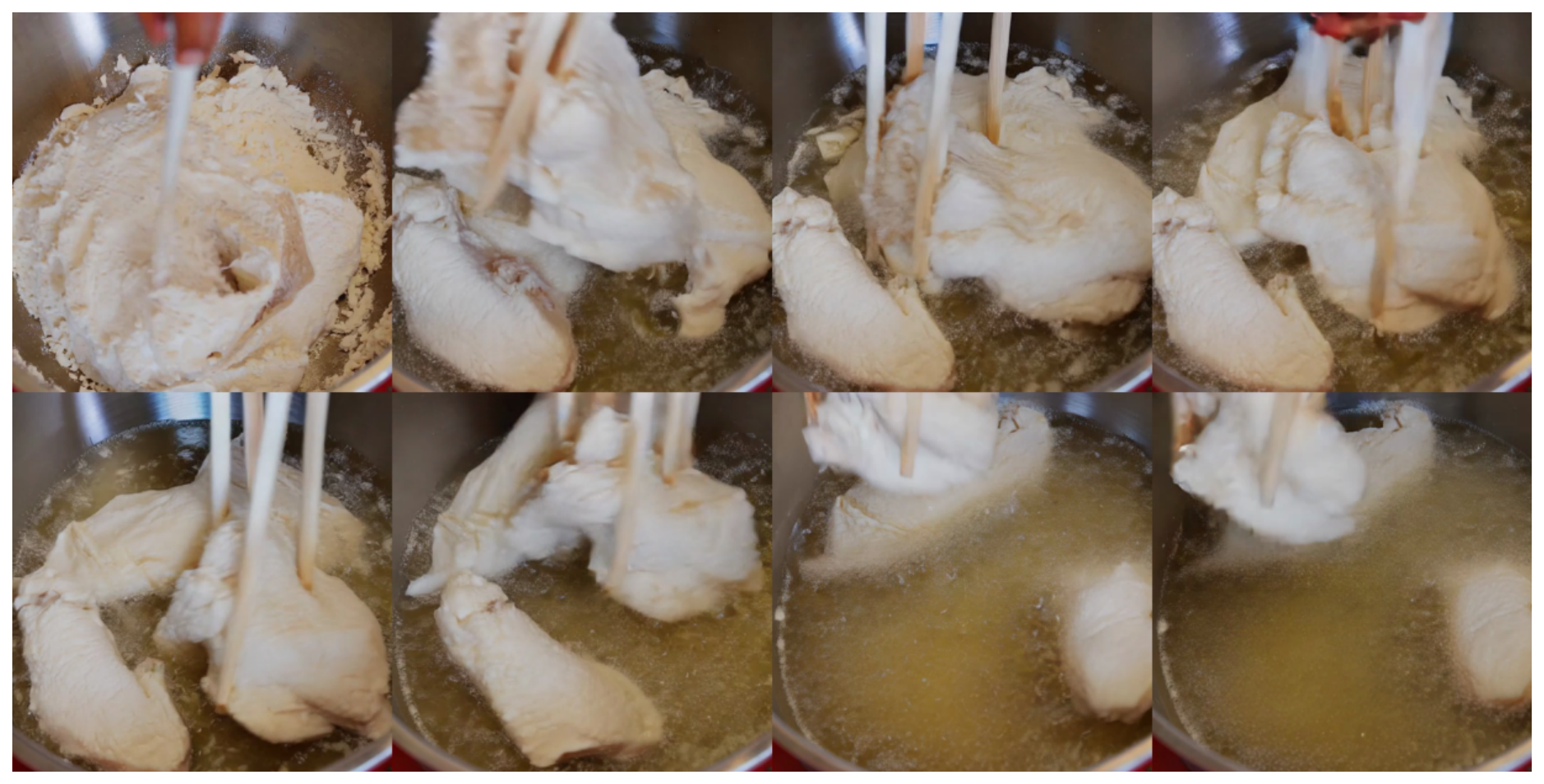}
        \caption{\textbf{Action:} Coat the battered chicken in the flour mixture.}
    \end{subfigure}

    \begin{subfigure}[b]{0.48\linewidth}
        \centering
        \includegraphics[width=\linewidth]{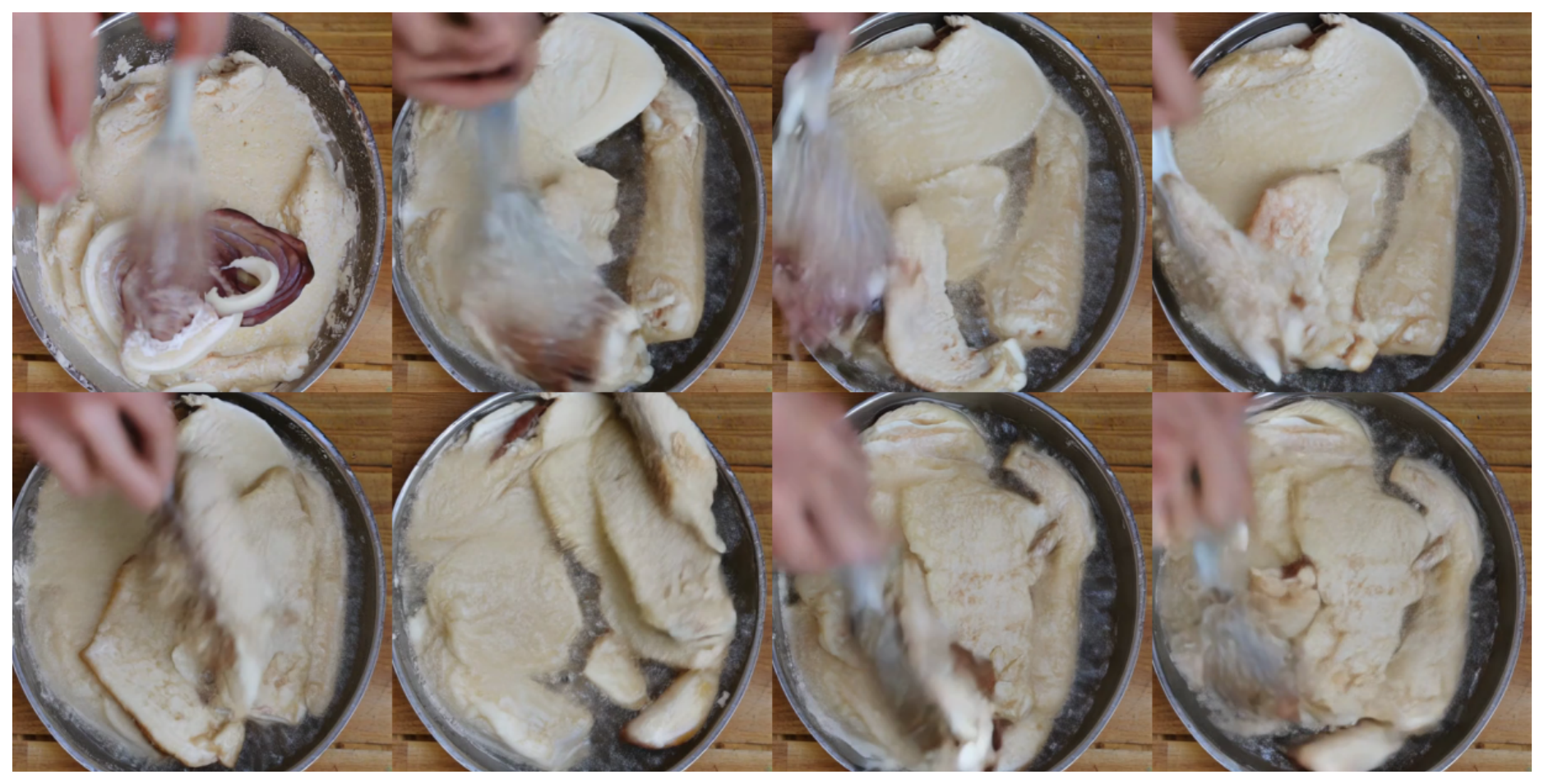}
        \caption{\textbf{Action:} Fry the coated chicken in hot oil until crispy and golden.}
    \end{subfigure}
    \hfill
    \begin{subfigure}[b]{0.48\linewidth}
        \centering
        \includegraphics[width=\linewidth]{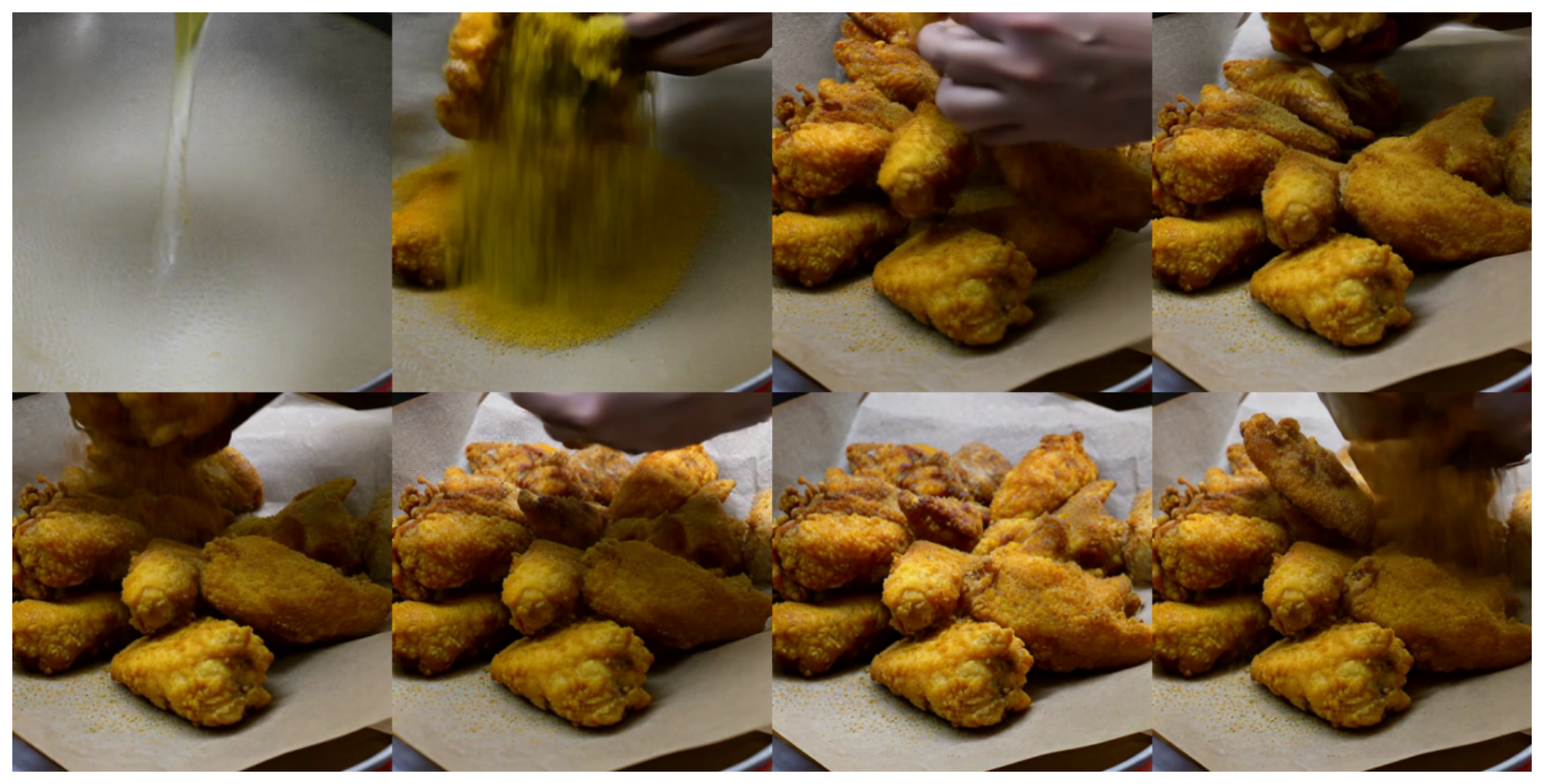}
        \caption{\textbf{Action:} Sprinkle seasoning on the fried chicken and serve.}
    \end{subfigure}

    \caption{\textbf{Video generation example.} Our pipeline effectively accomplishes long narrative video generation by producing six essential steps (\ie, video clips) for cooking "Fried Chicken." It delivers a clear, structured, and instructional step-by-step narrative, showcasing the model's capability to generate coherent and comprehensive videos.}
    \label{fig:fried_chicken_frames}
\end{figure*}

\begin{figure*}[h!]
    \centering
    \begin{subfigure}[b]{0.48\linewidth}
        \centering
        \includegraphics[width=\linewidth]{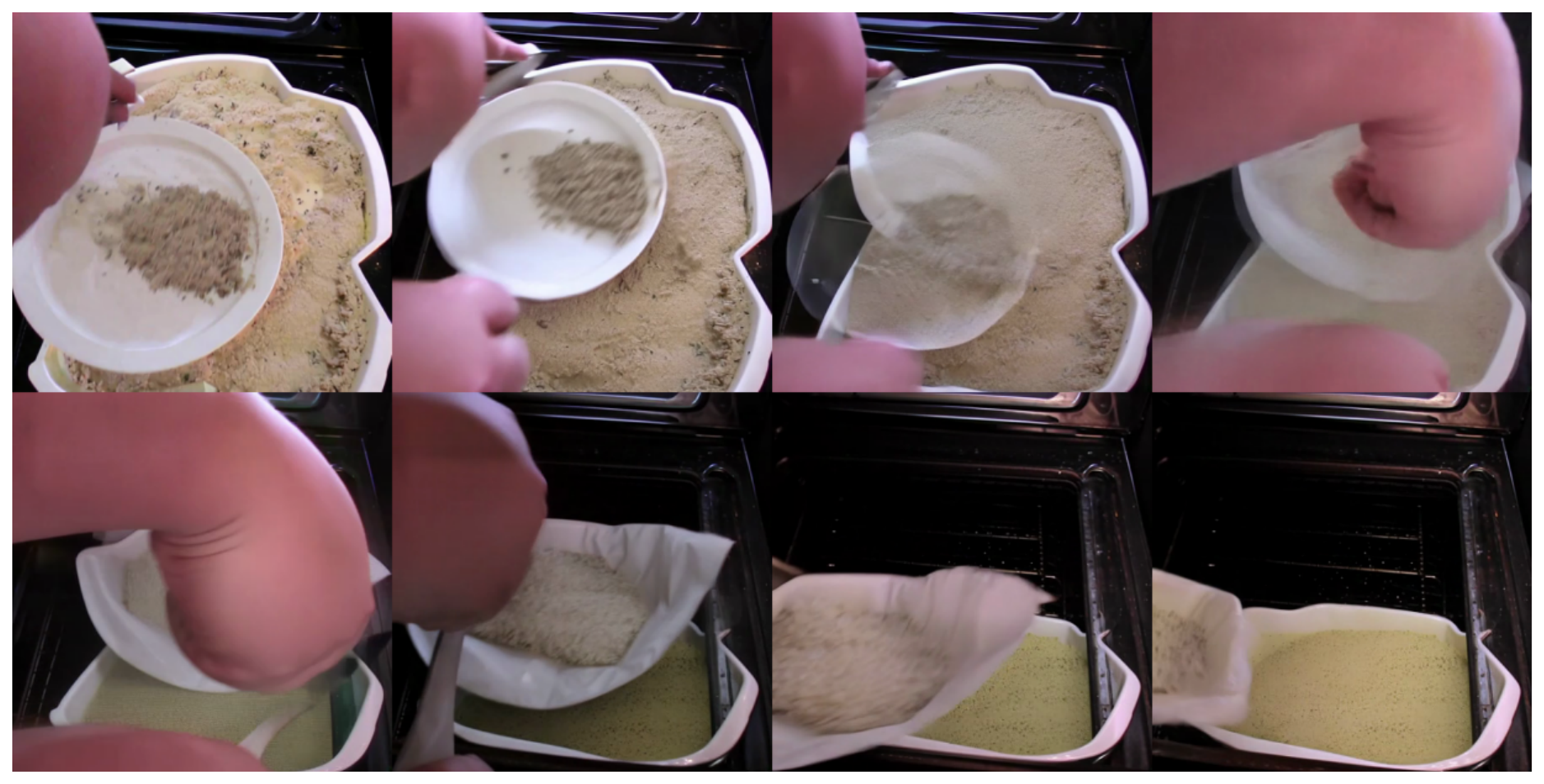}
        \caption{\textbf{Action:} Mix chopped vegetables in a glass bowl.}
    \end{subfigure}
    \hfill
    \begin{subfigure}[b]{0.48\linewidth}
        \centering
        \includegraphics[width=\linewidth]{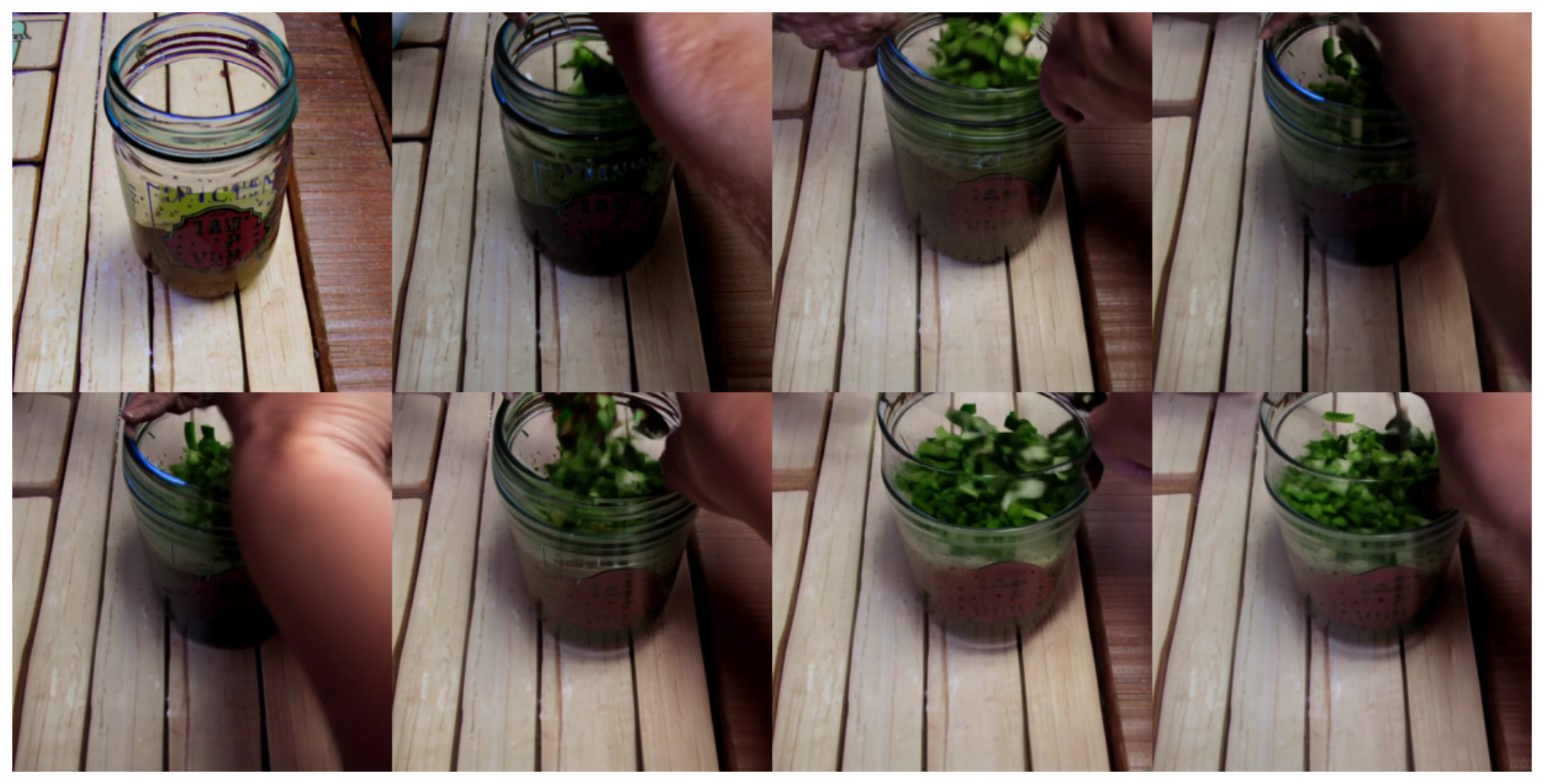}
        \caption{\textbf{Action:} Add seasoning to the mixture of chopped vegetables.}
    \end{subfigure}

    \begin{subfigure}[b]{0.48\linewidth}
        \centering
        \includegraphics[width=\linewidth]{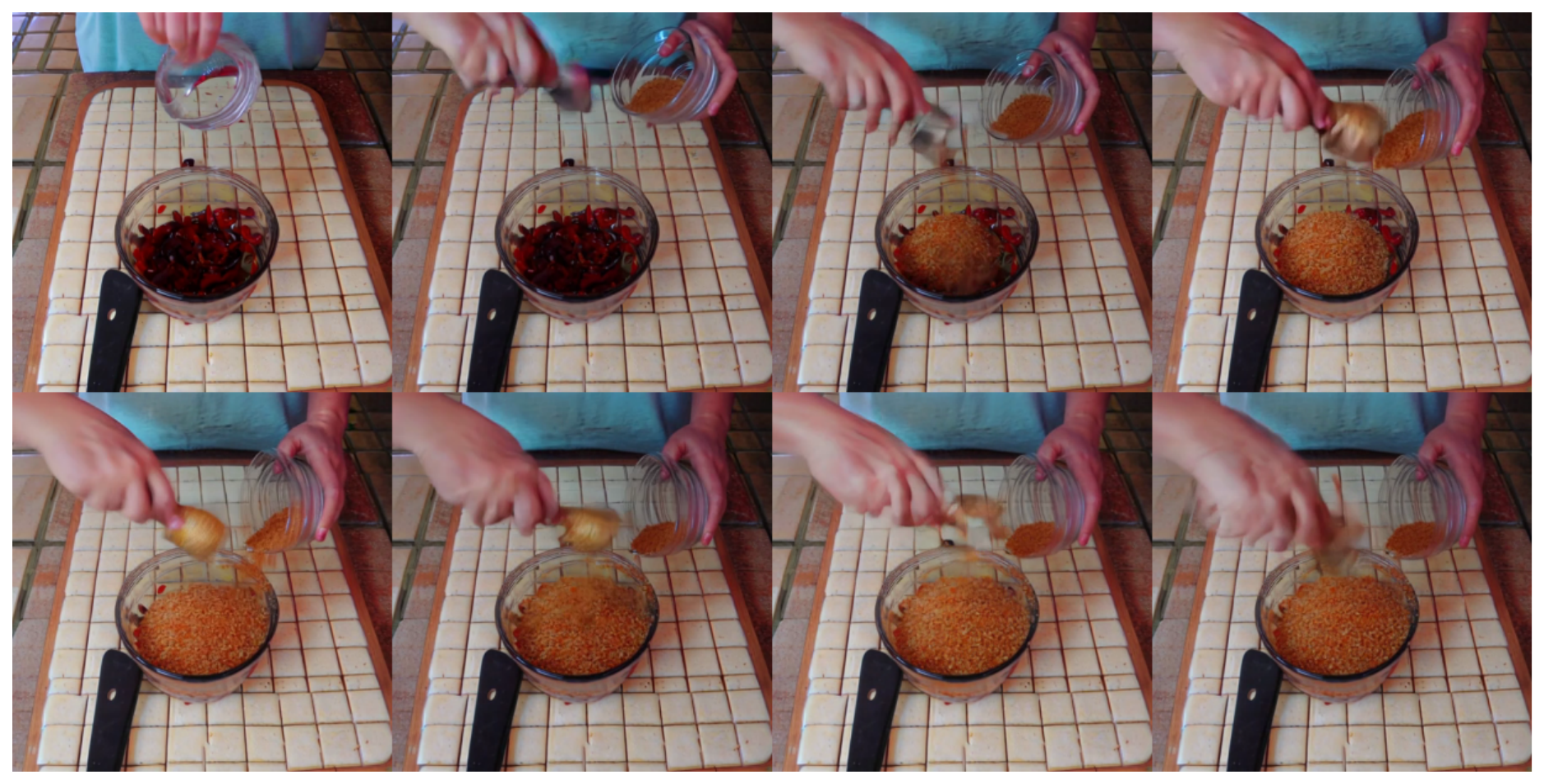}
        \caption{\textbf{Action:} Thoroughly mix the seasoned vegetable mixture.}
    \end{subfigure}
    \hfill
    \begin{subfigure}[b]{0.48\linewidth}
        \centering
        \includegraphics[width=\linewidth]{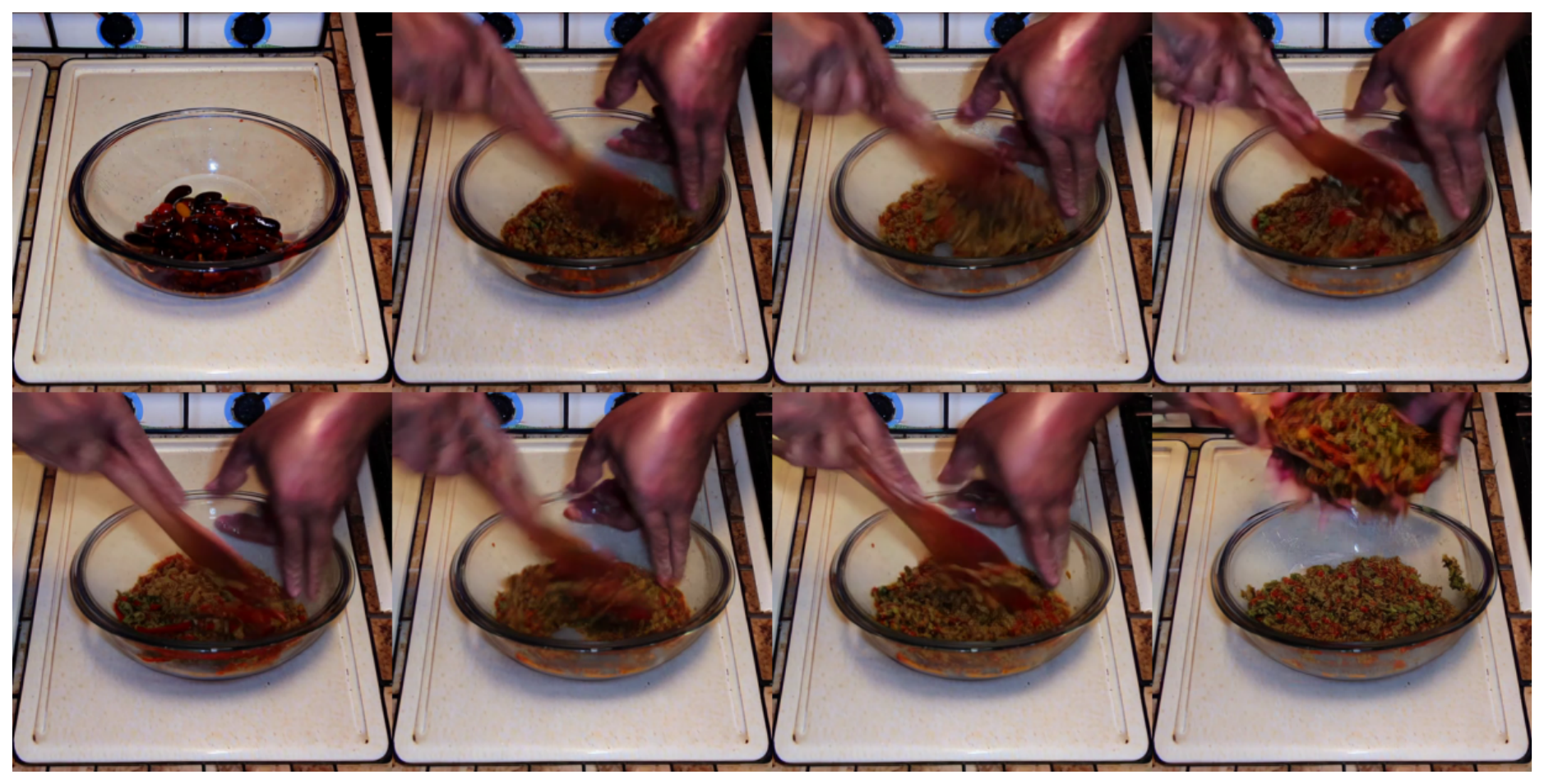}
        \caption{\textbf{Action:} Add chicken pieces to vegetable and chicken mixture.}
    \end{subfigure}

    \begin{subfigure}[b]{0.48\linewidth}
        \centering
        \includegraphics[width=\linewidth]{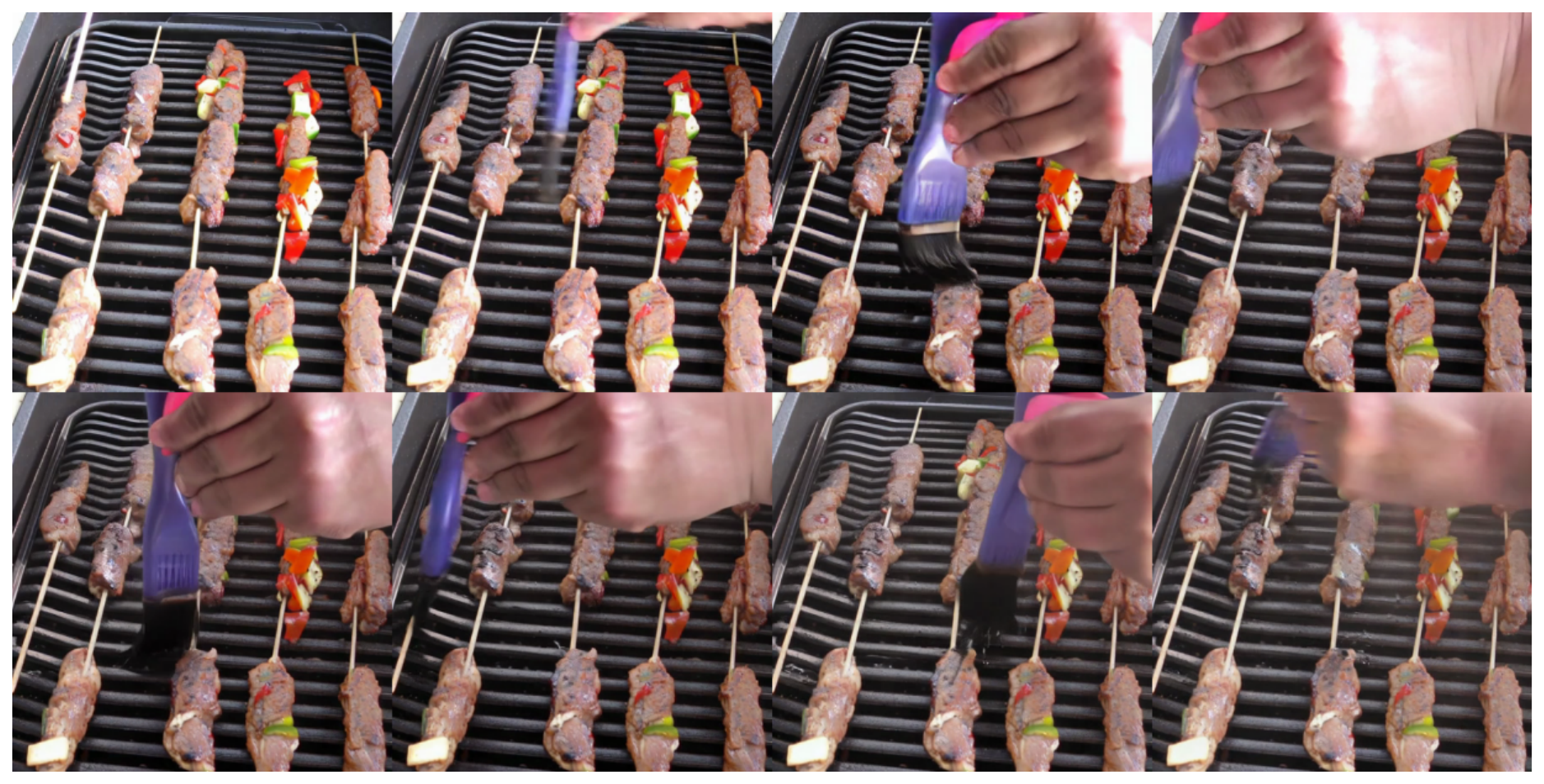}
        \caption{\textbf{Action:} Brush oil onto the skewered chicken and vegetable kebabs.}
    \end{subfigure}
    \hfill
    \begin{subfigure}[b]{0.48\linewidth}
        \centering
        \includegraphics[width=\linewidth]{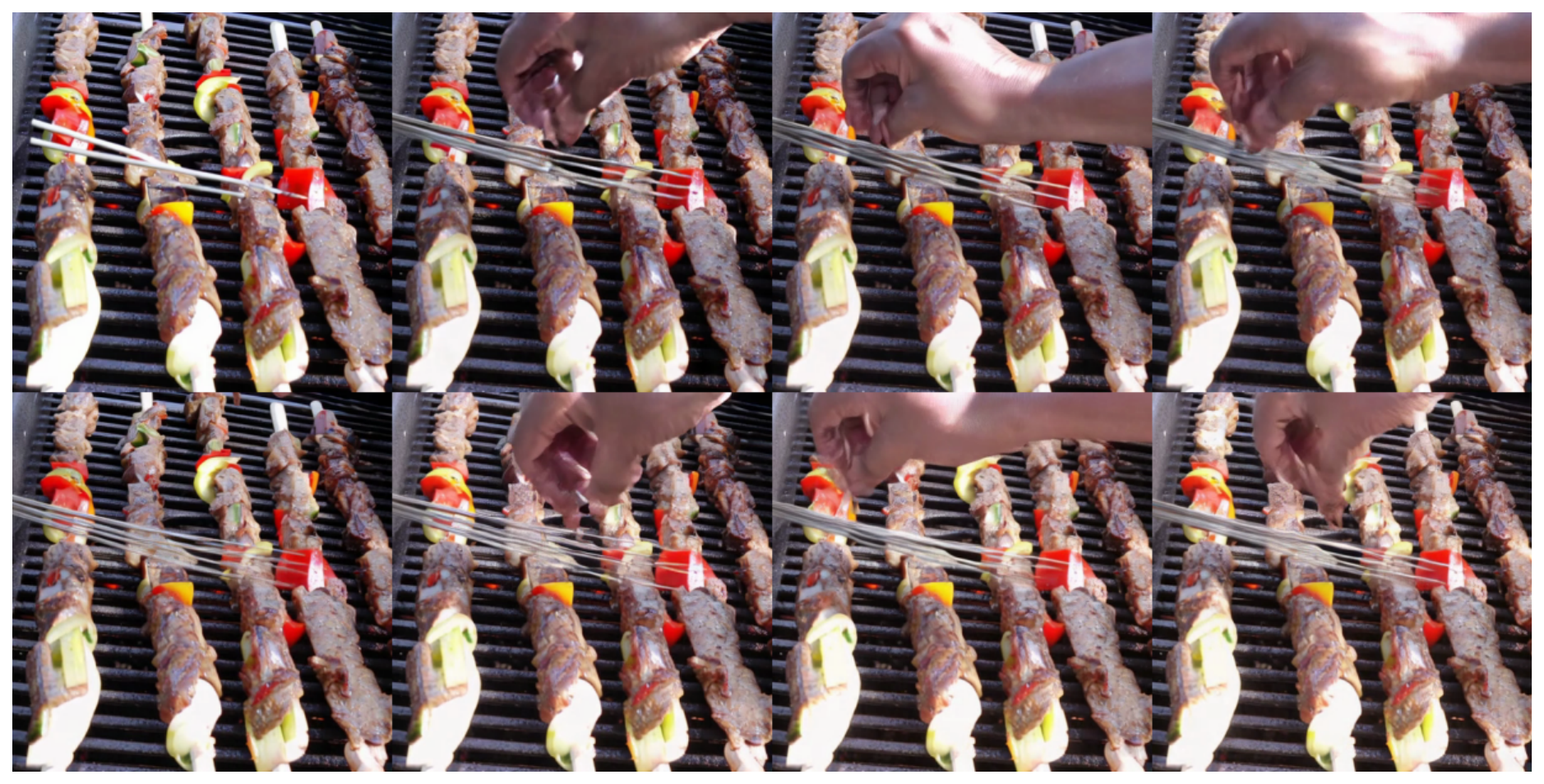}
        \caption{\textbf{Action:} Place the prepared chicken and vegetable kebabs onto a grill.}
    \end{subfigure}

    \begin{subfigure}[b]{0.48\linewidth}
        \centering
        \includegraphics[width=\linewidth]{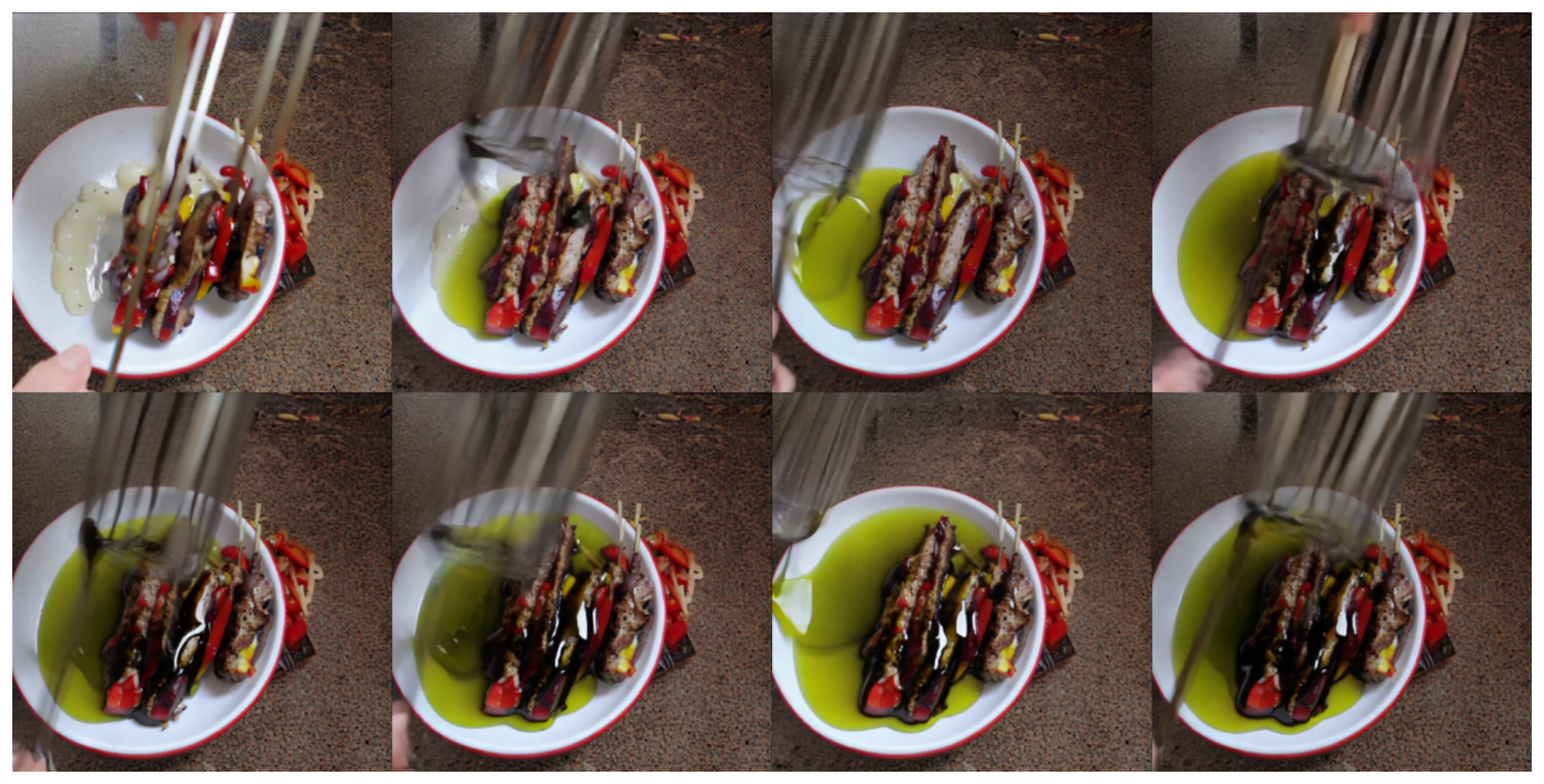}
        \caption{\textbf{Action:} Drizzle olive oil over the chicken and vegetable kebabs.}
    \end{subfigure}
    \hfill
    \begin{subfigure}[b]{0.48\linewidth}
        \centering
        \includegraphics[width=\linewidth]{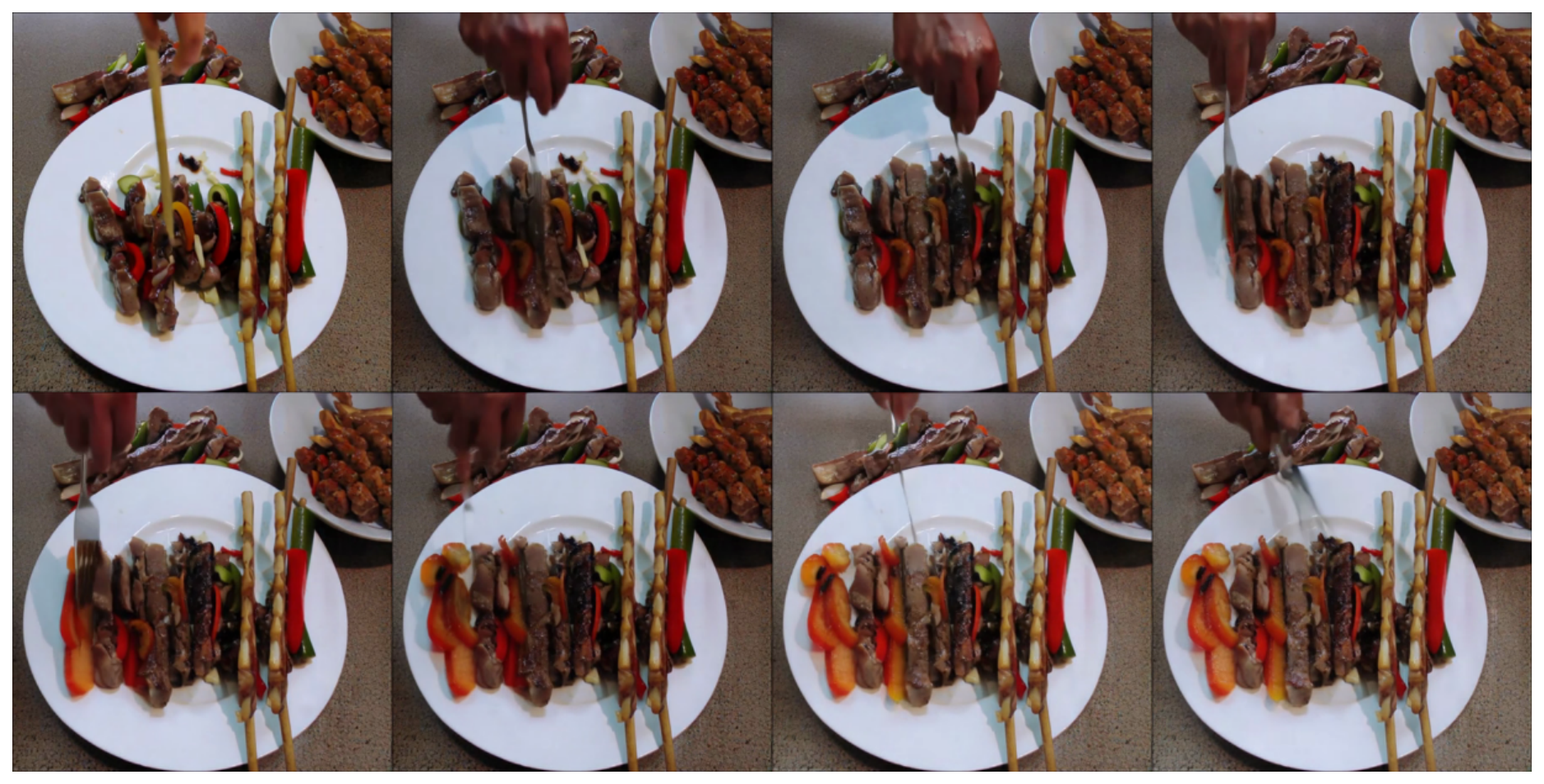}
        \caption{\textbf{Action:} Check on the grilling skewered chicken and vegetable kebabs.}
    \end{subfigure}

    \caption{\textbf{Video generation example.} Our pipeline successfully generates eight crucial steps (\ie, video clips) to prepare the dish "Shish Kabob." This showcases a clear, structured, and instructional step-by-step narrative, demonstrating the model's capability to produce coherent and comprehensive video content.}
    \label{fig:shish_kabob_frames}
\end{figure*}

\Cref{fig:fried_chicken_frames,fig:shish_kabob_frames} present two examples of long narrative video generation for cooking ``Fried Chicken" and ``Shish Kabob," illustrated step-by-step. The generation process begins with our interleaved auto-regressive director, which generates keyframe visual embeddings and their corresponding captions. These embeddings and captions are then used as conditions for the video generation model, which produces high-quality video clips that effectively narrate the cooking process and emphasize the crucial ``action" information. The resulting video clips demonstrate excellent performance in capturing the step-by-step cooking instructions. All video clips are also included in the supplementary materials for further review.

\clearpage
\vspace{-3mm}
\section{Limitations}
\label{sec:appendix_limitations}
\vspace{-2mm}
\subsection{Noisy ``Actions" from ASR}
While our CookGen dataset provides high-quality visual and contextual annotations, the action annotations derived from automatic speech recognition (ASR) have notable limitations. ASR-generated text often contains noise, resulting in action descriptions that are incomplete, ambiguous, or not directly informative for capturing the crucial steps in cooking processes. For instance, in  \Cref{fig:appendix_data_example_jello}(a), the action annotation \textit{“Hi everyone, this one's called rainbow broken glass jello”} offers little value for understanding the cooking process, while another annotation in \Cref{fig:appendix_data_example_jello}(b) \textit{“Now normally when you make jello you use two cups of boiling water”} provides vague guidance without specific details about the method. Such noisy annotations fail to align with the detailed and instructive nature of cooking instructions, which require precision and clarity to guide long narrative video generation effectively. This limitation underscores the importance of refining action annotations to improve their informativeness and utility for modeling cooking tasks.

\vspace{-3mm}
\subsection{Failure Cases}
\vspace{-2mm}
While our method generates high-quality long narrative videos, there are instances where the model fails to produce meaningful cooking steps, and the rendered video clips contain unrealistic or irrelevant content due to hallucination.

\begin{figure*}[h]

    \begin{subfigure}[b]{0.48\linewidth}
        \centering
        \includegraphics[width=\linewidth]{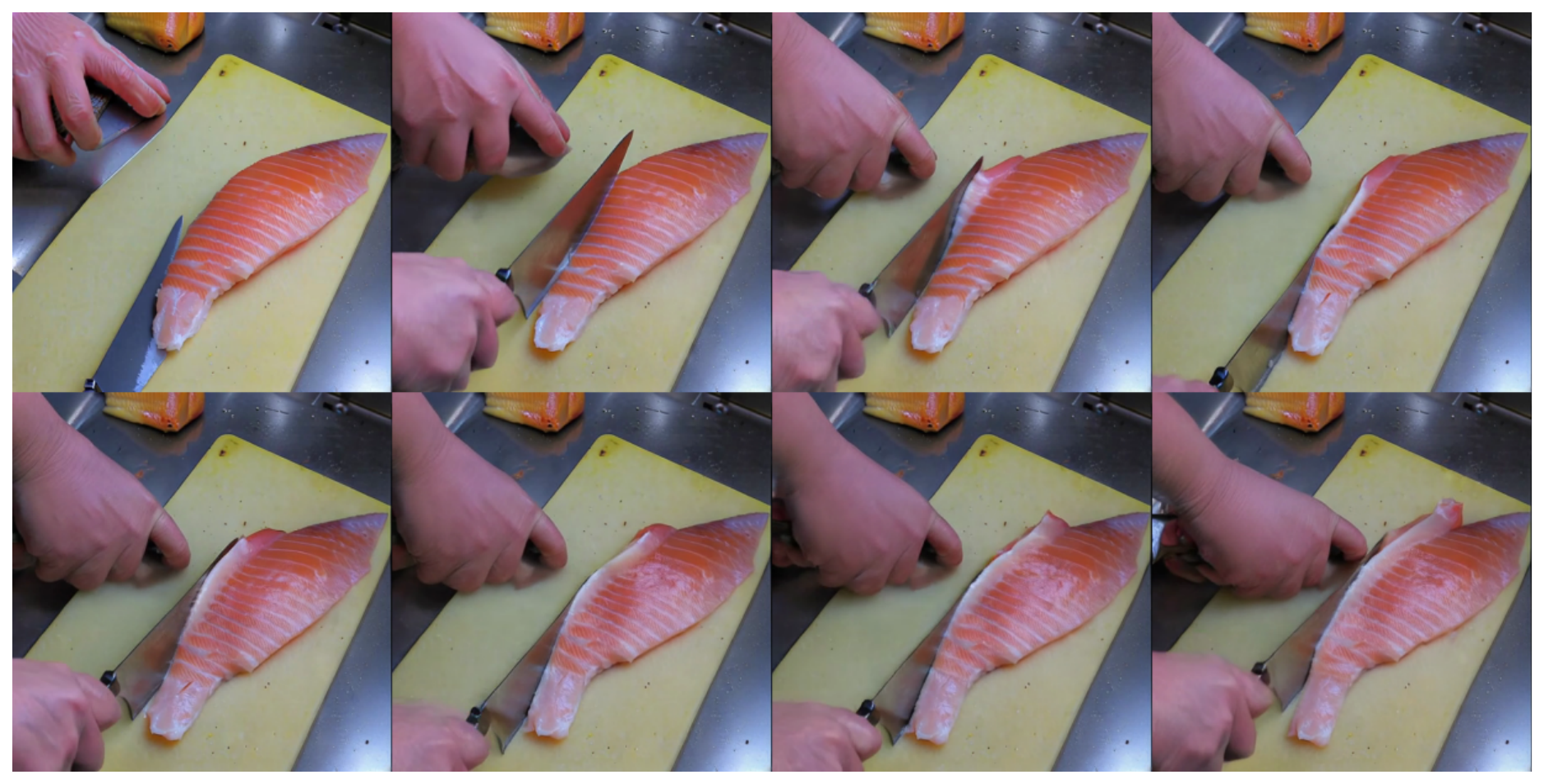}
        \caption{\textbf{Action:} Cutting away the salmon fillet from the backbone}
    \end{subfigure}
    \hfill
    \begin{subfigure}[b]{0.48\linewidth}
        \centering
        \includegraphics[width=\linewidth]{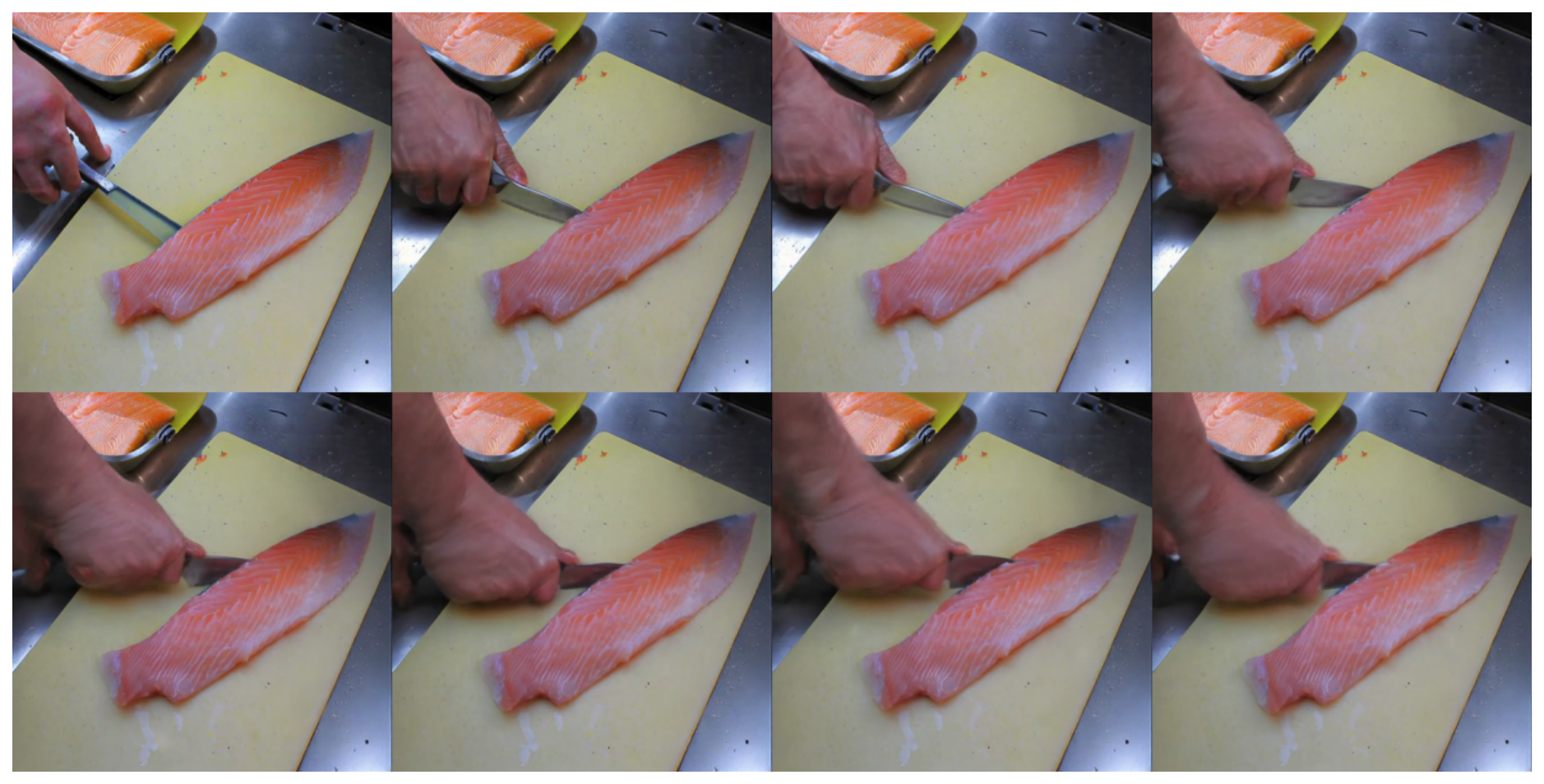}
        \caption{\textbf{Action:} Slicing the salmon fillet into even
    pieces}
    \end{subfigure}
    \vspace{-2mm}
    \caption{\textbf{Failure Case.} Auto-regressive model could  generate repeated ``Steps", which is not informative to viewer.}
    \label{fig:failure_case_repeated}
\end{figure*}

\paragraph{Auto-regressive Director: Repeated ``Steps".} \Cref{fig:failure_case_repeated} illustrates a failure case where the auto-regressive director repeatedly generates similar visual embeddings, resulting in redundant and uninformative cooking video clips. For example, in the provided frames, the generated steps involve repeatedly cutting the salmon fillet, which adds little value to the narrative and fails to progress meaningfully. This issue is a known limitation of auto-regressive models, often caused by a lack of diversity in the embedding generation process. A potential solution is to introduce penalties for repeated embeddings or add constraints to encourage greater variability in visual outputs.
\vspace{-3mm}

\begin{figure*}[h]
    \vspace{-2mm}
    \begin{subfigure}[b]{0.48\linewidth}
        \centering
        \includegraphics[width=\linewidth]{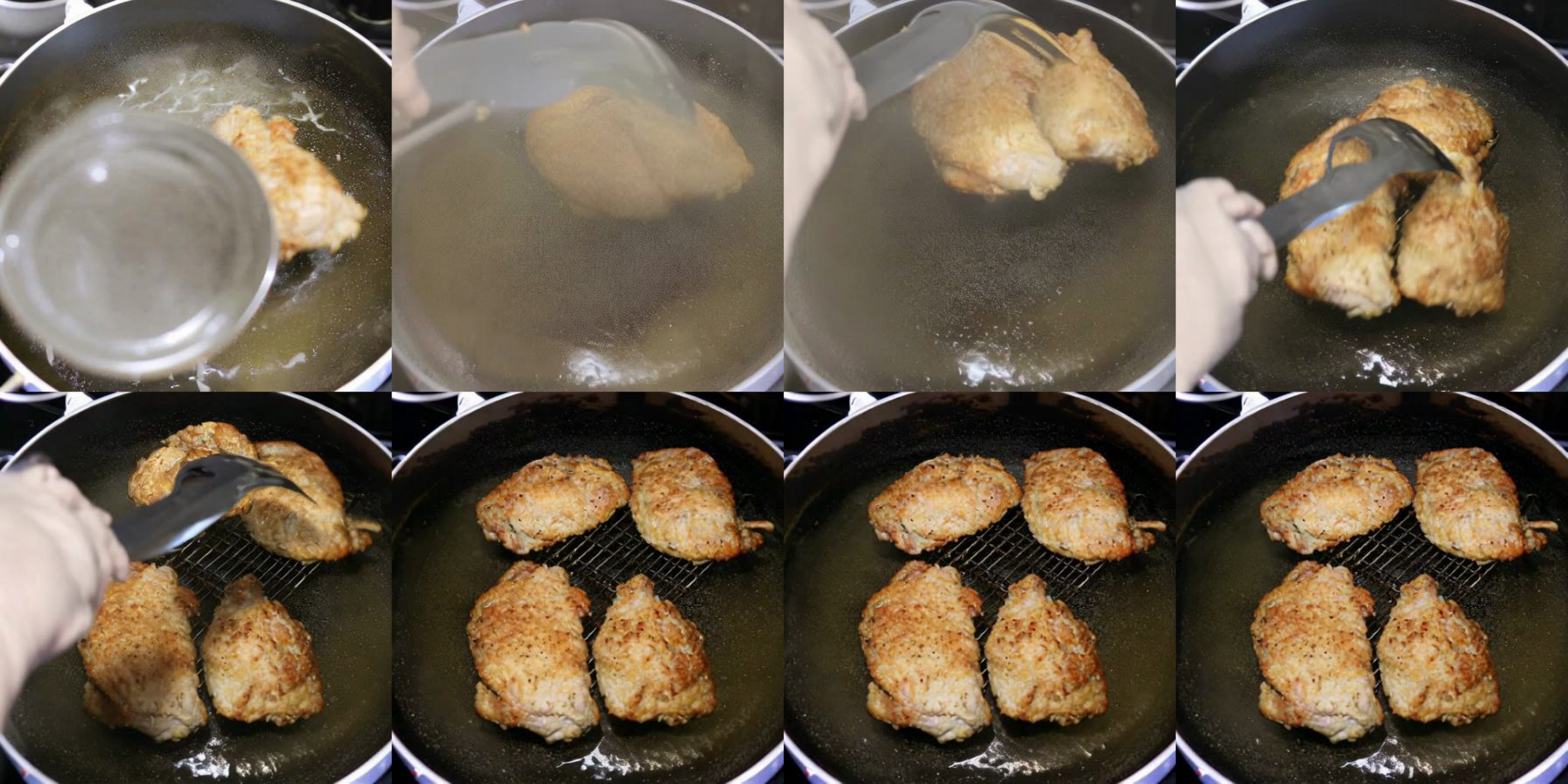}
        \caption{\textbf{Action:} Placing the fried chicken into a oven set to preheat}
    \end{subfigure}
    \hfill
    \begin{subfigure}[b]{0.48\linewidth}
        \centering
        \includegraphics[width=\linewidth]{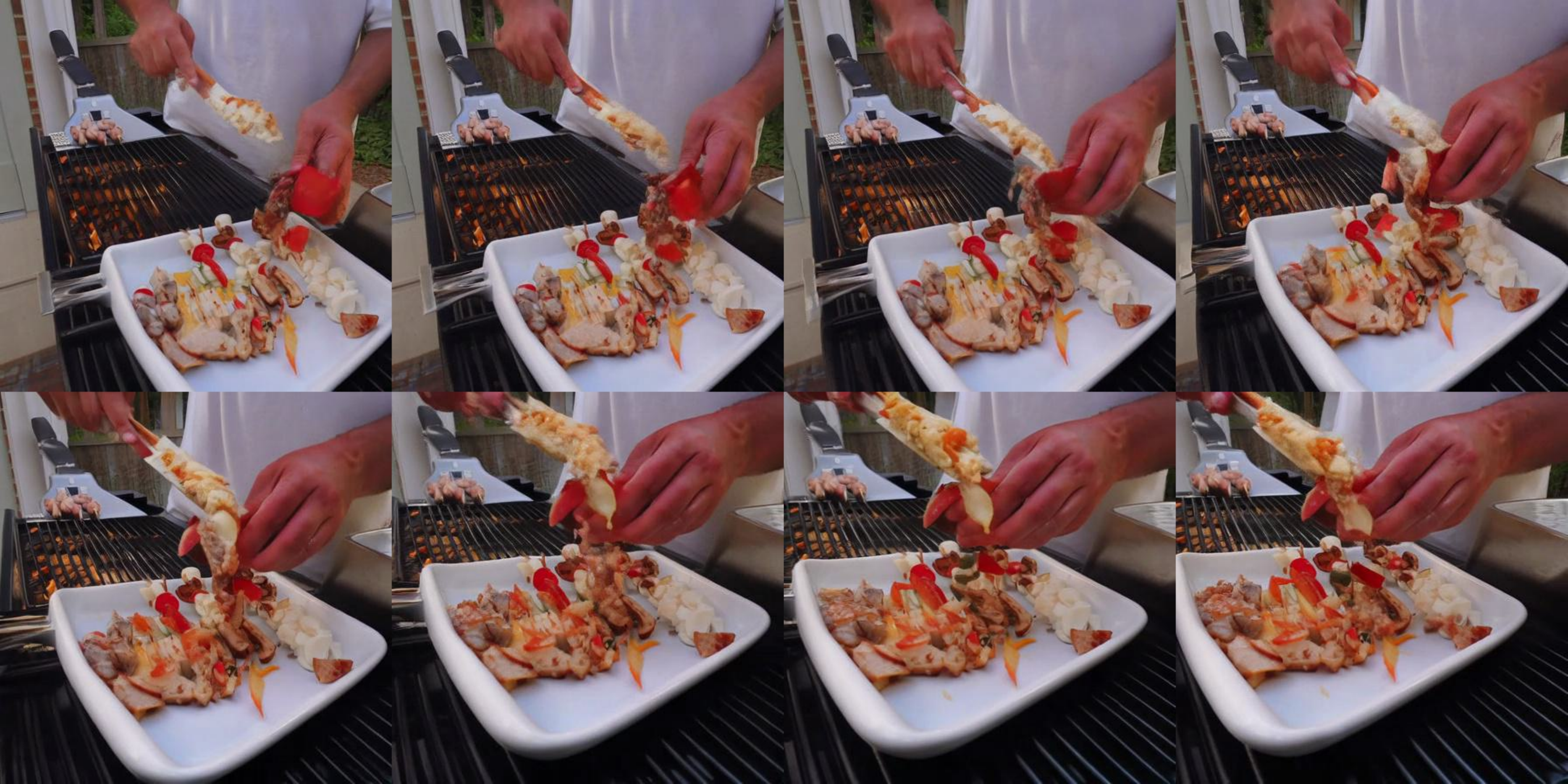}
        \caption{\textbf{Action:} Adding a drizzle of sauce to a
    plate of grilled skewers}
    \end{subfigure}

    \caption{\textbf{Failure Case.} Video generation model could make unrealisc hallucination to generate things from ``air". }
    \label{fig:failure_case_hallucination}
\end{figure*}

\paragraph{Video Generation Model:  Unrealistic Hallucination.} Unrealistic hallucination occurs when a video generation model produces content inconsistent with the intended narrative. In \Cref{fig:failure_case_hallucination}(a), the action "placing the fried chicken into an oven set to preheat" is misrepresented as frying chicken in a pan, with an unrealistic increase in the quantity of chicken, showing a lack of object continuity. In \Cref{fig:failure_case_hallucination}(b), the action "adding a drizzle of sauce to a plate of grilled skewers" introduces an illogical appearance of new grilled food items, deviating from the intended action and disrupting narrative coherence.

